\documentclass{article}

\PassOptionsToPackage{numbers, compress}{natbib}

 \usepackage[final]{neurips_data_2022}

\usepackage[utf8]{inputenc} 
\usepackage[T1]{fontenc}    
\usepackage{hyperref}       
\usepackage{url}            
\usepackage{booktabs}       
\usepackage{amsfonts}       
\usepackage{nicefrac}       
\usepackage{microtype}      
\usepackage{xcolor}         
\usepackage{makecell}

\usepackage{amsmath}
\usepackage{amsfonts}
\usepackage{amssymb}
\usepackage{wrapfig}
\usepackage{subcaption}
\usepackage{multirow}
 \usepackage{mathtools} 

\usepackage{verbatim}

\usepackage{anyfontsize}

\usepackage{microtype}
\usepackage{graphicx}
\usepackage{booktabs} 
\usepackage{multirow}
\usepackage{amsmath,amssymb}
\usepackage{booktabs}
\usepackage{caption,subcaption}

\usepackage{xcolor}
\definecolor{mygreen}{HTML}{3cb44b}
\definecolor{skyblue}{HTML}{beffff}
\definecolor{lightgreen}{HTML}{90ee90}

\usepackage{color, colortbl}

\definecolor{emerald}{rgb}{0.31, 0.78, 0.37}

\usepackage{tcolorbox}
\usepackage{enumitem}
\setitemize{itemsep=10pt,topsep=0pt,parsep=0pt,partopsep=0pt}
\usepackage{colortbl}

\usepackage{xcolor}
\definecolor{mygreen}{HTML}{3cb44b}
\colorlet{myyellow}{green!10!orange!90!}
\makeatletter

\usepackage{tikz}
\usetikzlibrary{arrows,shapes,snakes,automata,backgrounds,fit,petri}
\usepackage{adjustbox}

\newcommand{\RN}[1]{%
	\textup{\lowercase\expandafter{\it \romannumeral#1}}%
}
\usepackage{tabu}

\newcommand{\ie}[0]{\emph{i.e., }}

\newcommand{\eg}[0]{\emph{e.g., }}

\newcommand{\etc}[0]{\emph{etc.}}

\newcommand{\beq}{\vspace{0mm}\begin{equation}}
\newcommand{\eeq}{\vspace{0mm}\end{equation}}
\newcommand{\beqs}{\vspace{0mm}\begin{eqnarray}}
\newcommand{\eeqs}{\vspace{0mm}\end{eqnarray}}
\newcommand{\barr}{\begin{array}}
\newcommand{\earr}{\end{array}}

\newcommand{\Umat}[0]{{{\bf U}}}
\newcommand{\Vmat}[0]{{{\bf V}}}
\newcommand{\Wmat}[0]{{{\bf W}}}

\newcommand{\bv}[0]{{\boldsymbol{b}}}

\newcommand{\hv}[0]{{\boldsymbol{h}}}

\newcommand{\tv}[0]{{\boldsymbol{t}}}
\newcommand{\uv}{\boldsymbol{u}}
\newcommand{\vv}{\boldsymbol{v}}

\newcommand{\xv}{\boldsymbol{x}}

\newcommand{\thetav}{\boldsymbol{\theta}}

\newcommand{\phiv}{\boldsymbol{\phi}}

\newcommand{\R}{\mathbb{R}}

\newcommand{\Xcal}{\mathcal{X}}
\newcommand{\Ycal}{\mathcal{Y}}

\newcommand{\Bcal}{\mathcal{B}}
\newcommand{\Dcal}{\mathcal{D}}

\newcommand{\Tcal}{\mathcal{T}}

\usepackage{color, colortbl}
\definecolor{Gray}{gray}{0.93}

\usepackage{lipsum}

\newcommand\blfootnote[1]{%
  \begingroup
  \renewcommand\thefootnote{}\footnote{#1}%
  \addtocounter{footnote}{-1}%
  \endgroup
}

\usepackage{pifont}
\newcommand{\cmark}{\ding{51}}%
\newcommand{\xmark}{\ding{55}}%

\usepackage{makecell}

\usepackage{xcolor,amsmath}
\usepackage[linesnumbered,ruled,vlined]{algorithm2e}
\DontPrintSemicolon
\usepackage{color, colortbl}
\usepackage{colortbl}
            
\SetKwComment{Comment}{\color{green!50!black}\# }{}

\SetKwProg{Function}{def}{:}{}

\SetKwProg{For}{for}{:}{}
\SetKwProg{If}{if}{:}{}

\newcommand{\shortname}{\textsc{Elevater}}
\newcommand{\longname}{ \textbf{E}valuation of \textbf{L}anguage-augm\textbf{e}nted \textbf{V}isual T\textbf{a}sk-level \textbf{T}ransf\textbf{er}}

\newcommand{\fire}{\includegraphics[width=10px]{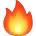}}

\title{\shortname{}: 
A Benchmark and Toolkit for Evaluating
Language-Augmented Visual Models}

\author{
\textbf{\normalsize{Chunyuan Li$^{*1\spadesuit}$, Haotian Liu$^{*2}$, Liunian Harold Li$^{3}$, Pengchuan Zhang$^{1}$, Jyoti Aneja$^{1}$}} \\ \textbf{\normalsize{Jianwei Yang$^{1}$, Ping Jin$^{1}$,   Houdong Hu$^{1}$, Zicheng Liu$^{1}$, Yong Jae Lee$^{2}$, Jianfeng Gao$^{1}$}}\\
\normalsize{$^1$Microsoft~~~~~~$^{2}$University of Wisconsin--Madison~~~~~~$^3$UCLA} 
}

\begin{document}

\maketitle

\maketitle

\begin{abstract}

Learning visual representations from natural language supervision has recently shown great promise in a number of pioneering works. In general, these language-augmented visual models demonstrate strong transferability to a variety of datasets and tasks.
However, it remains challenging to evaluate the transferablity of these models due to the lack of easy-to-use evaluation toolkits and public benchmarks. To tackle this, we build \shortname{}
\blfootnote{$^*$Equal Technical Contribution\hspace{3mm}$^\spadesuit$Project Lead}\footnote{\longname{}}
, the first benchmark and toolkit for evaluating (pre-trained) language-augmented visual models. \shortname{} is composed of three components.
$(i)$ Datasets. As downstream evaluation suites, it consists of 20 image classification datasets and 35 object detection datasets, each of which is augmented with external knowledge.
$(ii)$ Toolkit. An automatic hyper-parameter tuning toolkit is developed to facilitate model evaluation on downstream tasks. 
$(iii)$ Metrics. A variety of evaluation metrics are used to measure sample-efficiency (zero-shot and few-shot) and parameter-efficiency (linear probing and full model fine-tuning). \shortname{} is platform for {\it Computer Vision in the Wild (CVinW)}, and is publicly released  at \url{https://computer-vision-in-the-wild.github.io/ELEVATER/}. 
\end{abstract}

\section{Introduction}
Visual recognition has become ubiquitous in our society~\cite{szeliski2010computer}, with applications in geo-localization~\cite{radford2021learning}, action recognition~\cite{soomro2012ucf101}, street number transcription~\cite{netzer2011reading}, satellite remote sensing~\cite{helber2019eurosat}, medical imaging~\cite{Veeling2018-qh},  self-driving cars~\cite{geiger2013vision}, \etc ~Core to these applications are visual recognition tasks such as image classification (IC) and object detection (OD). It is of high value to develop transferable visual models that perform well on a wide range of downstream applications.
By leveraging large web crawled image-text corpora, recent advances in language-augmented visual models such as  CLIP~\cite{radford2021learning} and ALIGN~\cite{jia2021scaling} have demonstrated strong transfer performance, making this direction one of the most practical visual learning approaches. The reason is twofold: 
$(i)$ open-set recognition is made possible by reformulating classification tasks as retrieval; $(ii)$ model generalization is improved as language supervision significantly increases the coverage of visual concepts for model training.

The success has immediately inspired many studies of large-scale model pre-training~\citep{yang2022unicl,yuan2021florence,yao2021filip,li2021supervision,mu2021slip,gu2021zero,li2021grounded,zhong2021regionclip}. However, these studies use their own evaluation settings based on customized sets of downstream tasks where the detailed process of adapting the models to these tasks is typically not accessible to the public. Thus, it is extremely difficult for researchers to fairly compare models and develop new models based on other people's works.
%
To fill this gap, we develop an open-source benchmark and toolkit, \shortname{}, to make the research results (\eg model's task-level transferability)
more rigorous, and reproducible. 
\shortname{} is composed of three components.


\begin{itemize}[leftmargin=5.5mm]

\item {\bf Benchmark (Datasets and Knowledge).} 
We build the first publicly available benchmark to evaluate the {\em large-scale task-level transferability} of language-augmented visual models. The benchmark consists of two challenges: {\it Image Classification in the Wild (ICinW)} with 20 IC datasets and {\it  Object Detection in the Wild (ODinW)} with 35 OD datasets. A collected external knowledge base for each dataset which could be used for language data augmentation.
\vspace{-2mm}
\item {\bf Comprehensive Metrics.} To measure the cost of deploying models for real-world applications,
we measure a model's sample-efficiency in the zero-shot, few-shot and full-shot settings and parameter-efficiency in the linear probing and full model fine-tuning settings.
\vspace{-2mm}
\item {\bf Reproducible Toolkit \& Language-augmented Adaptation Methods.} 
We develop an open-source software toolkit to support model adaptation and evaluation. Automatic hyper-parameter tuning is employed to avoid human-in-the-loop tuning, thus reducing human labor and ensuring a fair comparison among different model checkpoints. 
We also present a set of new model adaptation methods for 
pre-trained language-augmented visual models. Our methods significantly outperform the traditional vision-only adaptation methods. These methods serve as baselines for the development of more advanced adaptation methods.
\end{itemize}

\begin{figure}[h!]
\vspace{-2mm}
	\centering
	\includegraphics[width=0.99\linewidth]{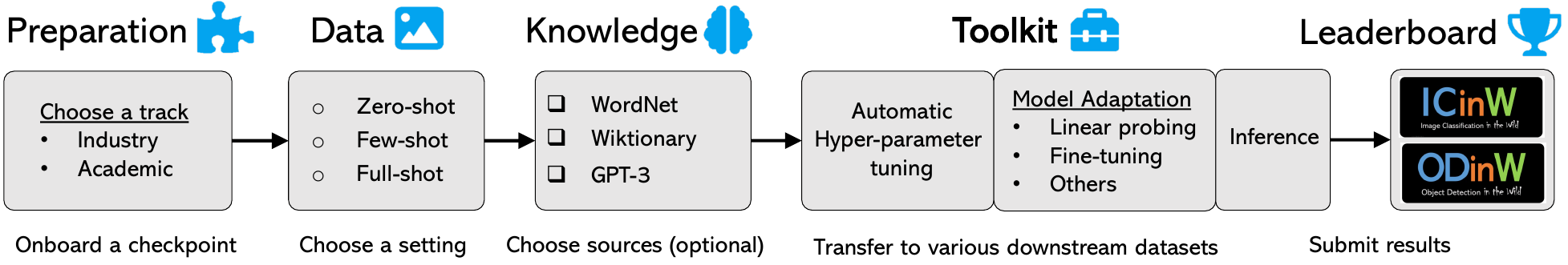}
    \vspace{-1mm}
    \caption{The illustrative pipeline to use \shortname{} to evaluate a model checkpoint.
    }
    \vspace{-4mm}
    \label{fig:pipeline}
\end{figure}

In addition, our empirical study leads to interesting findings.
$(i)$ Leveraging both text and vision in these models consistently yields better performance than vision-only in few-shot settings; In contrast, random initialization of the linear head in language-augmented visual models is sub-optimal.  We also find that few-shot results are always better than zero-shot results, which is different from the results reported in~\cite{radford2021learning}.
$(ii)$
For language-augmented visual models, linear probing performs better than full model fine-tuning in the few-shot settings.
As the task-specific training data increases, fine-tuning outperforms linear probing.
$(iii)$ Our study shows that the use of  external knowledge, including \emph{explicit} knowledge of human-compiled thesaurus/dictionaries/documents and \emph{implicit} knowledge stored in GPT3~\cite{brown2020language}, can improve zero-shot and few-shot learning performance. We summarize the pipeline to use \shortname{} in Figure~\ref{fig:pipeline}, and organize the paper to focus on the benchmark and toolkit.

\vspace{-0mm}
\section{Related Work: From Class-level to Task-level Transfer}\label{sec:related_work}
\vspace{-3mm}
Zero-shot learning in computer vision has been studied for decades. The research topic has witnessed two distinctive notions of zero-shot: the traditional {\it class-level zero-shot} that usually refers to the study of generalizing to unseen object categories~\cite{lampert2009learning}, and the recently popular {\it task-level zero-shot} that refers to the study of generalizing to unseen datasets/tasks~\cite{li2017learning,radford2021learning}.
In Table~\ref{tab:dataset}, we compare our benchmark against existing benchmarks.
Existing zero-shot learning benchmarks are developed for the class-level zero-shot setting. They are usually in a single domain, with a manual split of categories to produce disjoint training and test categories, \eg Animal with Attributes (AwA)~\citep{lampert2013attribute}, Caltech-UCSD Birds-200 (CUB)~\cite{wah2011caltech}, SUN~\cite{patterson2012sun}, aPY~\cite{farhadi2009describing}, and ZS-ImageNet~\cite{rohrbach2011evaluating,fu2016semi}. 
For the OD problem, COCO~\cite{lin2014microsoft} and LVIS~\cite{gupta2019lvis} represent two well established datasets to compare various OD methods in a single domain, while UODB~\cite{wang2019towards} is a multi-domain OD benchmark.



\begin{table}[h!]
    \centering
    \footnotesize
\setlength{\tabcolsep}{2.5pt}
 \scalebox{0.90}{
    \begin{tabular}{c | l cccc | p{8mm} p{8mm} p{8mm}   }
    \toprule
    &  \multicolumn{5}{c}{\bf Benchmark Statistics}   &  \multicolumn{3}{|c}{\bf Evaluation}  \\
   Problem &  & \#Datasets &  \#Image & \#Concepts & Knowledge Source &  Zero & Few & Full  \\
    \midrule
     \multirow{8}{*}{{\makecell{Image \\Classification\\ (IC)} }}
    & AwA~\citep{lampert2013attribute} & 1 & 30337 / 6985   & 40 / 10  & Attributes  & $\checkmark$ &   &   \\
    & CUB~\cite{wah2011caltech} & 1 &  8855 / 2933  & 150 / 50  & Attributes  & $\checkmark$ &   &  \\
    & SUN~\cite{patterson2012sun}  & 1 &  12900 / 1440  & 645 / 72  & Attributes  & $\checkmark$ &   &  \\
    & aPY~\cite{farhadi2009describing}  & 1 &  12695 / 2644  &  20 / 12  & Attributes  & $\checkmark$ &   &  \\
    & ZS-ImageNet~\cite{rohrbach2011evaluating}  & 1 &  1.2M / 54K  &  1K / 360 &  WordNet & $\checkmark$ &   &  \\
    & ImageNet-1K~\cite{deng2009imagenet}  & 1 &  1.2M / 50K  &  1K &  WordNet & $\checkmark$ &   & $\checkmark$ \\   
    & VTAB~\cite{zhai2019large}  & 19 &  2.2M / -  &  940 &  - &  & $\checkmark$  & $\checkmark$ \\       
    & \cellcolor{Gray}\shortname{} (ICinW) &
    \cellcolor{Gray}20 &
    \cellcolor{Gray}638K / 193K &
    \cellcolor{Gray}1151$^{\diamondsuit}$ &
    \cellcolor{Gray}WordNet, Wiki, GPT-3  &
    \cellcolor{Gray}$\checkmark$ &
    \cellcolor{Gray}$\checkmark$ &
    \cellcolor{Gray}$\checkmark$ \\    
    \midrule
    \multirow{4}{*}{\makecell{Object\\ Detection \\(OD)}}
    & COCO~\cite{lin2014microsoft} &  1 &  83K / 41K  & 80  & - &  &   &  $\checkmark$ \\    
    & LVIS~\cite{gupta2019lvis} &  1 &  120K / 40K  & 1723  & WordNet &  &   &  $\checkmark$ \\
    & UODB~\cite{wang2019towards} &  11 &  113K / 40K  & 109  & - &  &   &  $\checkmark$ \\    
    & \cellcolor{Gray}\shortname{} (ODinW) &  
    \cellcolor{Gray}35 &  
    \cellcolor{Gray}132K / 20K &  
    \cellcolor{Gray}314$^{\diamondsuit}$  & 
    \cellcolor{Gray}WordNet, Wiki, GPT-3 & \cellcolor{Gray}$\checkmark$ &
    \cellcolor{Gray}$\checkmark$ &
    \cellcolor{Gray}$\checkmark$ \\   
    \bottomrule
    \end{tabular}
    }
    \vspace{2pt}
    \caption{Comparison of dataset statistics and evaluation settings. For existing zero-shot datasets in IC, the number of images and concepts are reported for development / evaluation stages separately. $^{\diamondsuit}$ represents the total number of concepts in the benchmark to evaluate task-level transfer, and there is no train-evaluation category split as in class-level transfer. \vspace{-2mm}}
    \label{tab:dataset}
    \vspace{-5mm}
\end{table}

In contrast, our benchmark focuses on task-level transfer across domains, \ie it aims to evaluate the transferability of models, by pre-training from their own large corpus, then evaluating zero-shot performance on a diverse set of downstream datasets. This setting has been recently studied~\cite{li2017learning,radford2021learning,li2021align,yuan2021florence}, and is arguably more practical for real-world applications, as it brings the convenience towards the spirit of one-model-for-all. 
The well-known ImageNet-1K dataset~\cite{deng2009imagenet} was originally proposed as a large dataset for model training and testing. It has also recently been considered as one downstream task to study zero-shot transfer~\cite{radford2021learning,li2021align,yuan2021florence}.
Our work presents the first \emph{public benchmark} to standardize the zero-shot task-level transfer setting. Note that visual task transfer has been previously explored in VTAB~\cite{zhai2019large}, which measures good visual representations as those that adapt to diverse, unseen tasks with an emphasis on few training examples. The pre-trained models and task adaptation in VTAB are considered for vision backbones only, and no language model/modality is involved.
Our benchmark shares a similar spirit of task-level transfer to VTAB, but strives to analyze the vital role of language and knowledge in visual transfer.
All of them are usually evaluated in full-shot settings, without considering task-level transfer.
We have further made several novel contributions to consolidate the benchmark: $(i)$ We add external knowledge for each dataset to cultivate new research directions in knowledge-augmented visual models, inspired by the success of knowledge in traditional class-level transfer. $(ii)$ We consider the full spectrum in measuring the sample-efficiency of task adaptation, including zero-shot, few-shot, and full-shot.

\begin{wrapfigure}{r}{0.5\textwidth}
  \begin{center}
  \vspace{-4mm}
    \includegraphics[width=0.5\textwidth]{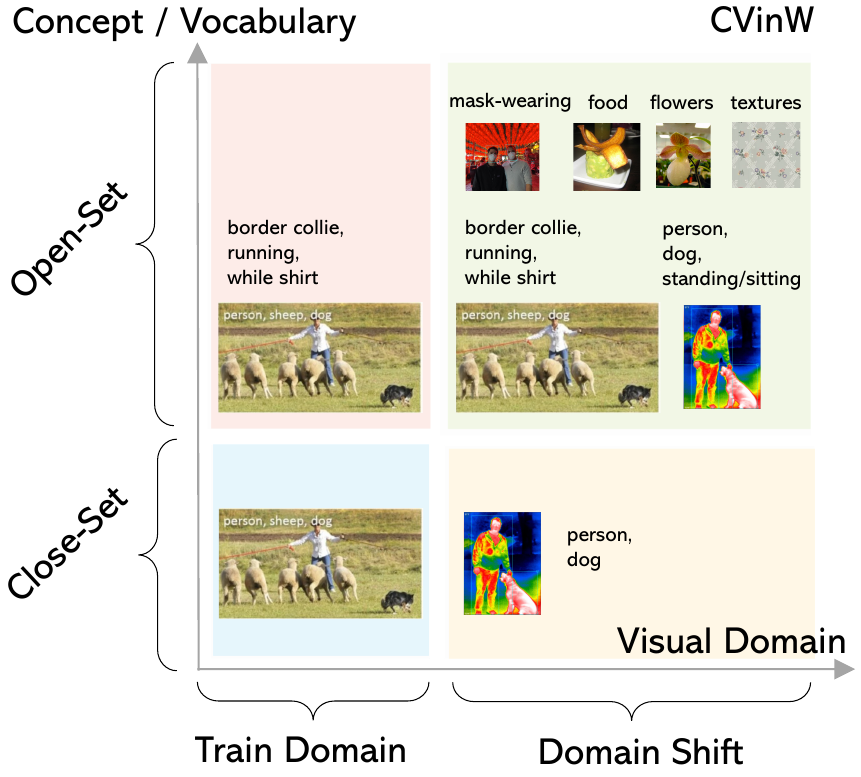}
  \end{center}
    \vspace{-2mm}
    \caption{Illustration of CVinW in comparison with close-set, open-set and domain shift.}
    \vspace{-5mm}
    \label{fig:setting_cvinw}
\end{wrapfigure}

In this paper, we develop \shortname{} as a platform for ``{\bf computer vision in the wild}'', whose ultimate goal is to develop a transferable foundation model/system that can {\it effortlessly} adapt to {\it a large range of visual tasks in the wild}. It consists of two key factors: 
$(i)$ {\bf The task transfer cost is low.}, which is formally defined in Section \ref{sec:evaluation_settings}, where our evaluation metrics is designed with efficiency considerations.
$(ii)$ {\bf The task transfer scenarios are broad}. We illustrate and compare CVinW with other settings using a 2D chart in Figure~\ref{fig:setting_cvinw}, where the space is constructed with two orthogonal dimensions: input image and output concept. The 2D chart is divided into four quadrants, based on how the model evaluation stage is different from model development stage. Both training and evaluation distributions are consistent in both dimensions for the traditional close-set recognition. Open-set recognition allows new concepts in evaluation, while typically remains the same visual domain~\citep{xian2018zero,zareian2021open}; Domain shift allows new visual domain in evaluation, while typically remains the same concept pool~\citep{peng2019moment,gulrajani2020search}. CVinW allows the flexibility in both dimensions, where any new tasks/datasets in the wild essentially fall into.

\vspace{-3mm}
\section{Benchmarks}
\label{sec:benchmark}
\vspace{-3mm}
\subsection{A Suite of Datasets with Language/Knowledge Augmentations}
\label{sec:benchmark_dataset}
\begin{figure}[t!]
\vspace{-3mm}
    \centering
    \begin{subtable}[h]{1.0\textwidth} \centering
    \scalebox{0.95}{
    \begin{tabular}{cc}
    \hspace{-3mm}
     \includegraphics[width=.48\linewidth]{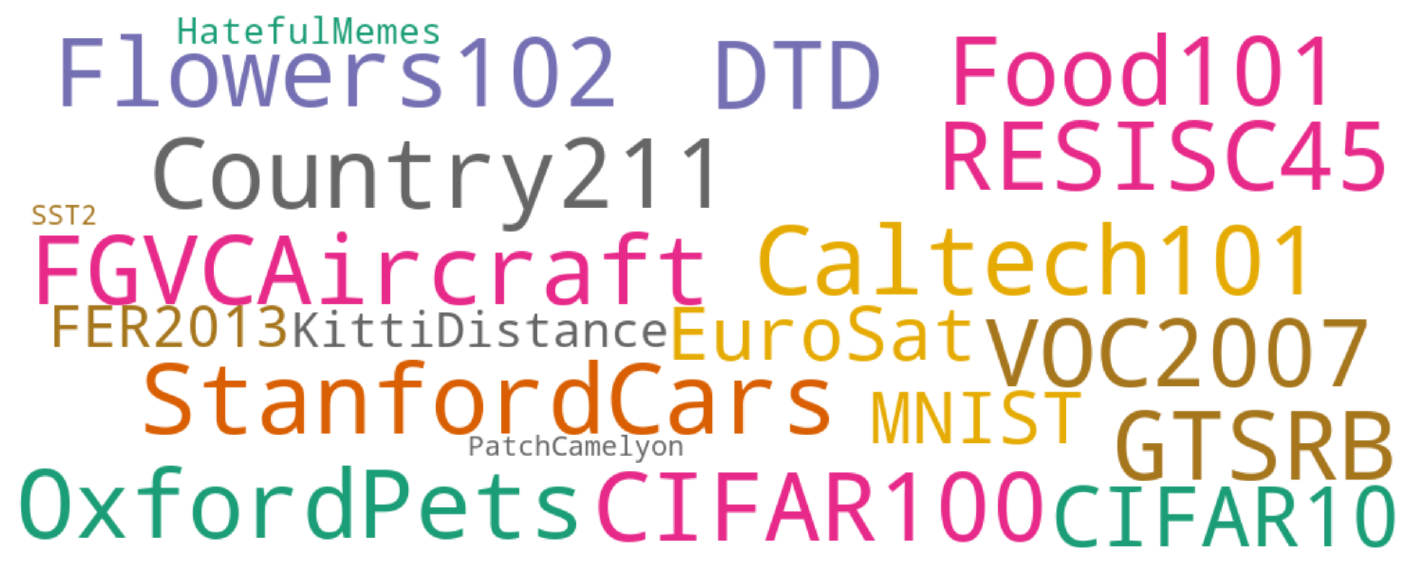}
     & 
     \hspace{-2mm}
     \includegraphics[width=.5\linewidth]{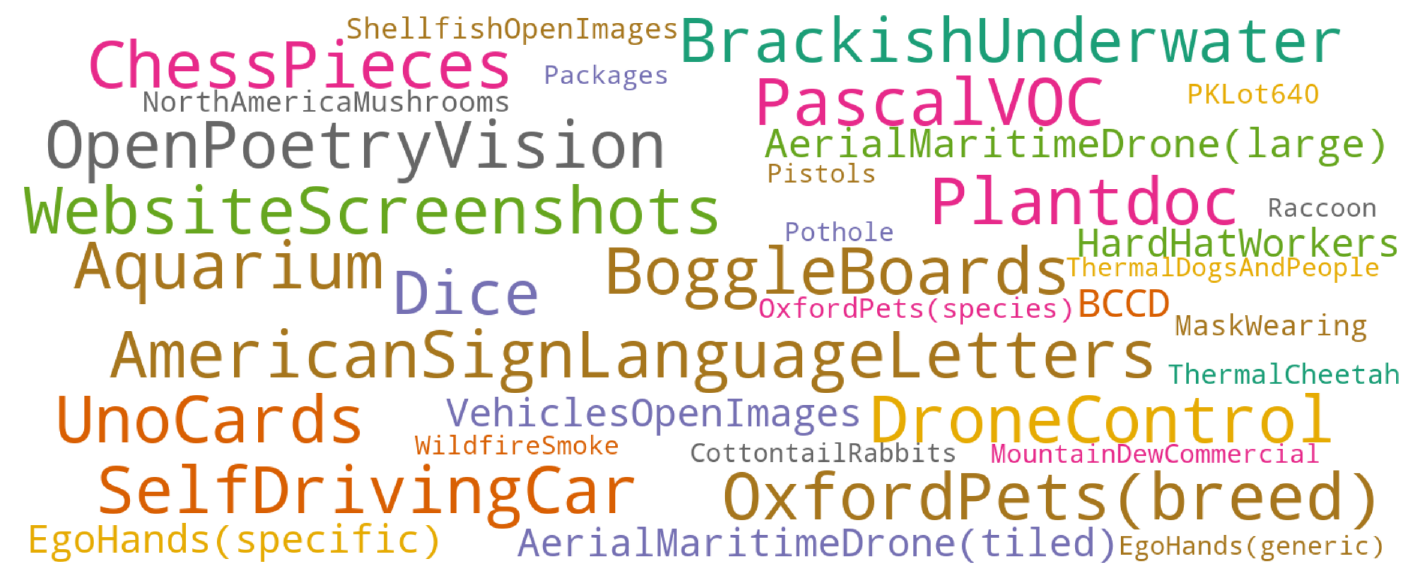}
  \\
    \end{tabular}
    }
    \vspace{-2mm}
   \caption{Dataset names. The font size is proportional to the number of concepts in each dataset.}
	\label{fig:benchmark_dataset_names}
    \end{subtable}    

    \begin{subtable}[h]{0.98\textwidth} \centering
    \vspace{2mm}
    \scalebox{0.95}{
    \begin{tabular}{cc}
    \hspace{-4mm}
     \includegraphics[width=.48\linewidth]{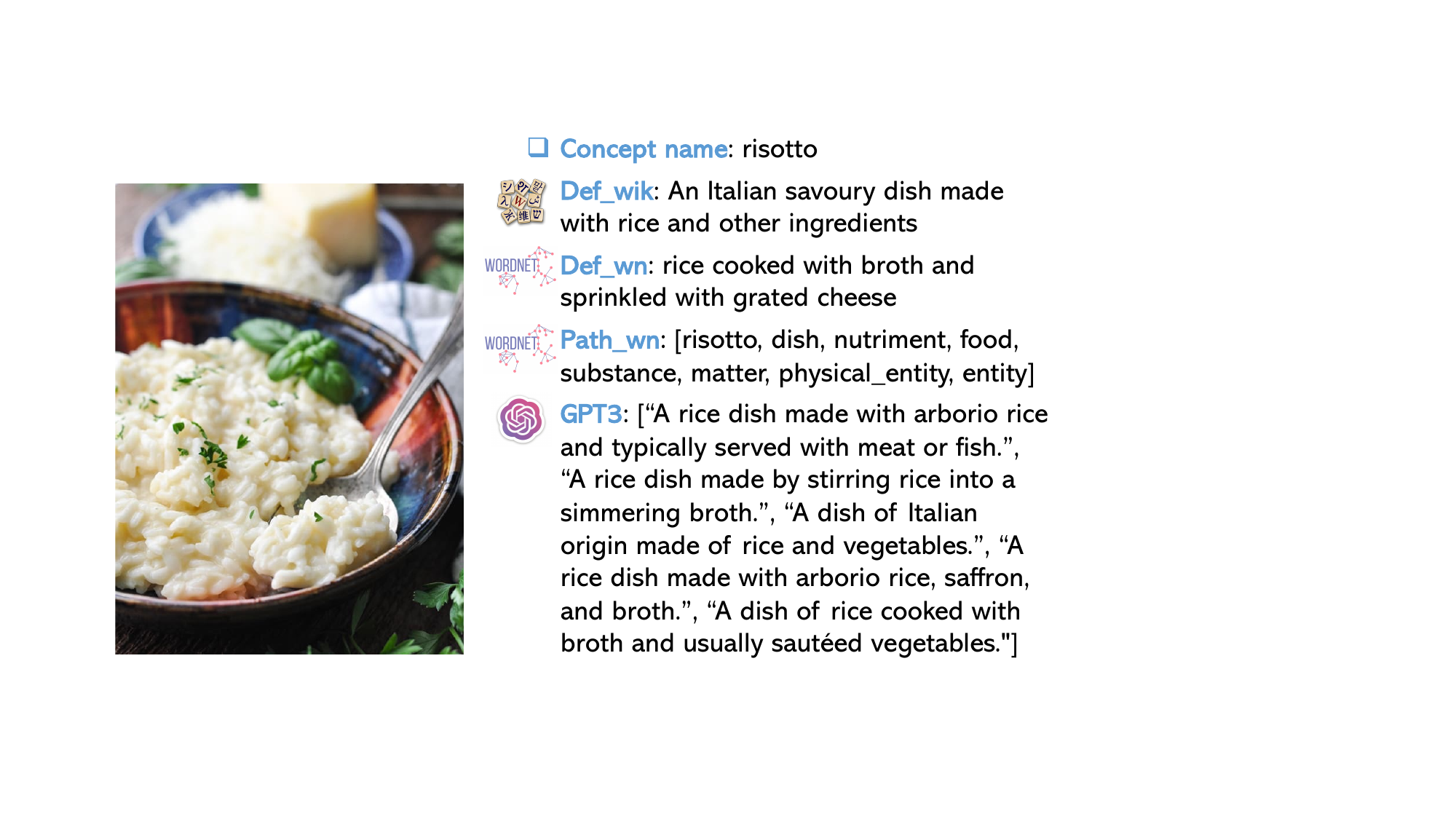}
     & 
     \hspace{-4mm}
     \includegraphics[width=.51\linewidth]{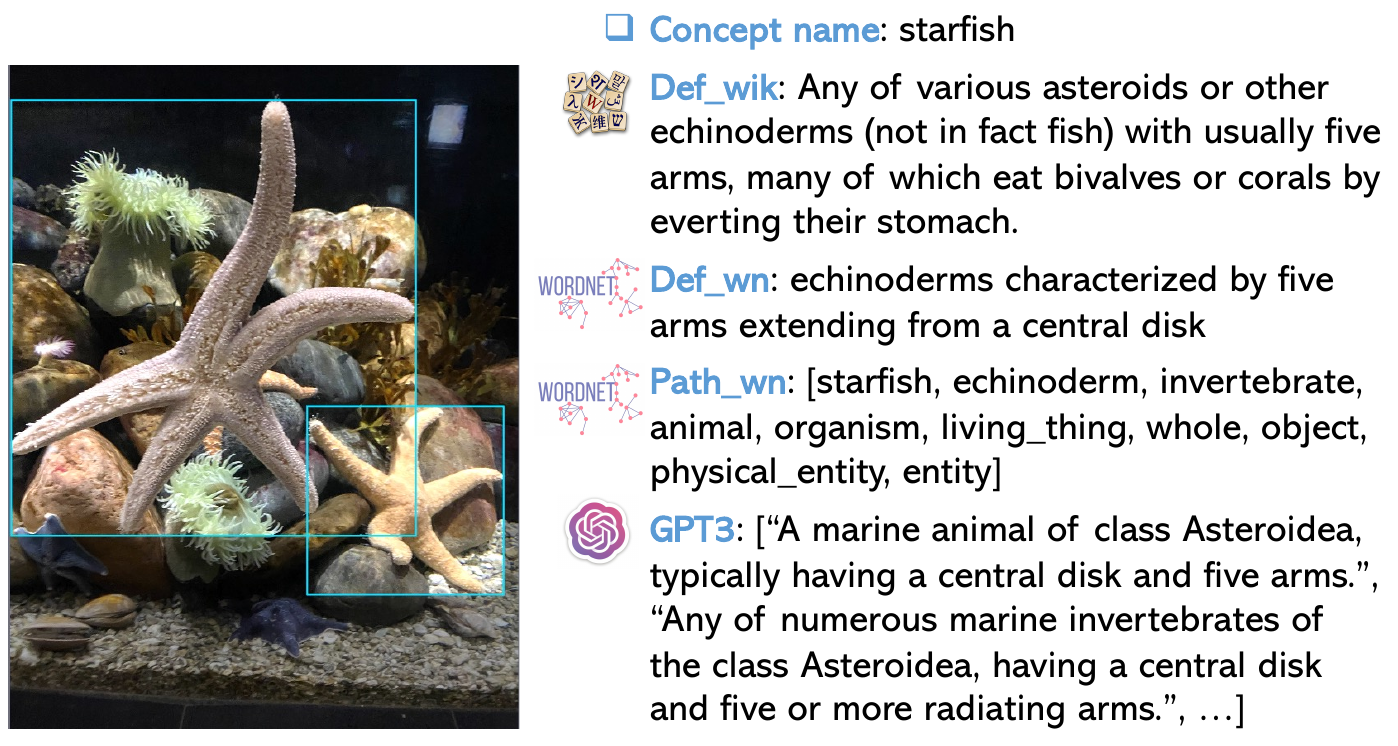}
  \\
    \end{tabular}
    }
    \vspace{-2mm}
   \caption{Examples of collected external knowledge.}
	\label{fig:benchmark_examples}
    \end{subtable}    
    \vspace{-2mm}
    \caption{Illustration of our benchmark. Left: Image classification, Right: Object detection.}
    \label{fig:benchmark}  
    \vspace{-4mm}
\end{figure}

\vspace{-2mm}
As a proxy for performing unseen tasks in the wild, we collect a diverse set of public datasets from various domains in computer vision, as the basis of our benchmark. Specifically, we consider 20 datasets for IC and 35 datasets for OD. We exhibit the dataset names in Figure~\ref{fig:benchmark} (a), and the detailed statistics of each dataset in Table~\ref{table:downstream_ic_dataset} and Table~\ref{table:downstream_od_dataset} in Appendix. It is recommended in~\cite{radford2021learning} that studying task-level zero-shot transfer is a way of measuring the task learning capabilities of machine learning systems. The task definition of each downstream recognition dataset is typically specified using category names. Adding user specification/note is a natural way to clarify the task definition \eg the attribute or explanation of a visual concept.
Importantly, a similar spirit has been implemented in traditional class-level zero-shot by adding individual domain-specific knowledge (see Table~\ref{tab:dataset}), and demonstrated promising zero-shot performance gains. In this paper, we generalize the notion of ``zero-shot'' to task-level, collecting external knowledge from general sources for our benchmark.


\begin{itemize}[leftmargin=2.5mm]
\vspace{-1mm}
\item \textit{WordNet Hierarchy}  ($\texttt{def\_path}$). The words along the traversal path from the query node in WordNet~\citep{miller1998wordnet} to the highest parent node is recorded as the hierarchy knowledge.
\vspace{-2mm}
\item \textit{WordNet Definition} ($\texttt{def\_wn}$). The definition in WordNet synsets~\citep{miller1998wordnet} is used to explain the query.
\vspace{-2mm}
\item \textit{Wiktionary Definition} ($\texttt{def\_wik}$). The definition of a query in  Wiktionary~\cite{meyer2012wiktionary} is used.
\vspace{-2mm}
\item \textit{GPT3 Knowledge} ($\texttt{gpt3}$). For the above three knowledge sources, it is not always feasible to retrieve valid knowledge for any query. To enable full knowledge coverage, we propose to use GPT3~\cite{brown2020language} to generate ``pseudo'' knowledge using in-context-learning, where prompts are constructed with multiple pairs of class names and their Wiktionary definitions. We generate five GPT3 knowledge sequences for each class name, by constructing different context prompts with randomly sampled pairs. See details in Section~\ref{sec:gpt3_prompt}.

\end{itemize}

In Fig.~\ref{fig:benchmark}~(b), we show examples to illustrate the knowledge sources. In practice, there is a trade-off between the knowledge quality and its coverage. For example, WordNet has relatively rich and precise knowledge, but the coverage is low; GPT3 knowledge has the full coverage (as it is generated via prompting a pre-trained neural language model), but it is hard to assess its quality. In the experiment section, we provide baseline results to demonstrate the benefits of external knowledge, and encourage the community to design advanced prompting techniques to leverage these knowledge sources.

\subsection{Pre-trained Models for Transfer Learning}
\vspace{-2mm}
\paragraph{Industry Track and Academic Track.} Our benchmark is an evaluation platform for pre-trained models, whose performance largely depends on the scale of the pre-training corpus. Larger corpus typically yields higher performance, but unfortunately results in a barrier to many participants, especially a majority of researchers from university labs. To increase inclusivity, we create two tracks with restrictions on the pre-training data scale: 
$(i)$
{\it Academic track} is a setting that limits the data in established public large datasets (\ie ImageNet-21K~\cite{deng2009imagenet}, GCC3M~\cite{sharma2018conceptual} \& 12M~\cite{changpinyo2021conceptual}, YFCC15M~\cite{thomee2016yfcc100m}). This track is more academic-friendly, aiming to encourage the exploration in data-efficient pre-training methodologies. 
$(ii)$
{\it Industry track} has no limit on pre-training data scale, except that images in our benchmark are not allowed in pre-training when reporting zero-/few-shot performance. This track aims to explore the scaling limit.
We encourage participants to report the pre-training datasets to enable reproducible research.

\paragraph{Pre-trained Models.}
To establish baseline results on \shortname{}, we evaluate the pre-trained model checkpoints in Table~\ref{tab:ckpts}. 
More details of the checkpoints are described in Appendix. Most existing visual models are language-free, where no text is used in model training. Till recently, visual models are trained in a language-augmented and/or knowledge-augmented manner using a language model~\cite{radford2021learning,yang2022unicl,shen2022klite,li2021grounded}, among which CLIP~\cite{radford2021learning} represents a strong baseline in the industry track. Please see the detailed taxonomy in Appendix Section~\ref{sec:taxonomy}.

\begin{table*}[t!]
    \centering
    \footnotesize
    \scalebox{0.85}{
\begin{tabular}{l| >{\centering}p{0.9cm} >{\centering}p{1.3cm}| cc|ccc}
    \toprule
        
       \multirow{2}{*}{Checkpoints}
       & \multicolumn{2}{c|}{ \multirow{1}{*}{Taxonomy} }
       & \multicolumn{2}{c|}{ \multirow{1}{*}{Pre-training Settings} } 
       & \multicolumn{3}{c}{ \multirow{1}{*}{Network Architecture} } 
        \\
       \cmidrule{2-8} 
       & Language & Knowledge & Training Objective & Dataset & Vision & Language & Others  \\
        \midrule
        \multicolumn{8}{c}{ ~~\bf{Image Classification} } \\
        MoCo-v3~\cite{chen2021empirical}  & \xmark & \xmark  & Self-Supervised & ImageNet-1K (1.2M)  & ViT-B & - & - \\ 
        MAE~\cite{he2021masked}  & \xmark & \xmark   & Self-Supervised & ImageNet-1K (1.2M)  &  ViT-B & - & - \\ 
        DeiT~\cite{touvron2021training}  & \xmark & \xmark  & Supervised & ImageNet-1K (1.2M)  &   ViT-B & - & - \\ 
        ViT~\cite{dosovitskiy2020image}  & \xmark & \xmark   & Supervised & ImageNet-22K (14M)  & ViT-B & - & - \\ 
        CLIP~\cite{radford2021learning} & \cmark & \xmark   & Image-Text Contrast  & WebImageText (400M)  &  ViT-B & T-B & - \\
        UniCL~\cite{yang2022unicl} & \cmark & \xmark  & Image-Text Contrast & ImageNet-21K (13M)    & Swin-T & T-B & - \\ 
        K-LITE~\cite{shen2022klite} & \cmark & \cmark & Image-Text Contrast & ImageNet-21K (13M)    & Swin-T & T-B & - \\
        \midrule
        \multicolumn{8}{c}{ ~~\bf{Object Detection} } \\
        DyHead~\cite{dai2021dynamic}  & \xmark & \xmark &  Supervised & Object365   & Swin-T  &  - &  - \\ 
        GLIP~\cite{li2021grounded} & \cmark & \xmark    & Supervised  & Object365 \& Grounding  & Swin-T & Bert-B & Fusion \\ 
        GLIP-A~\cite{li2021grounded} & \cmark & \xmark   & Supervised  & Object365   & Swin-T & Bert-B & -  \\ 
        K-LITE~\cite{shen2022klite} & \cmark & \cmark   & Supervised  & Object365  & Swin-T & Bert-B  & -  \\ 
        \bottomrule
    \end{tabular}
    }
    \vspace{-0mm}
    \caption{The pre-trained models evaluated in \shortname{} as baselines. In terms of taxonomy, \cmark~indicates the model checkpoint is pre-trained with the use of language / knowledge, while \xmark~indicates language- / knowledge-free. For image classification, the number of images in pre-training is reported. T-B indicate a Base-size Transformer architecture, using a 63M-parameter 12-layer 512-wide model with 8 attention heads. Swin-T is a Tiny-size Swin Transformer~\cite{liu2021Swin}, Bert-B is a Base-size Bert~\cite{devlin2018bert}, and Fusion indicates a cross-attention module to fuse the image-text features~\cite{li2021grounded}.
    }
    \label{tab:ckpts}
      \vspace{-3mm}
\end{table*}

\subsection{Evaluation Settings: Efficiency Considerations}
\label{sec:evaluation_settings}
\vspace{-2mm}

\begin{wrapfigure}{r}{0.6\textwidth}

  \begin{center}
  \vspace{-7mm}
    \includegraphics[width=0.6\textwidth]{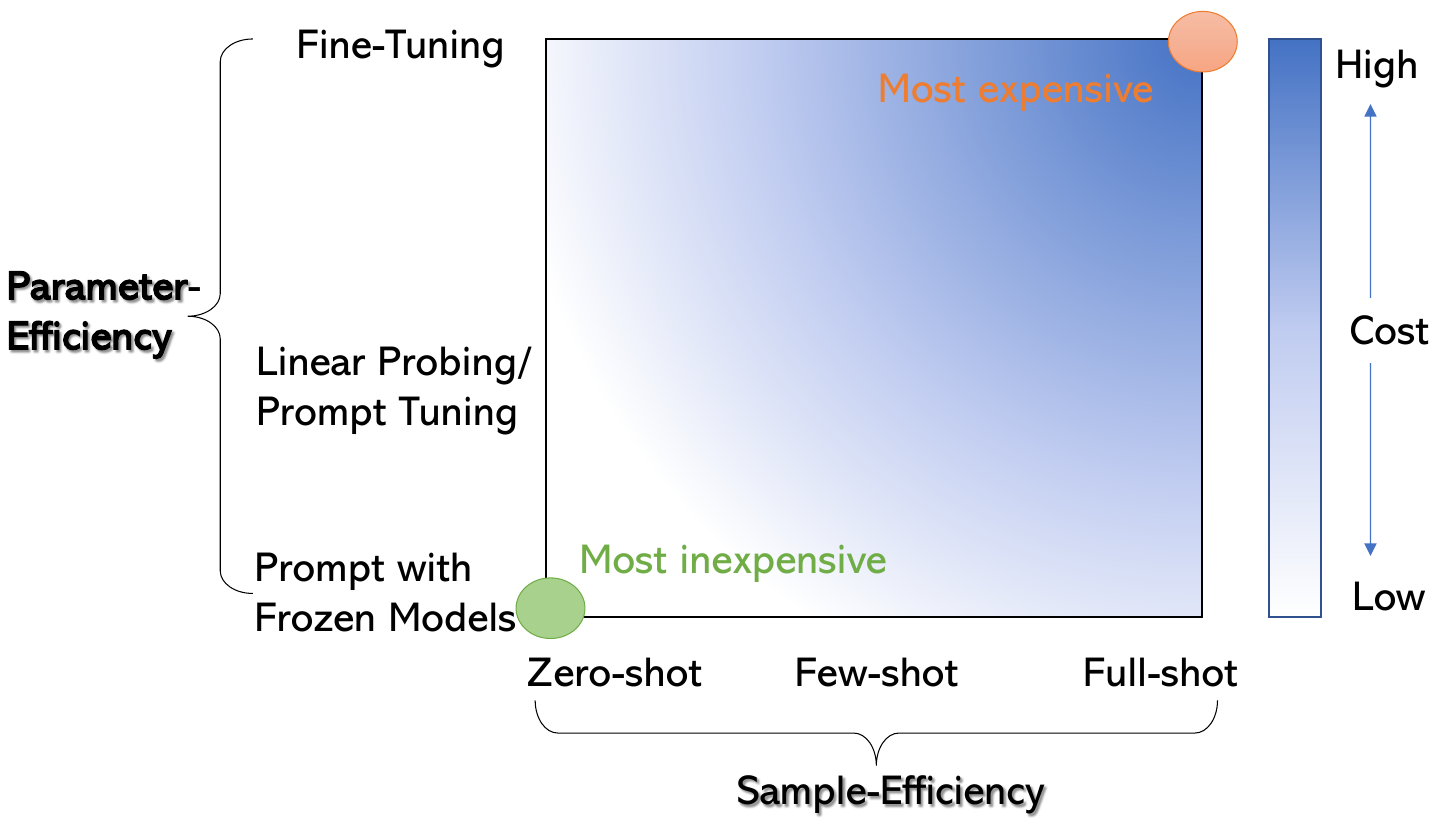}
  \end{center}
    \vspace{-4mm}
    \caption{The model adaptation cost chart.  }
    \vspace{-7mm}
    \label{fig:adapation_cost}
\end{wrapfigure}

One major advantage of pre-trained models is the promise that they can transfer to downstream tasks {\it effortlessly}.  The cost is considered in two orthogonal dimensions: sample-efficiency and parameter-efficiency, as illustrated in Figure~\ref{fig:adapation_cost}.  The bottom-left corner and  top-right corner is the most inexpensive and  expensive adaptation strategy, respectively. One may interpolate and  make combinations in the 2D space, to get different model adaptation methods with different cost.

\paragraph{Sample-efficiency: Zero-, Few-, and Full-shot.} Due to the high cost of annotating data, it is often desired to provide a small number of labeled image-label pairs in downstream datasets. Transferable models should be able to reach high performance in this data-limited scenario. To assess this ability, we vary the number of training set size $N$ per category in the downstream dataset.
For IC, $N=0, 5, 20, 50$. For OD\footnote{For OD, $N$-shot means providing at least $N$ images per category\cite{wang2020frustratingly,li2021grounded}}, $N=0, 1, 3, 5, 10$. Three random seeds are chosen, each of which identifies a subset of samples from the full dataset in a deterministic manner. Once the random seed is given, the indices of training samples in few-shot settings are fixed to encourage reproducible research. We also consider the full-shot setting, where all samples of a given dataset are used. \vspace{-2mm}

\paragraph{Parameter-efficiency: Linear Probing vs Full Model Fine-tuning.} Maintaining a small number of dataset-specific model parameters is often favored for model maintenance, as it can be expensive to maintain a unique copy of large model checkpoints for each of the thousands of downstream applications. In IC, linear probing provides a simple strategy for training a dataset-specific linear embedding matrix, while keeping the pre-trained visual backbone frozen. It arguably represents the minimum cost solution for parameter-efficiency. In contrast, fine-tuning often updates the entire weights in backbone and linear head, representing the most expensive solution to model adaptation. In OD~\cite{wang2020frustratingly}, linear probing means updating the linear heads for classification and localization tasks only, while fine-tuning means updating all model weights including the backbone and the detectors.


\vspace{-2mm}
\section{Toolkits}\label{sec:toolkit}
\begin{figure}[t!]
\vspace{-2mm}
	\centering
	\includegraphics[width=0.99\linewidth]{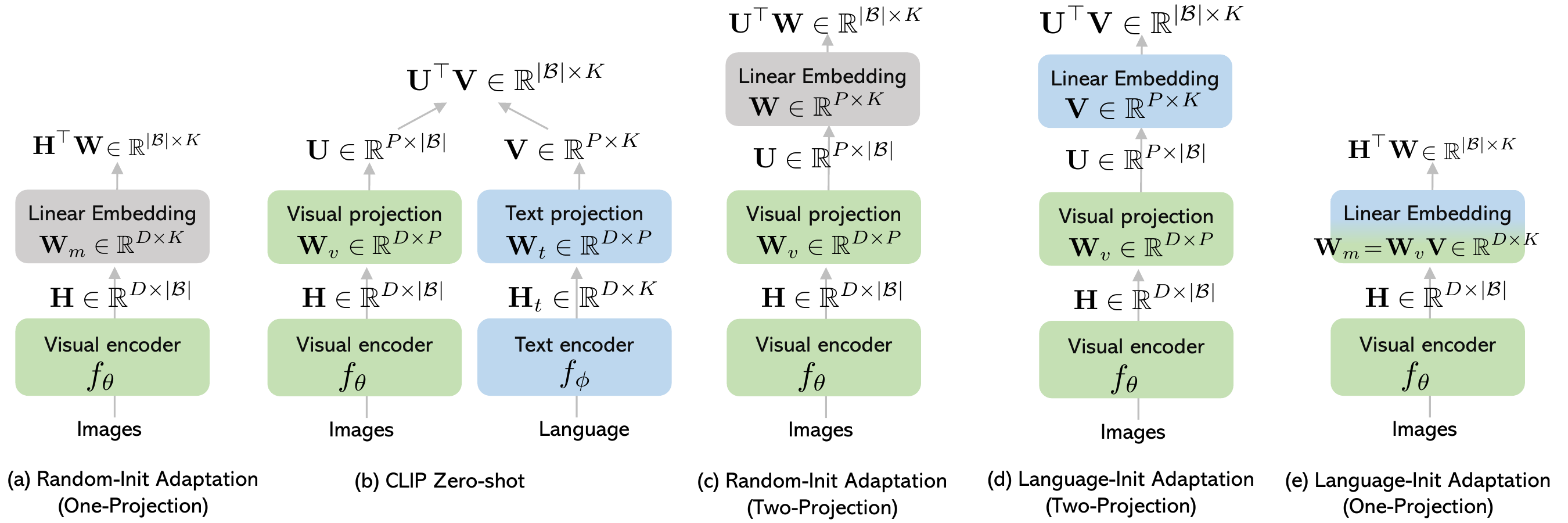}
    \vspace{-1mm}
    \caption{Illustrative comparison of different model evaluation and adaption methods.
    }
    \vspace{-4mm}
    \label{fig:adaptation}
\end{figure}

\vspace{-3mm}
To ease the process to onboard new checkpoints for evaluation, we provide a software toolkit, including $(i)$ automatic hyper-parameter tuning and $(ii)$ various strategies for model adaptation to downstream tasks.
First, automatic hyper-parameter tuning pipeline is developed to avoid human-in-the-loop tuning, thus reducing human labor and ensuring
fair comparisons of different model checkpoints. We follow CLIP~\cite{radford2021learning} to implement a simple grid-search style tuning pipeline, and leave more sophisticated methods like BOHB~\citep{falkner2018bohb} and DEHB~\cite{awad2021dehb} as future work. Details are provided in Appendix.
Second, we provide several model adaptation methods as strong baselines, which allow effective transfer learning of pre-trained visual models. The ideas are illustrated in Figure~\ref{fig:adaptation}. For a downstream dataset, we first represent it in a triplet-wise data format $\Dcal  = \{ (\xv_n, \tv_n, y_n) \}_{n=1}^N$, where $\xv \in \Xcal$  is the image, $\tv \in \Tcal$ is its corresponding language description, and $y \in \Ycal$ is a label indicating the index of the unique language description in the dataset. $|\Bcal|$ is batch size. In IC, the number of labels $|\Ycal| = K$, \ie the number of category names.  


\paragraph{Language-free Visual Models.} Most existing visual models are language-free, as language is often not considered in training, \eg supervised and self-supervised methods. Such models can not be directly used for zero-shot transfer, and model adaptation is often enabled by adding additional weights. For each image $\xv$, an image encoder $f_{\thetav}$ parameterized by $\thetav$ first represents $\xv$ as a visual feature vector $ \hv \in \R^{D\times 1}$: $  \hv  = f_{\thetav}(\xv)$. One randomly initialized linear projection layer with $\Wmat_m \in \R^{D\times K}$ (we absorb the bias $\bv$ in $\Wmat_m$ for simplicity) is used as the classifier; see  Figure~\ref{fig:adaptation} (a). 

\paragraph{Language-augmented Visual Models.} Recent works~\cite{radford2021learning,jia2021scaling} that learn visual models with language supervision often employ a two-encoder architecture. Besides the image encoder model $f_{\thetav}$, a text encoder $f_{\phiv}(\tv)$ parameterized by $\phiv$ is also used to encode text $\tv$. Additional projection layers $\Wmat_v$ and $\Wmat_t$ are introduced for image and language features, embedding them into a joint space with dimension $P$, with projected features as $\uv$ and $\vv$ respectively. Note that lowercase $\uv$ and $\vv$ are single feature vectors while $\Umat$ and $\Vmat$ are a batch with multiple feature vectors. As in Figure~\ref{fig:adaptation} (b), zero-shot learning can be directly performed in this space: the mean text feature $\vv$ is first obtained for each category, by averaging text features of the category name in different language prompts. The image is predicted as the category yielding the highest similarity $\uv^{\top}\vv$.

\begin{itemize}[leftmargin=2.5mm]
\vspace{1mm}

\item \textit{Random initialized Adaptation.} In the original CLIP paper~\cite{radford2021learning}, one randomly initialized linear projection layer $\Wmat \in \R^{P \times K}$ (similarly, we absorb the bias $\bv$ in $\Wmat$ for simplicity) is added on the pre-retained visual projection, which is shown as the two-projection method in Figure~\ref{fig:adaptation} (c).

\item \textit{Language-initialized Adaptation.} 
We argue that the full capacity of language-augmented visual models is not leveraged in~\cite{radford2021learning}. The power of pre-trained language encoder and text inputs must play a vital role in model adaptation. Hence, we propose two language-initialized adaption methods, each of which is ensured as a fair comparison variant for language-augmented and language-free models, respectively. 
$(i)$ {\it Two-Projection.} For the linear head $\Wmat \in \R^{P \times K}$ added on the projection space, we initialize $\Wmat$ with $\Vmat$ (bias terms are initialized as zeros), as shown in Figure~\ref{fig:adaptation} (d). In this way, the visual and text heads are separated. Note that the language-initialized Two-Projection scheme is also basically equivalent to Figure~\ref{fig:adaptation} (b) in zero-shot settings. Please see Appendix Section \ref{sec:gap_clip_ce} for discussions.
$(ii)$ {\it One-Projection.}
To fairly compare with language-free model adaptation in Figure~\ref{fig:adaptation} (a), one linear projection head should directly be added on the backbone (before the visual projection) to ensure that the same number of trainable parameters are updated. Therefore, we propose to initialize  $\Wmat_m \in \R^{D \times K}$ in this case with the multiplication result of two linear matrices  $\Wmat_v\Vmat$, as shown in Figure~\ref{fig:adaptation} (e).  

\end{itemize}

\paragraph{Discussion.} We highly recommend the proposed language-initialized methods as the standard to adapt language-augmented visual models for two reasons: 
$(i)$ This simple method yields surprisingly superior empirical performance, as demonstrated in our experiments.
$(ii)$ It provides an effective mechanism to leverage the external knowledge that is collected for a downstream task in our benchmark. Specifically, the knowledge can be concatenated with the original language prompt (with a simple ``;'' in our experiments), then encoded into contextualized text features. When multiple knowledge items exist (\eg the case of GPT-3) for each concept, we concatenate one of its prompts and one of its knowledge items, and get the encoded text embedding of the concatenated sequence via the language encoder. This is performed for all the combinations between all prompts and knowledge items for this concept, then the averaged embedding is computed to represent the concept.  In contrast, random initialization would ignore this knowledge source. The language-initialized method can serve as a strong baseline to encourage  more effective knowledge-augmented adaptation methods. 

In OD, GLIP is a language-augmented detector, whose overall architecture can be simply considered as adding a cross-modal module over the CLIP-like dual-encoder. In GLIP~\cite{li2021grounded}, its linear probing has been implemented via updating $\Wmat_v$ and $\Wmat_t$. A prompt-tuning strategy was proposed, by initializing the language input of the cross-modal module as $\Vmat$, and only updating $\Vmat$ during adaptation. This is similar to our language-initialized strategy. 

\vspace{-2mm}
\section{Empirical Results and Findings}
\label{sec:experiments}
\vspace{-2mm}
We present the experimental results with our benchmark to illustrate two points. 
\textbf{\texttt{Q1}}: The importance of language in visual model transfer in the  adaptation stage.
\textbf{\texttt{Q2}}: We present three playgrounds that our benchmark can help to cultivate research in, including sample-efficiency, parameter-efficiency and external knowledge for visual transfer. We also present novel empirical findings.  
\subsection{The Role of Language for Vision}
\label{sec:comparison_foundation_models_main}
\vspace{-2mm}
\paragraph{Effectiveness of Language-initialized Adaptation Methods.}
In Table~\ref{tab:perf_ic_overall}, we compare the effectiveness of the proposed language-initialization methods with the checkpoint CLIP ViT-B32. 
The one-projection scheme is consistently better than two-projection scheme in all settings (though the gain is minor). This is because the former often has less parameters than the latter, as $D=768 > P=512$. 
To ensure fair comparisons with the random initialization of linear head in CLIP~\cite{radford2021learning} (\ie \# trainable parameters is the same), in the ensuing experiments, we consider the two-projection language-initialization scheme as the default, unless the one-projection scheme is specified.


\begin{table*}[t!]
    \centering
    \footnotesize
    \scalebox{0.85}{
    \begin{tabular}{@{}p{2.7cm}@{}| c l | cll}
    \toprule
    \multirow{2}{*}{Checkpoint }  & 
       \multicolumn{2}{c|}{ Settings }   &  \multicolumn{3}{c}{ 20 IC datasets }  \\
       \cmidrule{4-6} 
         & Adaptation & Initialization &   Zero-shot$^\dagger$ &  Few-shot (5, 20, 50) & Full  \\
         \midrule
        \multicolumn{5}{l}{\em {\bf Industry Track} (No pre-train data scale limit)} \\
        \multirow{5}{4.2cm}{CLIP \\ (ViT-B32)}
         &  LP & Random-2P    &  \multirow{5}{0.8cm}{56.64}  &   58.09 {\tiny $\pm$ 2.80}, 69.97 {\tiny $\pm$ 1.30}, 74.09 {\tiny $\pm$ 0.69} & 78.38 \\      
         &  LP & Language-2P  &   & 65.35 {\tiny $\pm$ 1.24}, 71.69 {\tiny $\pm$ 0.93}, 74.89 {\tiny $\pm$ 0.79} & 78.40 \\
         &  LP & Language-1P  &   &65.88 {\tiny $\pm$ 0.79}, 72.05 {\tiny $\pm$ 0.85}, 75.08 {\tiny $\pm$ 0.73} & 78.96 \\
         & FT & Random-2P        &    & 29.75 {\tiny $\pm$ 6.64}, 46.76 {\tiny $\pm$ 11.9}, 61.70 {\tiny $\pm$ 9.97} & 77.77 \\    
         & FT & Language-2P      &   & 63.29 {\tiny $\pm$ 3.18}, 72.19 {\tiny $\pm$ 1.31}, 75.70 {\tiny $\pm$ 1.14} & 80.35 \\  
         \midrule
        \multirow{2}{4.2cm}{Supervised \\(ViT-B32)}         
         &  LP & Random-1P    &  \multirow{2}{0.8cm}{ -}  & 56.00 {\tiny $\pm$ 2.67}, 67.23 {\tiny $\pm$ 1.66}, 71.35 {\tiny $\pm$ 1.17} & 75.29 \\
         &  FT & Random-1P  &   & 58.55 {\tiny $\pm$ 2.58}, 71.27  {\tiny $\pm$ 1.25}, 75.36 {\tiny $\pm$ 1.42} & 80.39 \\
         \midrule
         \multicolumn{5}{l}{\em {\bf Academic Track} (Pre-trained on large established public datasets)} \\
        \multirow{2}{4.2cm}{UniCL \\(Swin-Tiny)}         
         &  LP & Language-2P    &  \multirow{2}{0.8cm}{27.15}  & 54.31 {\tiny $\pm$ 4.15}, 66.42 {\tiny $\pm$ 2.08}, 70.49 {\tiny $\pm$ 1.01} & 74.75 \\      
         &  FT & Language-2P  &   & 44.75 {\tiny $\pm$ 5.42}, 56.53 {\tiny $\pm$ 5.37}, 67.90 {\tiny $\pm$ 5.31} & 78.48 \\
         \midrule
        \multirow{2}{4.2cm}{\textsc{K-Lite} \\(Swin-Tiny)}         
         &  LP & Language-2P    &  \multirow{2}{0.8cm}{33.44}  & 55.06 {\tiny $\pm$ 2.36}, 66.26 {\tiny $\pm$ 1.56}, 70.16 {\tiny $\pm$ 1.09} & 74.47 \\      
         &  FT & Language-2P  &   & 48.41 {\tiny $\pm$ 2.84}, 58.06 {\tiny $\pm$ 4.30}, 71.66 {\tiny $\pm$ 2.02} & 78.05 \\         
        \bottomrule
    \end{tabular}
    }
    \vspace{-0mm}
    \caption{Averaged results on 20 IC datasets using linear probing (LP) and fine-tuning (FT). 
    Random-1P, Random-2P, Language-1P and Language-2P indicates the initialization method in Figure~\ref{fig:adaptation} (a), (c), (e) and (d), respectively. $\dagger$ Note that one zero-shot result is reported for each model checkpoint using the method in  Figure~\ref{fig:adaptation} (b), which is independent from adaptation/initialization methods.}
    \label{tab:perf_ic_overall}
      \vspace{-3mm}
\end{table*}
\begin{table*}[t!]
    \centering
    \footnotesize
    \scalebox{0.85}{
    \begin{tabular}{@{}p{2.0cm}@{} l|  cll}
    \toprule
    \multirow{2}{*}{Checkpoint}& \multirow{2}{*}{Adaptation}
     &  \multicolumn{3}{c}{ 35 OD datasets }  \\
       \cmidrule{3-5} 
         &  &    Zero-shot &  Few-shot (1, 3, 5, 10) & Full  \\
         \midrule
        \multirow{3}{4.2cm}{GLIP \\(Swin-Tiny)}    
         &  Prompt   &   \multirow{3}{0.8cm}{ 19.7}  
         &  29.7{\tiny $\pm$ 0.4}, 36.5{\tiny $\pm$ 0.6}, 39.0{\tiny $\pm$ 1.1}, 41.8{\tiny $\pm$ 1.2} &  54.4 \\    
         &  LP  &    & 22.2{\tiny $\pm$ 0.1}, 24.4{\tiny $\pm$ 0.2}, 25.1{\tiny $\pm$ 0.2}, 25.6{\tiny $\pm$ 0.6} &  35.2  \\      
         &  FT    &   & 32.2{\tiny $\pm$ 0.7}, 39.2{\tiny $\pm$ 0.3}, 42.5{\tiny $\pm$ 0.9}, 49.1{\tiny $\pm$ 0.6}    & 63.2 \\
         \midrule
        \multirow{2}{4.2cm}{DyHead \\(Swin-Tiny)}         
         &  LP      &  \multirow{2}{0.8cm}{-}  
         & 15.2{\tiny $\pm$ 0.6}, 19.2{\tiny $\pm$ 0.9}, 19.8{\tiny $\pm$ 1.0}, 20.6{\tiny $\pm$ 1.1} &  31.4 \\      
         &  FT    &   
         & 25.6{\tiny $\pm$ 0.4}, 37.1{\tiny $\pm$ 0.5},  40.1{\tiny $\pm$ 1.5}, 44.6{\tiny $\pm$ 0.7}    & 63.9 \\ 
        \bottomrule
    \end{tabular}
    }
    \vspace{-2mm}
    \caption{Averaged results on 35 OD datasets. \vspace{-2mm}}
    \label{tab:perf_od_overall}
      \vspace{-2mm}
\end{table*}

\begin{wrapfigure}{r}{0.48\textwidth}
  \begin{center}
  \vspace{-5mm}
    \includegraphics[width=0.48\textwidth]{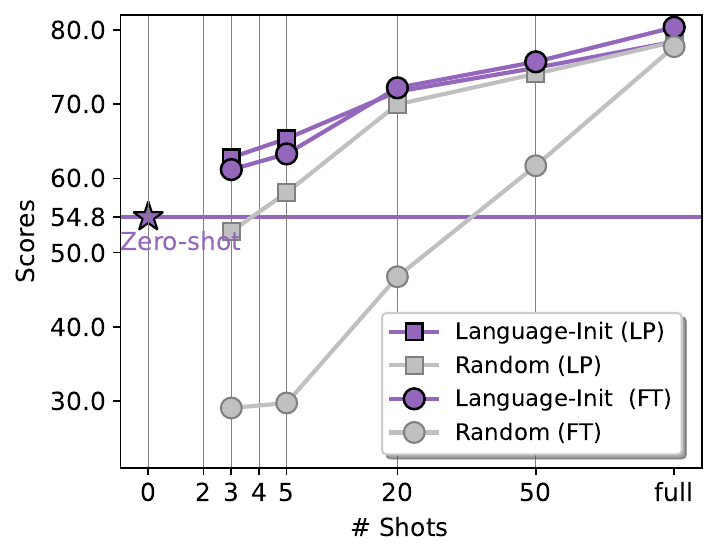}
  \end{center}
    \vspace{-5mm}
    \caption{Comparison of random- and language- initialized adaptation.
    }
    \vspace{-3mm}
    \label{fig:adaptation_compare_language_random}
\end{wrapfigure}

As shown in Fig.~\ref{fig:adaptation_compare_language_random}, under both linear probe (LP) and fine-tuning (FT) settings, language-based initialization significantly outperforms random initialization.  Notably, we show that even with very few shots (\eg 2-shots), both our LP and FT is able to outperform the zero-shot CLIP.  This is contradictory to the finding in the original CLIP paper~\cite{radford2021learning}, where zero-shot outperforms linear probing in the fewer shot (less than 4) settings. With the proposed language-init method, one can ensure that few-shot performance is always better than zero-shot, as we essentially reduce to zero-shot when zero iteration is updated in our language-init method.
Moreover, we also find that with random initialization, FT performs significantly worse than LP under few-shot settings. However, with language-init, FT starts to outperform LP with more than 20 shots.
Both findings demonstrate the proposed language-based initialization is consistently effective, suggesting that it is an important technique, and should be the standard adaptation method for language-augmented visual models like CLIP. Further, the correct adaptation methods for language-augmented visual models should leverage both the pre-trained visual and text encoder. It is not sufficient to solely transfer from the visual encoder, pre-trained language encoder plays an important role in task transfer.

\paragraph{The Competition of Pre-trained Models: Language-free vs Language-augmented.} 
We summarize the transfer performance of pre-trained models for IC in Table~\ref{tab:perf_ic_overall} and OD in Table~\ref{tab:perf_od_overall}. For IC, we also compare CLIP against other language-free visual models including MoCov3, MAE, ViT, DeiT in Appendix. We see that the language-augmented model (CLIP) outperforms language-free model (Supervised ViT) in the limited data settings. The gap is closed when more training examples are observed (\eg, 50-shot and full-shot). This is probably because the pre-training power is gradually dominated by larger-scale downstream training. Further, language-augmented models are able to perform zero-shot task transfer, while traditional language-free models cannot. Similar conclusions can be drawn for OD in Table~\ref{tab:perf_od_overall}. Hence, we recommend the use of language-augmented visual models for task-level transfer.

\begin{figure}[t!]
   \vspace{-2pt}
   \centering
    \scalebox{1.0}{
    \begin{tabular}{cc}
     \hspace{-3mm}
      \includegraphics[width=0.47\linewidth]{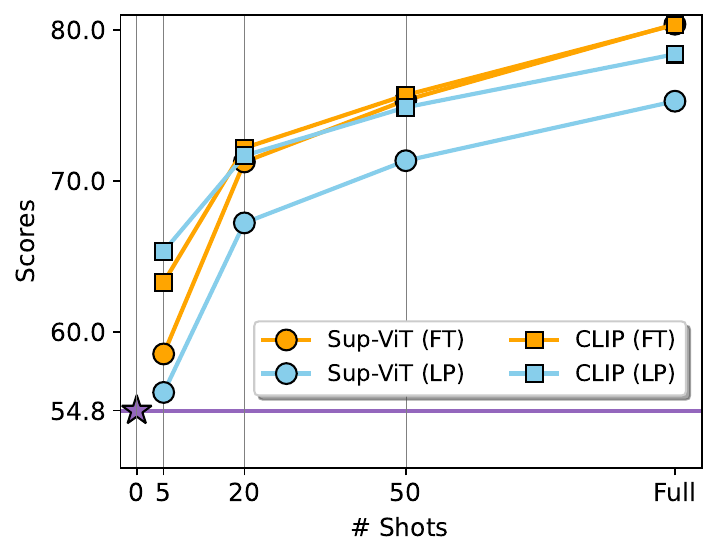}
        & 
      \hspace{-3mm}
      \includegraphics[width=0.47\linewidth]{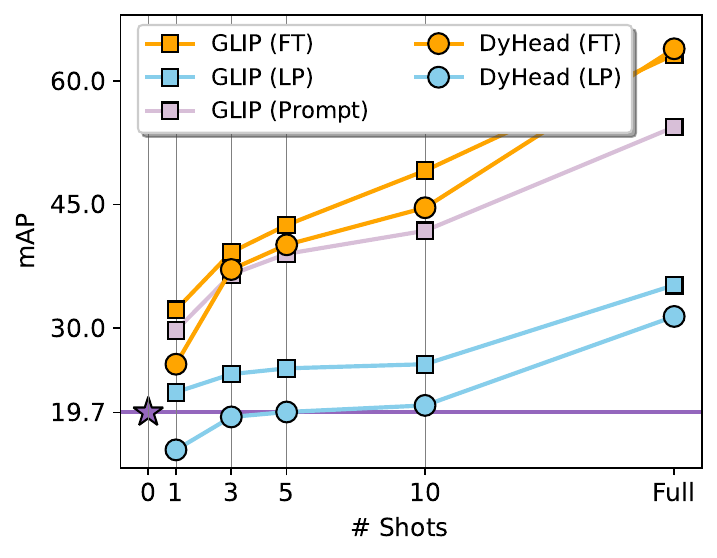}
      
      \vspace{-2mm}
       \\
      (a) IC: Sample Efficiency & 
      (b) OD: Sample Efficiency
      \vspace{4mm} \\
     \hspace{-2pt}
     
    \includegraphics[width=0.46\linewidth]{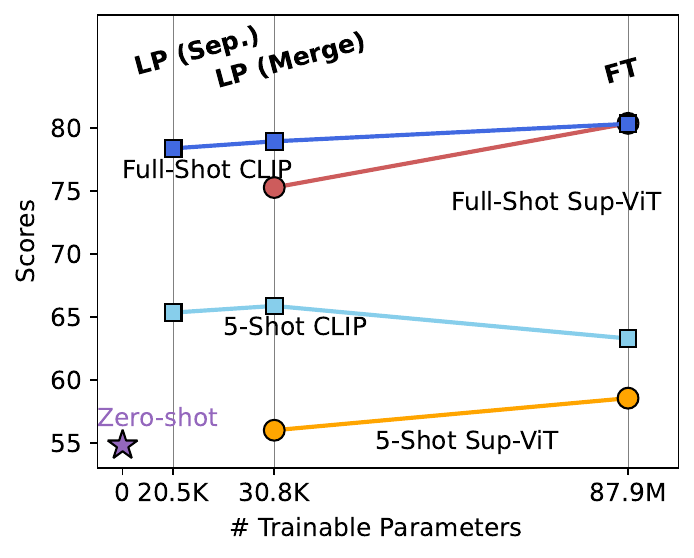}      
        & 
    \hspace{-3mm}
      \includegraphics[width=0.45\linewidth]{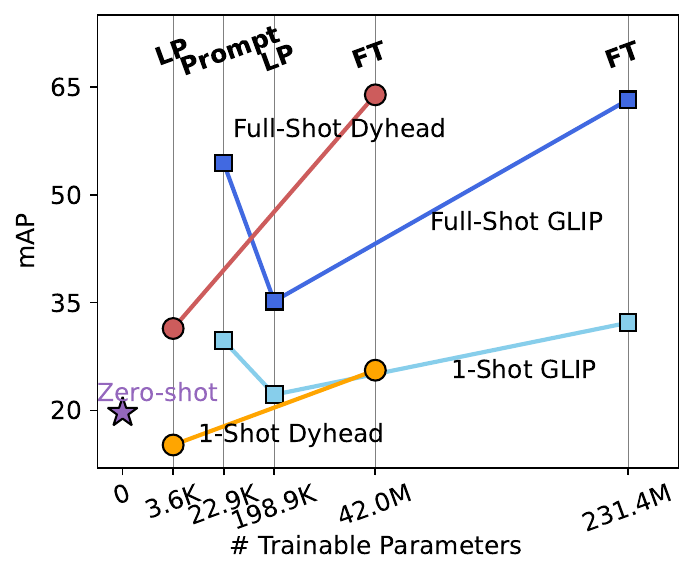}
       \vspace{-2mm}
       \\
      (c) IC: Parameter Efficiency  & 
      (d) OD: Parameter Efficiency
      \end{tabular}
    }
    \vspace{-0mm}
    \caption{Adaptation efficiency considerations. For IC, comparison of adaptation efficiency between CLIP and supervised ViT (Sup-ViT). For OD, comparison of adaptation efficiency between GLIP and DyHead.}
    \vspace{-5mm}
    \label{fig:adaptation_efficiency}
\end{figure}


\vspace{-1mm}
\subsection{Playground I: Sample Efficiency}
\vspace{-3mm}
We explore sample efficiency in Fig.~\ref{fig:adaptation_efficiency} (a) for IC.
First, we find that CLIP consistently outperforms supervised ViT (Sup-ViT), yielding a significant 5\textasciitilde10\% gain in the 5-shot settings. 
This suggests that CLIP is more sample-efficient than supervised ViT .
%
Furthermore, we find that fine-tuning CLIP yields better performance than linear probing in >20-shot settings, while being worse in the 5-shot setting.  
This is a bit surprising, as it is contradictory to the common convention that fine-tuning is always better than linear probing. We hypothesize this is because fine-tuning tends to over-fit in the scenarios with a large number of trainable parameters and a small number of training samples.
Overall, it suggests that fine-tuning CLIP potentially has a better sample efficiency than linear probing, and a better adaptation strategy on fewer-shot settings can be explored in the future. For supervised ViT, FT is always better than LP, the performance gap becomes larger when more samples are used. In 5-shot settings, the gap is minor, which is similar to observations made for supervised CNNs~\cite{yosinski2014transferable,karayev2013recognizing,karianakis2018reinforced}.
To compare pre-trained models, we suggest to report the evaluation results on the entire spectrum of sample-efficiency to fully study the behaviors of a pre-trained model. If compute resource is limited, zero-shot or few-shot evaluation can be used as a quick assessment.

We explore sample efficiency for OD in Fig.~\ref{fig:adaptation_efficiency} (b). The conclusion is similar to IC in that the language-augmented visual model (GLIP) is more sample-efficient than the language-free visual model (DyHead), when the models are adapted using either LP or FT settings. The performance gap is large in the fewer-shot settings and is small in the full-shot settings. The difference is that FT consistently outperforms LP in all settings for OD.  This is probably because there are a lot of boxes (training instances) per image in OD, which makes OD less likely to over-fit compared to IC.

\vspace{-1mm}
\subsection{Playground II: Parameter Efficiency}
\vspace{-3mm}
We study the parameter efficiency in Fig.~\ref{fig:adaptation_efficiency} (c) for IC.
For CLIP, we experiment with two different settings of linear probing on whether to merge the last two linear projection layers ($\mathbf{W}_v$ and $\mathbf{V}$ in Fig.~\ref{fig:adaptation}).  Merging these two layers in CLIP allows 1.5$\times$ trainable parameters in the linear probe classifier as keeping them separated. 
First, it shows a trend that a larger number of trainable parameters leads to better performance, as demonstrated by three curves/scenarios: full-shot CLIP, full-shot Sup-ViT and 5-shot Sup-ViT. This also verifies that LP and FT provide the lower bound and upper bound, respectively, in terms of both \#parameter and performance. Most existing parameter-efficient adaptation methods play a trade-off game.
However, in the scenario of 5-shot CLIP, we do notice a slight drop in performance when we further increase the number of trainable parameters to full-model fine-tuning. It suggests that the scenario of adapting language-augmented visual models for data-limited settings is a more meaningful playground to explore the line of research in parameter-efficient adaptation methods, as the best performance may require an optimal number of trainable parameters, which has been less explored.

We study the parameter efficiency in Fig.~\ref{fig:adaptation_efficiency} (d) for OD. The overall trend is similar in that better performance comes with more parameters. It turns out that prompt tuning an language-augmented OD model is an effective parameter-efficient approach. For example, prompting is better than linear probing in GLIP. Further, prompt tuning GLIP outperforms fine-tuning DyHead in the 1-shot setting, where the former has less than 0.1\% parameters of the latter. 

\vspace{-1mm}
\subsection{Playground III: The Benefit of External Knowledge for Vision}
\label{sec:benefit_knowledge_main}
\begin{figure}[t!]
	\centering
	\begin{tabular}{cc}
	\hspace{-4pt}
	\includegraphics[width=0.44\linewidth]{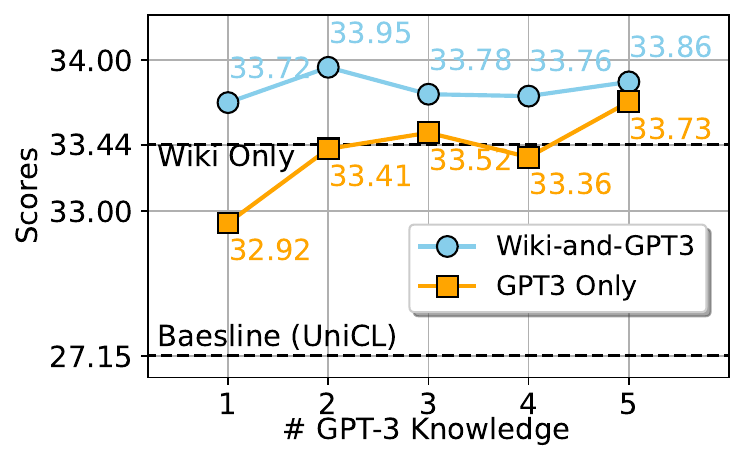}
	& 
	\hspace{-4pt}
	\includegraphics[width=0.44\linewidth]{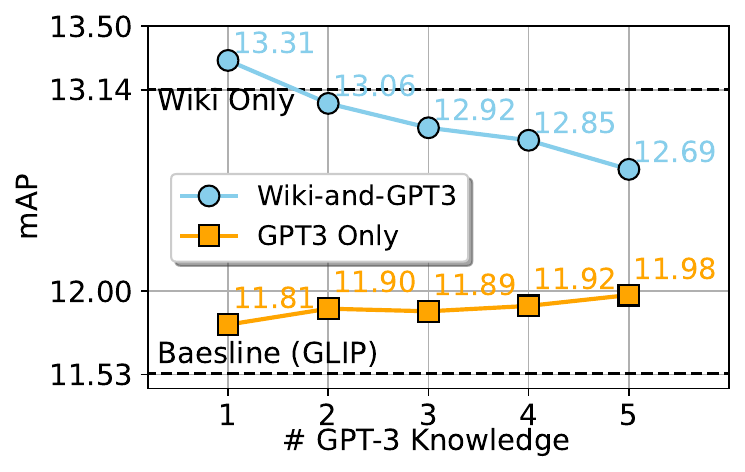} \vspace{-4pt}	\\
	(a) Image classification &
	(b) Object detection
	\end{tabular} \\
	\vspace{3pt}
    \vspace{4pt}
    \hspace{2mm}
    \scalebox{0.9}{
    \begin{tabular}{l | l c | ccccc}
    \toprule
          \multirow{2}{*}{Problem} 
          & \multirow{2}{*}{Model}
          & \multirow{2}{*}{Baseline}
          & 
         \multicolumn{5}{c}{External Knowledge Sources in Evaluation} \\
           &  &  & 
         $\texttt{wn\_path}$ &  
         $\texttt{wn\_def}$  &  
         $\texttt{wiki\_def}$  &
         $\texttt{gpt3}$  &
         $\texttt{wiki\_def}$ \& $\texttt{gpt3}$
          \\          
        \midrule
        IC &  UniCL
        & \cellcolor{orange!30}27.15 
        & \cellcolor{emerald!30}30.68
        & \cellcolor{emerald!30}29.92
        & \cellcolor{emerald!30}33.44          
        & \cellcolor{emerald!30}33.73         
        & \cellcolor{emerald!30}33.95 \\ 
        OD &  GLIP-A
        & \cellcolor{orange!30}11.53 
        & \cellcolor{emerald!30}12.43
        & \cellcolor{emerald!30}11.70
        & \cellcolor{emerald!30}13.14          
        & \cellcolor{emerald!30}11.98        
        & \cellcolor{emerald!30}13.30 \\         
        \bottomrule
    \end{tabular}
    }
    \vspace{-2pt}
    \caption{Zero-shot task transfer with various external knowledge sources in the evaluation stage. In (a) and (b), a varying number of generated GPT-3 knowledge sequences is utilized for inference, and ``Wiki-and-GPT3''  indicates both Wiktionary and GPT3 knowledge are used simultaneously. The bottom table summarizes the prediction result for each knowledge source.}
    \label{fig:gpt3}
    \vspace{-3mm}
\end{figure}

\vspace{-2mm}
We investigate the effectiveness of external knowledge in Fig.~\ref{fig:gpt3}, measured by zero-shot task transfer performance. The model \textsc{K-Lite} is evaluated, as its pre-training is knowledge-augmented. We find that leveraging external knowledge improves upon the knowledge-free pre-training counterparts (UniCL and GLIP). For example, UniCL is improved from 27.15 to 29.92\textasciitilde33.93, and GLIP-A is improved from 11.53 to 11.70\textasciitilde13.30. Further, for GPT3 knowledge, a larger number of generated knowledge items often leads to higher performance. When combining GPT3 knowledge with Wiktionary knowledge, we see a further performance boost. With an increasing number of GPT3 knowledge items, the gain is consistently improved for IC, but not for OD.
In Table~\ref{tab:perf_ic_overall}, we study the role of knowledge for task transfer in model adaptation. Initializing the linear head using features encoded with knowledge is an effective way to leverage the collected knowledge sources, especially for the fewer-shot settings. 

One may wonder if the collected knowledge in \shortname{} benchmark can also benefit knowledge-free pre-trained models such as CLIP during model adaptation? We confirm its effectiveness in Appendix. In zero-shot transfer, external knowledge improves the baseline on four datasets. In few- and full-shot transfer,  one may selectively choose whether to update the model using external knowledge, by observing the best performance in the auto-tuning stage. This selective strategy with the availability of external knowledge demonstrates a consistent improvement (or tie) on 15+ out of 20 datasets.

\vspace{-1mm}
\section{Conclusions}
\vspace{-3mm}
We have presented \shortname{}, a platform to evaluate the recently emerging language-augmented visual models for task-level transfer. It consists of 20 image classification datasets and 35 object detection datasets. All of them are collected from public domains, and are enriched with various external knowledge sources to enhance the language modality. We have developed open-source toolkits with an auto hyper-parameter pipeline and novel language-initialized adaptation methods to ensure easy utilization and fair comparisons. Strong baseline results are produced from the toolkit to cultivate research in a variety of topics, \eg more transferable language-augmented visual models, advanced model adaption methods (sample-efficiency and parameter-efficiency), and external knowledge for task-level transfer. The question of how to design general-purpose task-level transferable visual models remains largely unanswered. Given benchmarks and tookits we have developed from the perspective of language-augmented visual models, we believe that \shortname{} can provide fertile soil for addressing this challenge.

\section*{Acknowledgments}
The authors gratefully acknowledge Haotian Zhang for building the ODinW leaderboard on Eval AI, Pengcheng He for helpful discussions to have a separate track dedicated for users from academia, Baolin Peng and and Zhengyuan Yang for the inspirations of GPT3 to generate knowledge for dialogue and OK-VQA tasks,  Bo Li for insights on the topic of domain generalization, Zhuowen Tu for the inspirations to make benchmark scope wider to measure all pre-trained vision models, Ce Liu for suggestions to compare the benchmark with well-established vision datasets such as ImageNet and COCO.
The benchmark depends on publicly available datasets; we acknowledge all the original authors who made their datasets public. Please follow the original license of each dataset and keep this benchmark for academic purposes.  This work was supported in part by NSF CAREER IIS-2150012, the Wisconsin Alumni Research Foundation, and Institute of Information \& communications Technology Planning \& Evaluation(IITP) grants funded by the Korea government(MSIT) (No. 2022-0-00871, Development of AI Autonomy and Knowledge Enhancement for AI Agent Collaboration) and (No. RS-2022-00187238, Development of Large Korean Language Model Technology for Efficient Pre-training).



{\small
\bibliography{egbib}
\bibliographystyle{plain}
}

\appendix
\newpage
\clearpage
\newpage

\noindent\makebox[\linewidth]{\rule{\linewidth}{3.5pt}}
\begin{center}
	\bf{\Large Supplementary Material for 
``\shortname{}: A Benchmark and Toolkit for Evaluating
Language-Augmented Visual Models''}
\end{center}
\noindent\makebox[\linewidth]{\rule{\linewidth}{1pt}}

This appendix is organized as follows. 

\begin{itemize} 

\item In Section \ref{sec:societal} (referred by CheckList), we discuss the societal impact.

\item In Section \ref{sec:related_work_nlp} (referred by Section~\ref{sec:related_work}), we discuss the related work in pre-trained language models in NLP.

\item In Section \ref{sec:appendix_benchmark} (referred by Section~\ref{sec:benchmark_dataset}), we summarize the datasets statistics and license  used in our benchmark suite. We also describe how to obtain external knowledge from GPT-3, and construct language prompts.


\item In Section \ref{sec:auto_tuning} (referred by Section~\ref{sec:toolkit}), we introduce more details of our toolkits, including the automatic hyper-parameter tuning pipeline and implementation details.

\item In Section \ref{sec:gap_clip_ce} (referred by Section~\ref{sec:toolkit}), we discuss the gap between language-image model pre-training and adaptation.

\item In Section \ref{sec:comparisons_foundation_models} (referred by Section~\ref{sec:comparison_foundation_models_main}), we provide performance of comparison different vision pre-trained models.

\item In Section \ref{sec:benefit_knowledge}  (referred by Section~\ref{sec:benefit_knowledge_main}), we provide empirical evidence that external knowledge improves CLIP adaptation.
\end{itemize}

\section{Societal Impact}\label{sec:societal} 
\vspace{-2mm}
We do not anticipate a specific negative impact, but, as with any Machine Learning method, we recommend to exercise caution. The existing knowledge bases such as Word-Net and Wiktionary are the results of crowd-sourcing various human knowledge or commonsense into a centered place. \shortname{} provides evidence to leverage such knowledge bases for AI research. It encourages the community to contribute more to improve the coverage and quality of knowledge items, which will further benefit AI research. We also leverage GPT3 to generate knowledge, which is stored as a part of benchmark for public academic use. The related societal impact on the usage of AI-generated content may apply to our work.

\section{Our Position}

\subsection{Computer Vision in the Wild}
\label{sec:appendix_cvinw}
In this paper, we advocate our perspective on ``{\bf Computer Vision in the Wild (CVinW)}'', whose ultimate goal is to develop a transferable foundation model/system that can {\it effortlessly} adapt to {\it a large range of visual tasks in the wild}. We further illustrate two key factors as follows. 

\paragraph{Factor I: The Task Transfer Cost is Low.}
One major advantage of pre-trained/foundation models is the promise that they can transfer to downstream tasks {\it effortlessly} (or in an inexpensive manner). It means that model adaptation efficiency is an important factor to measure the performance of the pre-trained models. To concretely illustrate the notion of inexpensive adaptation, we provide a 2D chart on the model adaptation cost in Figure~\ref{fig:adapation_cost}. The cost is considered in two orthogonal dimensions: sample-efficiency and parameter-efficiency.
One may interpolate and  make combinations in the 2D space, to get different model adaptation methods with different cost. This is design philosophy behind our comprehensive evaluation metrics. Two playgrounds with different efficiency considerations presented in the main paper are simplified settings to study model performance. As a north star, one foundation could with fixed weights should zero-shot transfer well on many downstream tasks, the most inexpensive regime in the bottom-left corner of Figure~\ref{fig:adapation_cost}.

\paragraph{Factor II: The Task Transfer Scenarios are Broad.}
We illustrate and compare the settings of CVinW using a 2D chart in Figure~\ref{fig:setting_cvinw}. It consists of two dimensions: the input visual content  and output concept prediction.  For the example provided in the standard setting, the natural image with concept ``person, sheep, dog'' is presented. 
We divide the the 2D chart into four quadrants 

\begin{enumerate}
 [leftmargin=5.5mm]

\item {\bf The Standard Close-Set Setting.} The bottom-left quadrant is the standard setting, where most existing visual recognition lie in, training and evaluation are consistent in both their visual input distributions and output category sets. For example, only natural images with concept ``person, sheep, dog'' are presented in training and evaluation.

\item {\bf Open-Set/Vocabulary/World Setting.} 
In the top-left quadrant, the recognition of new concepts is enabled, while the visual input distributions of training and evaluation are in the same domain. This research problem is usually tackled by traditional class-level zero-shot transfer, or some experimental settings in the open-set recognition. For example, natural images with concepts ``person, sheep, dog'' are presented in training, but natural images with concepts ``border collie, running, while shirt'' are presented in evaluation. Though the testing concepts are closely to training concepts, but they have not been observed by the models in training.

\item {\bf The Domain Shift Setting.} 
In the bottom-right quadrant, the input image distributions are shifted between training and evaluation sets, while the output category sets are the same. This research problem is often tackled in the area of domain adaptation and out-of-distribution. For example, natural images with concepts ``person, sheep, dog'' are presented in training, but thermal images are presented in evaluation, though the concepts have been observed in training.

\item {\bf Computer Vision in the Wild Setting.} 
In the top-right quadrant, the strong generalization ability to both new concepts and new visual distributions is required. Therefore, the model can perform well on new tasks of any customized set of concepts in any visual domains. This is a setting we advocate for computer vision in the wild, where any new downstream tasks can appear in this quadrant, and it requires models with a strong task-level visual transfer ability.

\end{enumerate}

For the readers who are interested in the literature on Computer Vision in the Wild, we create an up-to-date CVinW reading list at \url{https://github.com/Computer-Vision-in-the-Wild/CVinW_Readings}.

\subsection{Related Works in NLP: Benchmarks, Adaptation, and Knowledge} 
\label{sec:related_work_nlp}
\vspace{-2mm}

With a focused scope, our benchmark evaluates language-image models on two core CV problems: IC and OD.
Though language-image models can also be deployed and evaluated in other scenarios, including joint visual-text evaluation~\cite{zhou2022vlue,bugliarello2022iglue}(\eg visual question answering~\citep{antol2015vqa,marino2019ok}, video-and-language understanding~\citep{li2021value}) and the scenario of improving language encoders with vision~\citep{tan2020vokenization}. Our benchmark is complementary to them in its focus on evaluating vision encoders.


Our work takes major inspiration from the development of pre-trained language models in natural language processing (NLP) in several aspects:
$(i)$
{\it Benchmarks}. Platforms with a suite of small datasets such as GLUE~\cite{wang2018glue}/SuperGLUE~\cite{wang2019superglue} have been extensively used to evaluate the general language understanding ability of pre-trained models~\cite{devlin2018bert}. Recently, there is a trend in NLP to develop task-agnostic models such as the GPT family~\cite{brown2020language} that demonstrate task-level transfer learning ability, enabling zero-shot and few-shot transfer to downstream datasets. The success in NLP encourages us to build a generic benchmark to measure the similar transferability for visual models.
$(ii)$
{\it Efficient adaptation}. The democratization of large pre-trained models for efficient adaptation in downstream applications is an important topic in practice. Many algorithms have been developed for various efficiency considerations, including adapters~\cite{houlsby2019parameter} and prompt tuning~\cite{li2021prefix,liu2021pre}. 
In particular, natural language prompting is the method of reformatting NLP tasks in the format of a natural language response to natural language input,  has attracted attentions in zero-shot and few-shot learning in NLP~\citep{sanh2021multitask}.
It has inspired a few recent works for language-augmented visual models~\citep{zhou2022conditional,sung2021vl,yao2021cpt,gao2021clip}. Our benchmark can serve as a comprehensive playground to quantify the progress in the emerging field of visual model adaptation. We also propose to use external knowledge for prompt engineering, and a novel language/knowledge-initialized model adaptation method as a strong baseline. 
$(iii)$
{\it Knowledge}. Knowledge-intensive tasks~\cite{marino2019ok,petroni2020kilt} — those where a human can only be expected to perform the task with access to a knowledge source such as Wikipedia — are challenging for even cutting edge NLP and vision-and-language models, as it is infeasible to train large models to memorize everything. KILT~\cite{petroni2020kilt} is a benchmark that contains a suite of tasks/datasets for evaluating and analyzing knowledge-intensive NLP models. Similarly, we also add various external knowledge sources in each downstream dataset for our vision benchmark.

\section{Benchmark Suite}
\label{sec:appendix_benchmark}

\subsection{Detailed Dataset Statistics}
\label{sec:appendix_dataset_statistics}
In Table~\ref{table:downstream_ic_dataset} and Table~\ref{table:downstream_od_dataset}, we list the basic statistics of 20 image classification datasets and 35 object detection datasets in the benchmark.

The benchmark may inherit data biases from the public datasets we have considered, both in the images and the annotations. Such biases might be reflected in the predictions of the systems trained on these data. Users should not completely rely on such systems for making real-world decisions.

\label{app:ic}

\begin{table*}[t!]
  \centering
    \setlength{\tabcolsep}{2.2pt}
  \scalebox{0.76}{
\begin{tabular}{c | c c c c c c } 
 \toprule
 Dataset & \#Concepts & Train size & Test size & Evaluation metric & Source link  \\ 
 \midrule

Hateful Memes~\cite{kiela2020hateful} & 2 & 8,500 & 500 & ROC AUC & \href{https://ai.facebook.com/blog/hateful-memes-challenge-and-data-set/}{Facebook} \\
PatchCamelyon~\cite{veeling2018rotation}  & 2 & 262,144 & 32,768 & Accuracy & \href{https://www.tensorflow.org/datasets/catalog/patch_camelyon}{Tensorflow} \\
Rendered-SST2~\cite{radford2021learning} & 2 & 6,920 & 1,821 & Accuracy & \href{https://github.com/openai/CLIP/blob/main/data/rendered-sst2.md}{OpenAI} \\
KITTI Distance~\cite{fritsch2013new} & 4 & 6,347 & 711 & Accuracy & \href{http://www.cvlibs.net/datasets/kitti/}{KITTI website} \\
FER 2013~\cite{fer2013} & 7 & 28,709 & 3,589 & Accuracy & \href{https://www.kaggle.com/c/challenges-in-representation-learning-facial-expression-recognition-challenge/data}{Kaggle fer2013} \\
CIFAR-10~\cite{krizhevsky2009learning} & 10 & 50,000 & 10,000 & Accuracy & \href{https://www.tensorflow.org/datasets/catalog/cifar10}{Tensorflow} \\
EuroSAT~\cite{helber2019eurosat} & 10 & 5,000 & 5,000 & Accuracy & \href{https://www.tensorflow.org/datasets/catalog/eurosat}{Tensorflow} \\
MNIST~\cite{deng2012mnist} & 10 & 60,000 & 10,000 & Accuracy & \href{https://www.tensorflow.org/datasets/catalog/mnist}{Tensorflow} \\
VOC 2007 Classification~\cite{everingham2010pascal} & 20 & 2,501 & 4,952 & 11-point mAP & \href{http://host.robots.ox.ac.uk/pascal/VOC/voc2007/index.html}{VOC 2007} \\
Oxford-IIIT Pets~\cite{parkhi2012cats} & 37 & 3,680 & 3,669 & Mean-per-class & \href{https://www.tensorflow.org/datasets/catalog/oxford_iiit_pet}{Tensorflow} \\
GTSRB~\cite{stallkamp2011german} & 43 & 26,640 & 12,630 & Accuracy & \href{https://benchmark.ini.rub.de/gtsrb_dataset.html}{GTSRB website} \\
Resisc-45~\cite{cheng2017remote} & 45 & 3,150 & 25,200 & Accuracy & \href{https://www.tensorflow.org/datasets/catalog/resisc45}{Tensorflow} \\
Describable Textures~\cite{cimpoi2014describing} & 47 & 1,880 & 1,880 & Accuracy & \href{https://www.tensorflow.org/datasets/catalog/dtd}{Tensorflow} \\
CIFAR-100~\cite{krizhevsky2009learning} & 100 & 50,000 & 10,000 & Accuracy & \href{https://www.tensorflow.org/datasets/catalog/cifar100}{Tensorflow} \\
FGVC Aircraft (variants)~\cite{maji2013fine} & 100 & 3,334 & 3,333 & Mean-per-class & \href{https://www.robots.ox.ac.uk/~vgg/data/fgvc-aircraft/}{FGVC website} \\
Food-101~\cite{bossard2014food} & 101 & 75,750 & 25,250 & Accuracy & \href{https://www.tensorflow.org/datasets/catalog/food101}{Tensorflow} \\
Caltech-101~\cite{fei2004learning} & 102 & 3,060 & 6,084 & Mean-per-class & \href{https://www.tensorflow.org/datasets/catalog/caltech101}{Tensorflow} \\
Oxford Flowers 102~\cite{nilsback2008automated} & 102 & 1,020 & 6,149 & Mean-per-class & \href{https://www.tensorflow.org/datasets/catalog/oxford_flowers102}{Tensorflow} \\
Stanford Cars~\cite{krause20133d} & 196 & 8,144 & 8,041 & Accuracy & \href{https://www.tensorflow.org/datasets/catalog/cars196}{Tensorflow} \\
Country-211~\cite{radford2021learning} & 211 & 31,650 & 21,100 & Accuracy & \href{https://github.com/openai/CLIP/blob/main/data/country211.md}{OpenAI} \\

\midrule
Total & 1151 & 638429 & 192677 & -- & -- &    \\ 

\bottomrule
\end{tabular}
}
\vspace{-2mm}
\caption{Statistics of 20 datasets used in image classification.}
\label{table:downstream_ic_dataset}
\end{table*}

\begin{table*}[t!]
  \centering
    \setlength{\tabcolsep}{2.2pt}
  \scalebox{0.76}{
\begin{tabular}{c | c | c c | c c | c } 
 \toprule
  \multirow{2}{*}{Dataset}  & \multirow{2}{*}{\#Concepts}  &  
  \multicolumn{2}{c|}{\#Image} & \multicolumn{2}{c|}{\#Annotated Regions}
  &   \multirow{2}{*}{Source link }   \\ 
  &   & Train & Test & Train & Test &  \\ 
 \midrule

CottontailRabbits   &     1  &    1980 &   10 &   2070 &     11 &  \href{https://public.roboflow.com/object-detection/cottontail-rabbits-video-dataset}{Roboflow}   \\
EgoHands(generic)~\cite{egohands2015iccv}  &     1  &    3840 &   480 &   12015 &     1514 & \href{https://public.roboflow.com/object-detection/hands}{Roboflow}   \\
MountainDewCommercial   &     1  &    17 &   1 &   453 &     32 & \href{https://public.roboflow.com/object-detection/mountain-dew-commercial}{Roboflow}   \\
Packages   &     1  &    19 &   3 &   31 &     5 & \href{https://public.roboflow.com/object-detection/packages-dataset}{Roboflow}\\
Raccoon   &     1  &    150 &   17 &   164 &     20 & \href{https://public.roboflow.com/object-detection/raccoon}{Roboflow}   \\
WildfireSmoke   &     1  &    516 &   74 &   516 &     74 & \href{https://public.roboflow.com/object-detection/wildfire-smoke}{Roboflow} \\
Pistols   &     1  &    2377 &   297 &   2728 &     358 & \href{https://public.roboflow.com/object-detection/pistols}{Roboflow}   \\
Pothole   &     1  &    465 &   67 &   1256 &     154 & \href{https://public.roboflow.com/object-detection/pothole}{Roboflow}   \\
MaskWearing   &     2  &    105 &   15 &   696 &     96 & \href{https://public.roboflow.com/object-detection/mask-wearing}{Roboflow}   \\
NorthAmericaMushrooms   &     2  &    41 &   5 &   67 &     9 & \href{https://public.roboflow.com/object-detection/na-mushrooms}{Roboflow}   \\
OxfordPets(species)~\cite{parkhi2012cats}   &     2  &    2523 &   358 &   2527 &     358 & \href{https://public.roboflow.com/object-detection/oxford-pets/2}{Roboflow}   \\
PKLot640   &     2  &    8691 &   1242 &   497856 &     70684 &  \href{https://public.roboflow.com/object-detection/pklot}{Roboflow}  \\
ThermalCheetah   &     2  &    90 &   14 &   152 &     31 & \href{https://public.roboflow.com/object-detection/thermal-cheetah}{Roboflow}   \\
ThermalDogsAndPeople   &     2  &    142 &   20 &   181 &     27 & \href{https://public.roboflow.com/object-detection/thermal-dogs-and-people}{Roboflow}   \\
BCCD   &     3  &    255 &   36 &   3450 &     471 & \href{https://public.roboflow.com/object-detection/bccd}{Roboflow}   \\
HardHatWorkers   &     3  &    5069 &   1766 &   19455 &     6808 & \href{https://public.roboflow.com/object-detection/hard-hat-workers}{Roboflow}   \\
ShellfishOpenImages   &     3  &    407 &   58 &   859 &     116 & \href{https://public.roboflow.com/object-detection/shellfish-openimages}{Roboflow}   \\
EgoHands(specific)   &     4  &    3840 &   480 &   12015 &     1514 & \href{https://public.roboflow.com/object-detection/hands/1}{Roboflow}   \\
AerialMaritimeDrone(large)   &     5  &    52 &   7 &   873 &     78 &  \href{https://public.roboflow.com/object-detection/aerial-maritime/1}{Roboflow}  \\
AerialMaritimeDrone(tiled)   &     5  &    371 &   32 &   1237 &     98 & \href{https://public.roboflow.com/object-detection/aerial-maritime/9}{Roboflow}   \\
VehiclesOpenImages   &     5  &    878 &   126 &   1676 &     258 & \href{https://public.roboflow.com/object-detection/vehicles-openimages}{Roboflow}   \\
BrackishUnderwater~\cite{pedersen2019brackish}  &     6  &    11739 &   1468 &   28518 &     3466 & \href{https://public.roboflow.com/object-detection/brackish-underwater}{Roboflow}   \\
Dice   &     6  &    576 &   71 &   1439 &     225 &  \href{https://public.roboflow.com/object-detection/dice}{Roboflow}  \\
Aquarium   &     7  &    448 &   63 &   3324 &     584 & \href{https://public.roboflow.com/object-detection/aquarium}{Roboflow}   \\
DroneControl   &     8  &    32688 &   4675 &   32734 &     4694 & \href{https://public.roboflow.com/object-detection/drone-gesture-control}{Roboflow}   \\
WebsiteScreenshots   &     8  &    1688 &   242 &   76820 &     10656 & \href{https://public.roboflow.com/object-detection/website-screenshots}{Roboflow}   \\
SelfDrivingCar   &     11  &    24000 &   3000 &   156730 &     19598 & \href{https://public.roboflow.com/object-detection/self-driving-car}{Roboflow}   \\
ChessPieces   &     13  &    202 &   29 &   2108 &     376 & \href{https://public.roboflow.com/object-detection/chess-full}{Roboflow}   \\
UnoCards   &     15  &    6295 &   899 &   18885 &     2697 &  \href{https://public.roboflow.com/object-detection/uno-cards}{Roboflow}  \\
PascalVOC~\cite{everingham2010pascal}   &     20  &    13690 &   3422 &   31356 &     7835 &  \href{https://public.roboflow.com/object-detection/pascal-voc-2012}{Roboflow}  \\
AmericanSignLanguageLetters   &     26  &    1512 &   72 &   1512 &     72 & \href{https://public.roboflow.com/object-detection/american-sign-language-letters}{Roboflow}   \\
Plantdoc~\cite{singh2019plantdoc}    &     30  &    2128 &   239 &   7629 &     454 &  \href{https://public.roboflow.com/object-detection/plantdoc}{Roboflow}  \\
BoggleBoards   &     36  &    285 &   35 &   5727 &     647 & \href{https://public.roboflow.com/object-detection/boggle-boards}{Roboflow}   \\
OxfordPets(breed)   &     37  &    2437 &   345 &   2441 &     345 &  \href{https://public.roboflow.com/object-detection/oxford-pets/1}{Roboflow}  \\
OpenPoetryVision   &     43  &    2798 &   402 &   8392 &     1198 & \href{https://public.roboflow.com/object-detection/open-poetry-vision}{Roboflow}   \\ 
\midrule
Total   &     314  &    132314 &   20070 &   937892 &     135563 &  --  \\ 
\bottomrule
\end{tabular}
}
\vspace{-2mm}
\caption{Statistics of 35 datasets used in object detection. Box mAP is used as the evaluation metric. Datasets are downloaded from \href{https://roboflow.com/}{Roboflow}. For the datasets without a citation, we refer to Roboflow links for the original sources.}
\label{table:downstream_od_dataset}
\end{table*}










\subsection{Visualization Comparison with Established Vision Datasets}
\label{sec:visualization_comparison}

\begin{figure}[t!]
   \vspace{-2pt}
   \centering
    \scalebox{1.0}{
    \begin{tabular}{cc}
     \hspace{-3mm}
      \includegraphics[width=0.48\linewidth]{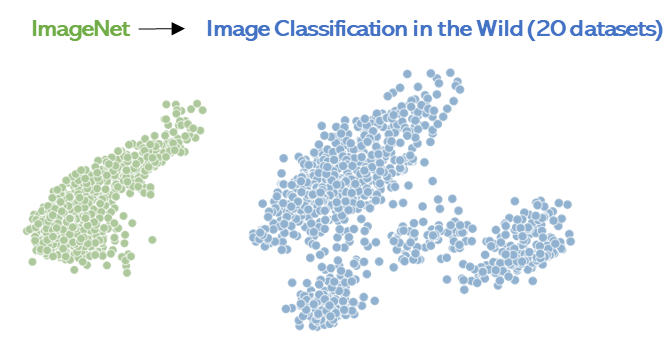}
        & 
      \hspace{-3mm}
      \includegraphics[width=0.48\linewidth]{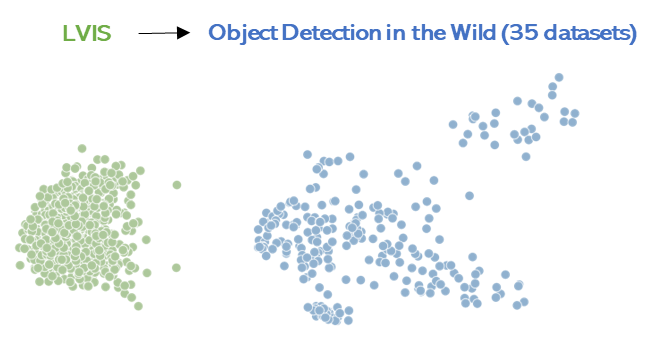}
      
      \vspace{-0mm}
       \\
      (a) Image Classification & 
      (b) Object Detection
      \end{tabular}
    }
    \vspace{-0mm}
    \caption{Semantic space comparison with 2D PCA. For IC or OD, the CLIP text feature of category names in each benchmark are projected together with PCA, and visualized separately.}
    \vspace{-0mm}
    \label{fig:visualization_comparison}
\end{figure}

We also compare our benchmark with well established datasets in computer vision: ImageNet-1K for IC and COCO/LVIS for OD. Note that LVIS is much diverse than COCO in terms of concept coverage. The visualization of concept semantic space is Figure~\ref{fig:visualization_comparison}. The semantics is computed by extracting the CLIP text features from the category names.
To quantitatively measure the diversity of different benchmarks, we compute the standard derivation (STD) over text features. The STD of ImageNet1-K and ICinW is 0.610 and 0.680, respectively.  The STD of LVIS and ODinW is 0.533 and 0.619, respectively.

\subsection{License}
\label{sec:licenses}

As per the original authors, the licenses of each dataset include 
CC BY-NC-SA 3.0\footnote{\url{https://creativecommons.org/licenses/by-nc-sa/3.0/}}, 
CC BY-NC-SA 4.0\footnote{\url{https://creativecommons.org/licenses/by-nc-sa/4.0/}},
CC BY 4.0\footnote{\url{https://creativecommons.org/licenses/by/4.0/}}, 
ODbL v1.0\footnote{\url{https://opendatacommons.org/licenses/odbl/1-0/}},
MIT\footnote{\url{https://choosealicense.com/licenses/mit/}}, 
CC0 1.0\footnote{\url{https://creativecommons.org/publicdomain/zero/1.0/}}.
Some datasets have published dedicated usage aggrements: Hateful Memes\footnote{\url{https://www.drivendata.org/competitions/64/hateful-memes/page/214/}}.  All datsets allow the usage for research purposes.
The images used in the datasets are from Internet, on non-offensive topics. The annotations in the datasets do not contain personally identifiable information.  

For external knowledge collected on \shortname{}, we suggest the users to follow the corresponding licenses: WordNet\footnote{\url{https://wordnet.princeton.edu/license-and-commercial-use}}, Wiktionary\footnote{\url{https://en.wiktionary.org/wiki/Wiktionary:Main_Page}}, GPT-3\footnote{\url{https://openai.com/api/policies/sharing-publication/}}. For the GPT-3 generated knowledge, we have the approval from OpenAI to release it as a part of \shortname{} to encourage future research.

\vspace{5mm}
\subsection{Generating GPT-3 Knowledge with In-Context-Learning}
\label{sec:gpt3_prompt}

\begin{figure}[h!] 
  \begin{center}
  \vspace{0.5mm}
    \frame{\includegraphics[width=0.5\textwidth]{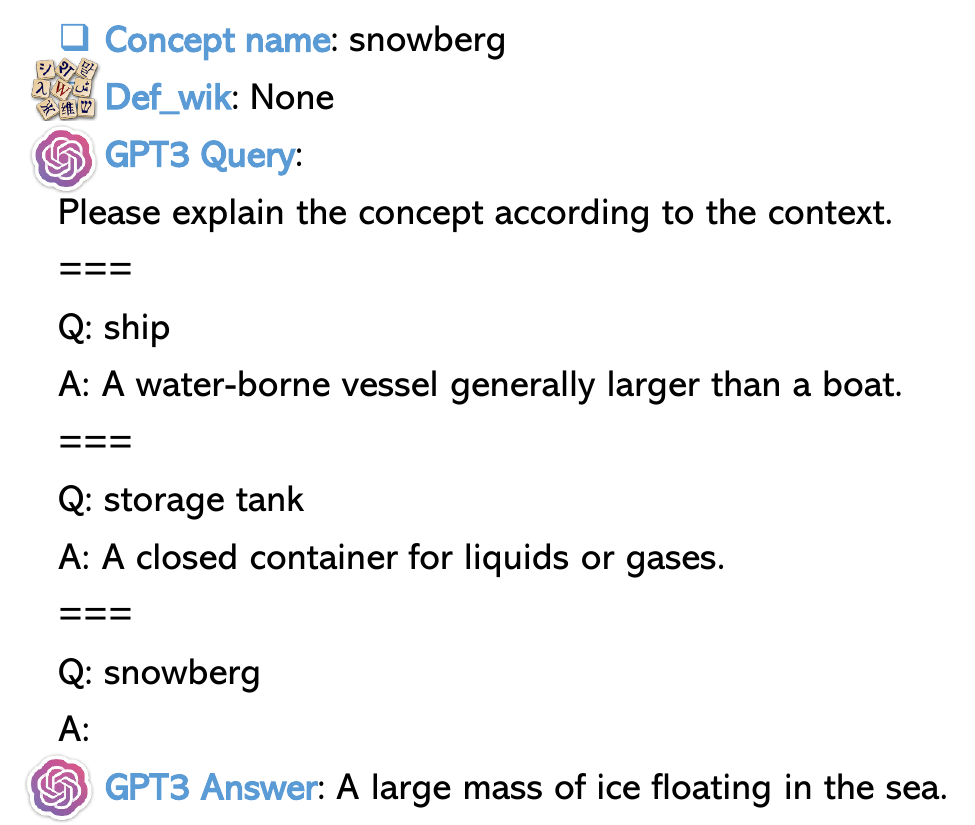}}
  \end{center}
    \vspace{-2mm}
    \caption{Example of generating external knowledge with GPT3 using in-context learning even when Wiktionary knowledge is missing.}
    \vspace{-2mm}
    \label{fig:gpt3_incontext_example}
\end{figure}

Wiktionary and WordNet do not provide a 100\% coverage for all downstream concepts. As shown in \citep{shen2022klite}, an incomplete knowledge coverage can lead to deteriorated model performance.  In this paper, we show that GPT3 can be used for generating additional external knowledge and providing a full coverage for downstream concepts.

We use in-context-learning to prompt GPT-3. As an input to GPT3, we start by asking ``Please explain the concept according to the context''.  In addition, we provide multiple concept-explaining Q (concept)-A (explanation) pairs.  Each pair of the concept and explanation are sampled from the concepts that have the Wiktionary knowledge available.  Finally, we send a different concept to GPT3, and ask for the explanation.  In this way, GPT3 is able to generate explanatory descriptions for the concepts even when its Wiktionary knowledge is missing.  For example, as shown in Fig.~\ref{fig:gpt3_incontext_example}, there is no Wiktionary knowledge available for ``snowberg'', while ``ship'' and ``storage tank'' have their corresponding Wiktionary explanations.  By providing the concept-explanation pairs of ``ship'' and ``storage tank'', GPT3 recognizes this as a concept explaining task, and when a new concept ``snowberg'' is given, it explains the concept \emph{without} the need for its external knowledge.  By randomly sampling different Q-A groups from the concepts \emph{with} Wiktionary knowledge, we are able to generate a diverse set of GPT3 responses.



\begin{table*}[t!]\centering
\begin{minipage}{\textwidth}\vspace{0mm}    \centering
\begin{tcolorbox} 
\vspace{-2mm}
\begin{adjustbox}{scale=0.8,tabular=p{2.6cm}p{14cm},center}
\textbf{Dataset name} & Oxford Flowers 102 \\
\textbf{Category names} & \textcolor{red!60}{['pink primrose', $\cdots$]} \\
\textbf{Templates} &  [
    'a photo of a \{\}, a type of flower.',
] \\
\textbf{Knowledge} & 
\textcolor{blue}{
[{"classname": "pink primrose",  "def\_wiki": "A flowering plant of the genus Primula.", "path\_wn": "", "def\_wn": "", "gpt3": [" A plant of the genus Primula, having a pink flower.", " Primula vulgaris, a plant of the primrose family, with pink flowers.", " A flowering plant of the genus Primula.", " A primrose, Primula $\times$ polyantha, with pink flowers.", " A plant of the genus Primula, of the family Primulaceae, having showy flowers of various colors."]}, $\cdots$ ]}
\vspace{1mm}
\\
\midrule
\\
\textbf{Prompt} & 
    $\bullet$ 'a photo of a \textcolor{red!60}{pink primrose}, a type of flower.' \\
\midrule
\\
\textbf{Prompt +} &  
    $\bullet$ 'a photo of a \textcolor{red!60}{pink primrose}, a type of flower ; \textcolor{blue}{ A flowering plant of the genus Primula.}'
 \\
\textbf{Knowledge} &  
    $\bullet$ 'a photo of a \textcolor{red!60}{pink primrose}, a type of flower ; \textcolor{blue}{ A plant of the genus Primula, having a pink flower.}'
 \\ 
  &  
    $\bullet$ 'a photo of a \textcolor{red!60}{pink primrose}, a type of flower ; \textcolor{blue}{Primula vulgaris, a plant of the primrose family, with pink flowers.}'
 \\ 
  &  
    $\bullet$ 'a photo of a \textcolor{red!60}{pink primrose}, a type of flower ;  \textcolor{blue}{A flowering plant of the genus Primula.}'
 \\ 
  &  
    $\bullet$ 'a photo of a \textcolor{red!60}{pink primrose}, a type of flower ; \textcolor{blue}{A primrose, Primula $\times$ polyantha, with pink flowers.}'
 \\ 
  &  
   $\bullet$ 'a photo of a \textcolor{red!60}{pink primrose}, a type of flower ; \textcolor{blue}{A plant of the genus Primula, of the family Primulaceae, having showy flowers of various colors.}'
\end{adjustbox}
\end{tcolorbox}
\vspace{-0mm}
\caption{Examples of prompt construction with and without external knowledge for the concept `pink primrose' on dataset `Oxford Flowers 102'.}
\label{tab:prompt_samples}
\end{minipage}
\vspace{-5mm}
\end{table*}

\subsection{Prompting and Knowledge}
\label{sec:prompt_and_knowledge}
For each visual recognition dataset, there comes naturally with a set of category names. A specific set of natural language templates are created for each dataset, following~\cite{radford2021learning}. In our toolkit (\href{https://github.com/Computer-Vision-in-the-Wild/Elevater_Toolkit_IC/blob/main/vision_benchmark/datasets/prompts.py}{$\mathtt{ vision\_benchmark/datasets/prompts.py}$}), we maintain the mappings from a dataset to its specific category names and template sets, respectively. External knowledge for each dataset is maintained at the folder \href{https://github.com/Computer-Vision-in-the-Wild/Elevater_Toolkit_IC/tree/main/vision_benchmark/resources/knowledge}{$\mathtt{ vision\_benchmark/resources/knowledge}$}. To construct the language prompt, we suggest the following steps:

\begin{enumerate}
    \item For a given dataset, choose one category from a set of its category names 
    \item Choose one template from a set of pre-defined dataset-specific language templates.
    \item Fill in the category name into the template, which yields the constructed language prompt for this category. 
    \item (Optional) If external knowledge is preferred to add into the prompt construction, please select a knowledge source with non-empty value, and concatenate the knowledge sequence after the text sequence in Step 3, separated by ``;''.
\end{enumerate}

In Table~\ref{tab:prompt_samples}, we provide examples to construct prompts with and without external knowledge, by following the above procedure.

\section{Evaluation}
\label{sec:evaluation_appendix}

\subsection{Leaderboards}
As demonstrated in Section~\ref{sec:evaluation_settings}, we advocate an evaluation setting with efficiency considerations, which decomposes the adaptation cost into two orthogonal dimensions: sample-efficiency and parameter-efficiency. To encourage future users compare their models with efficiency considerations,  We build the public leaderboards on EvalAI:

\begin{itemize} 

\item Image Classification in the Wild (ICinW)\\
\url{https://eval.ai/web/challenges/challenge-page/1832/overview}
\item Object Detection in the Wild (ODinW)\\
\url{https://eval.ai/web/challenges/challenge-page/1839/overview}

\end{itemize}

\subsection{A new metric with performance-efficiency trade-off}
For parameter-efficiency track, to compare different methods with a single number that considers both prediction accuracy and parameter-efficiency, we define the performance-efficiency (PE) metric: 

\begin{align}\label{eq:sampling}
    \text{PE} = \texttt{score} * \exp(\log_{10}( \texttt{\# trainable-parameters} / M_0 + 1))
\end{align}
where $\texttt{score}$ measures the prediction accuracy, while $\texttt{\# trainable-parameters} $ is the number of updated parameters in the model adaptation stage, and $M_0$ is the normalization constant. We set $M_0=10^8$ because most existing vision backbone model size are designed in this magnitude, for example, ViT-Base (80M parameters) and ViT-Large (300M parameters). With larger models designed in the future, one may increase $M_0$ for sensible measurement.

\section{Toolkit}
\label{sec:toolkit_appendi}
Our code is under MIT license.

\subsection{Automatic Hyper-parameter Tuning}
\label{sec:auto_tuning}

\paragraph{Image Classification.}
For a given dataset, we split its training set into training and validation with a ratio 80\% vs 20\%. At least one training sample per class is ensured for training and validation.  Grid search is applied over learning rate $\eta$  and weight decay $\alpha$. In the hyper-parameter search stage, the model is trained with a given configuration $(\eta, \alpha)$ for 10 epochs, the best hyper-parameter configuration is chosen as the one with the best validation performance along the entire process. After that, a final run is performed for 50 epochs to report the performance on the testing set.

\paragraph{Object Detection.} A validation set is chosen in the hyper-parameter search stage. We consider validation set size $(1, 1, 1, 3, full)$ for $N=1, 3, 5, 10$, respectively. For each type of checkpoints (DyHead, GLIP) and each adaption method, we have a set of pre-selected hyper-parameters, \ie batch size |$\mathcal{B}$|, initial learning rate $\eta_0$ and weight decay $\alpha$, as shown in Table~\ref{tab:odinw_hyperparams} in Appendix. They are determined by either empirical rules or simple hyper-parameter tuning. For each setting and each train/val split, we evaluate on the val split after every training epoch to decrease the learning rate in a step-wise manner. More specifically, we use the PyTorch {\it ReduceLROnPlateau} with patience 3 and factor 0.1 to decrease the learning rate when there is no improvement on val. We terminate the fine-tuning process if we do not see improvements for continuously 9 epochs, return the checkpoint with the best score on val, and report its score on the test split. For each few-shot setting, we random sample the train/val split 3 times, and report the average score and standard deviation on the test split. 
For each type of checkpoints (DyHead, GLIP) and each adaption method, we have a set of pre-selected hyper-parameters, \ie batch size |$\mathcal{B}$|, initial learning rate $\eta_0$ and weight decay $\alpha$, as shown in Table~\ref{tab:odinw_hyperparams}. They are determined by either empirical rules or simple hyper-parameter tuning.

\begin{table*}[h!]
    \centering
    \footnotesize
    \scalebox{0.99}{
    \begin{tabular}{@{}p{2.0cm}@{} l|  lll}
    \toprule
       \multicolumn{2}{c|}{ \multirow{2}{*}{Settings} }   &  \multicolumn{3}{c}{ 35 OD datasets }  \\
       \cmidrule{3-5} 
       Checkpoint  & Adaptation &     |$\mathcal{B}$| &  $\eta_0$ & $\alpha$  \\
         \midrule
        \multirow{3}{4.2cm}{GLIP \\(Swin-Tiny)}    
         &  Prompt  &  \multirow{3}{0.8cm}{4}  &   0.05 &  0.25 \\    
         &  Linear Probing  &    & 0.0001   & 0.05  \\      
         &  Fine-tuning    &   &  0.0001   & 0.05 \\
         \midrule
        \multirow{2}{4.2cm}{DyHead \\(Swin-Tiny)}         
         &  Linear Probing      &  \multirow{2}{0.8cm}{4}  
         & 0.0001 &  0.05 \\      
         &  Fine-tuning    &   
         & 0.0001 &  0.05 \\ 
        \bottomrule
    \end{tabular}
    }
    \vspace{-0mm}
    \caption{Pre-selected hyperparameters for OD datasets.}
    \label{tab:odinw_hyperparams}
      \vspace{-3mm}
\end{table*}

\subsection{Implementation details}
\label{sec:implementation_detail_appendix}

\paragraph{Image Classification.}
To make a fair comparison between different methods in image classification, we conduct experiments with $\texttt{FP32}$ precision. Our preliminary experiments show that on average $\texttt{FP16}$ and $\texttt{FP32}$ yields similar zero-shot performance, while $\texttt{FP32}$ models outperform  $\texttt{FP16}$ ones on 16 out of 20 datasets. 

%
\paragraph{Object Detection.} 
For OD, one image could contain multiple classes. We run an algorithm to go over the images in the full training set one by one, and add the image to the $N$-shot training set if the image contains some classes that do not have $N$ images yet. We stop if all classes have at least $N$ images or we have exhausted the full training set. Thus, the total number of images in the dataset could be between $N \sim N * K$, where $K$ is the number of categories. We will release all the $N$-shot samples we used for experiments~\cite{li2021grounded}.
For OD full fine-tuning, the common practice is to freeze the bottom two layers of the backbone\footnote{\url{shorturl.at/AOZ13}}.


\section{Close the Gap between Pre-training and Adaption for CLIP}
\label{sec:gap_clip_ce}
In Section~\ref{sec:toolkit}, we have proposed language-initialized adaptation strategy, which consistently improves the linear probing and fine-tuning performance of language-image pre-trained models like CLIP.  By initializing the linear head of CLIP model with the embeddings from the language encoder, it allows the model update and prediction of CLIP in few- / full-shot adaptation settings behaving in a similar way as in the zero-shot setting.  This, in other words, narrows the gap between the pre-training CLIP objective and the downstream image classification objective (cross-entropy).  In this section, we explore other factors that differs in the pre-training CLIP and downstream CLIP adaptations.

\begin{table}[h!]
    \centering
    \footnotesize
    \setlength{\tabcolsep}{2.2pt}
    \scalebox{0.76}{
    \begin{tabular}{cc|cc|cc|cccccccccccccccccccc}
        \toprule
        BN & $\ell_2$ & $\exp(\tau)$ & \fire & mean & std & \rotatebox[origin=l]{90}{Caltech101} & \rotatebox[origin=l]{90}{CIFAR10} & \rotatebox[origin=l]{90}{CIFAR100} & \rotatebox[origin=l]{90}{Country211} & \rotatebox[origin=l]{90}{DTD} & \rotatebox[origin=l]{90}{EuroSat} & \rotatebox[origin=l]{90}{FER2013} & \rotatebox[origin=l]{90}{FGVCAircraft} & \rotatebox[origin=l]{90}{Food101} & \rotatebox[origin=l]{90}{GTSRB} & \rotatebox[origin=l]{90}{HatefulMemes} & \rotatebox[origin=l]{90}{KittiDistance} & \rotatebox[origin=l]{90}{MNIST} & \rotatebox[origin=l]{90}{Flowers102} & \rotatebox[origin=l]{90}{OxfordPets} & \rotatebox[origin=l]{90}{PatchCamelyon} & \rotatebox[origin=l]{90}{SST2} & \rotatebox[origin=l]{90}{RESISC45} & \rotatebox[origin=l]{90}{StanfordCars} & \rotatebox[origin=l]{90}{VOC2007} \\
        \midrule
        \rowcolor{emerald!30} \cmark & \xmark & 1.0 & \xmark & 63.3 & 3.2 & 88.8 & 91.3 & 73.0 & 16.6 & 51.8 & 79.3 & 52.2 & 23.1 & 84.0 & 60.4 & 55.8 & 44.3 & 60.5 & 67.3 & 86.9 & 61.8 & 59.2 & 70.8 & 56.3 & 82.4 \\
        \midrule
        \xmark & \cmark & 1.0 & \xmark & 61.5 & 2.0 & 88.9 & 90.2 & 72.2 & 17.3 & 48.5 & 79.5 & 53.5 & 21.1 & 84.2 & 36.5 & 55.8 & 42.0 & 54.4 & 67.4 & 87.7 & 65.3 & 56.9 & 67.1 & 59.6 & 82.3 \\
        \cmark & \cmark & 1.0 & \xmark & 60.8 & 2.4 & 86.0 & 90.4 & 70.3 & 16.6 & 45.6 & 71.7 & 53.9 & 19.6 & 83.6 & 35.9 & 55.8 & 41.8 & 66.5 & 64.9 & 85.6 & 65.6 & 58.9 & 65.8 & 54.6 & 82.5 \\
        \xmark & \xmark & 1.0 & \xmark & 62.7 & 3.1 & 88.8 & 91.2 & 72.7 & 17.3 & 49.9 & 73.5 & 53.2 & 21.9 & 84.4 & 37.0 & 55.8 & 52.8 & 52.0 & 80.7 & 87.8 & 64.9 & 59.3 & 68.9 & 59.9 & 82.0 \\
        \midrule
        \cmark & \xmark & 1.0 & \cmark & 65.0 & 2.8 & 90.0 & 91.1 & 71.4 & 16.9 & 57.8 & 80.0 & 52.7 & 26.5 & 83.4 & 69.2 & 55.8 & 41.6 & 61.4 & 79.5 & 87.3 & 64.2 & 59.1 & 76.3 & 54.2 & 82.7 \\
        \cmark & \xmark & 100 & \cmark & 58.5 & 3.9 & 86.8 & 90.5 & 45.4 & 7.8 & 47.2 & 71.8 & 42.3 & 20.0 & 79.4 & 59.1 & 54.2 & 40.1 & 61.5 & 58.9 & 86.0 & 62.8 & 59.6 & 69.0 & 48.4 & 79.1 \\
        \bottomrule
    \end{tabular}
    }
    \vspace{-0mm}
    \caption{Effect of the normalization and temperature with 5-shot finetuning CLIP (ViT/B-32).  The linear head is initialized with the proposed language-initialization adaptation strategy. \fire Trainable $\tau$.
    }
    \label{table:effect_norm_temp}
    \vspace{-3mm}
\end{table}

\subsection{Visual Feature Normalization}

There are two difference in the normalization strategy between the CLIP pre-training and fine-tuning.  In CLIP, visual features $\Umat$ are normalized per-instance using $\ell_2$-norm ~\cite{radford2021learning}; while in downstream adaptation, usually a batch normalization (BN)~\cite{ioffe2015batch} without the learnable affine transformation is used for feature normalization~\cite{doersch2015unsupervised,he2021masked}.  We compare between these two normalization strategies as well as the setting without feature normalization.

As shown in Table~\ref{table:effect_norm_temp} (Row 1-4), using the channel BN yields the best performance.  In addition, adding instance-wise $\ell_2$ normalization does not help improve the performance.  This suggests that it is not always beneficial to adopt the objectives / tricks from CLIP, as there are still differences in the training objectives between CLIP and downstream classification, which we discuss in Sec.~\ref{sec:ablation_training_objective}.

\subsection{Training objective}
\label{sec:ablation_training_objective}

Although the training objective is aligned between the pre-training and downstream adaptation already with the proposed language-initialization adaptation strategy, there are several factors that may cause a difference in the gradient flow between pre-training and downstream adaptation, which can potentially hurdle the model training.

\paragraph{The size of Softmax: $|\mathcal{B}|$ vs $K$.} In CLIP, a scaled pairwise cosine similarity is first computed between all image-text pairs, and the bidirectional cross entropy loss is then applied to the computed similarity score.  Although the loss function of pre-training CLIP and downstream adaptation can be reduced to the same objective, one key difference is the size of the similarity matrix.  For each image, the similarity is computed with \emph{all} text embeddings.  In CLIP, it is the number of all text samples in a \emph{large} batch (\eg $|\mathcal{B}|$=32,768); while in downstream, it is the number of text embeddings of all classes $K$ (which is typically less than 200).  Such disparity can cause a significant change in the pattern of the gradient flow.

\paragraph{Temperature.} In CLIP, a trainable log-parameterized temperature $\tau$ controls the range of the logits in the Softmax, which is typically not used in downstream adaptation.  Although the temperature parameter does not alter the ranking of its predictions, it modifies the scale of the gradients when backward propagation is performed in downstream adaptation.

\paragraph{Experiment/Analysis.} 

Based upon the above analysis, we design experiments to explore the effect of these factors on the gradient flow and the downstream adaptations.

We compare the initialization of the temperature $\tau$ and whether to keep it frozen during the adaptation in Table~\ref{table:effect_norm_temp} (Row 1,5-7).  First, setting it to trainable has minimal effect to the training process; as there are now only $K$ classes, it might not be as important in CLIP to have a learnable $\tau$.  Second, initializing it with the pretrained checkpoint (after training with CLIP, $\exp(\tau)=100$) yields a significant performance drop.  We attribute this performance drop to the change in the size of Softmax from $|\mathcal{B}_{\text{CLIP}}|$ to $K$, where $|\mathcal{B}_{\text{CLIP}}| \gg K$.  Having a large temperature coefficient like $\exp(\tau)=100$ dramatically increases the sharpness in the pattern of Softmax and its gradient flow, which is inappropriate for training.


\subsection{Conclusion}

The language-augmented initialization is the most critical componenet in aligning the training behavior of CLIP models (30\%+ mean score improvement for 5-shot finetuning), without which the pre-trained capacity in the language encoder would be completely lost.  Other factors like visual feature normalization, batch size, temperature, \etc~ have a much smaller effect to the training procedure.  We choose to use the parameter-free batch normalization, keep the traditional batch size, and not bring in additional parameters like temperatures, for trading off between the performance and the simplicity of the model.

\section{Empirical Comparisons of Existing  Pre-trained Vision Models}
\label{sec:comparisons_foundation_models}

\subsection{A Taxonomy of Pre-trained Vision Models}
\label{sec:taxonomy}

We provide the taxonomy for pre-trained  vision models from the perspective whether language and/or is employed in pre-training, as shown in Table~\ref{tab:model_taxonomy}.
The taxonomy is a two-level hierarchy.

\begin{enumerate}
    \item In the 1st level hierarchy, given a visual recognition problem (IC or OD), the models are first categorized into language-augmented or language-free, depending on whether language is used or not in pre-training.
    \item In the 2nd level hierarchy, the language-augmented models are further categorized into knowledge-augmented or knowledge-free, depending on whether the textual external knowledge is used or not in pre-training.
\end{enumerate}
Note that our taxonomy is only related to pre-training, which is independent from how the model is adapted to a downstream task.

For knowledge-augmented pre-trained models such as K-LITE~\cite{shen2022klite}, the model is pre-trained with both natural language supervision and external knowledge supervision. The external knowledge is employed in the following manner: (1) For image-text pairs, query is identified using entity extraction on the text, (2) The relevant “knowledge text” of the query is retrieved from knowledge bases; (3) The retrieved “knowledge text” is appended to the original text. In the downstream adaptation stage, it follows the same prompting process with other pre-trained models, as described in Section~\ref{sec:prompt_and_knowledge}.

\begin{table}[h!] 
  \begin{center}
  \vspace{0.5mm}
    \includegraphics[width=0.95\textwidth]{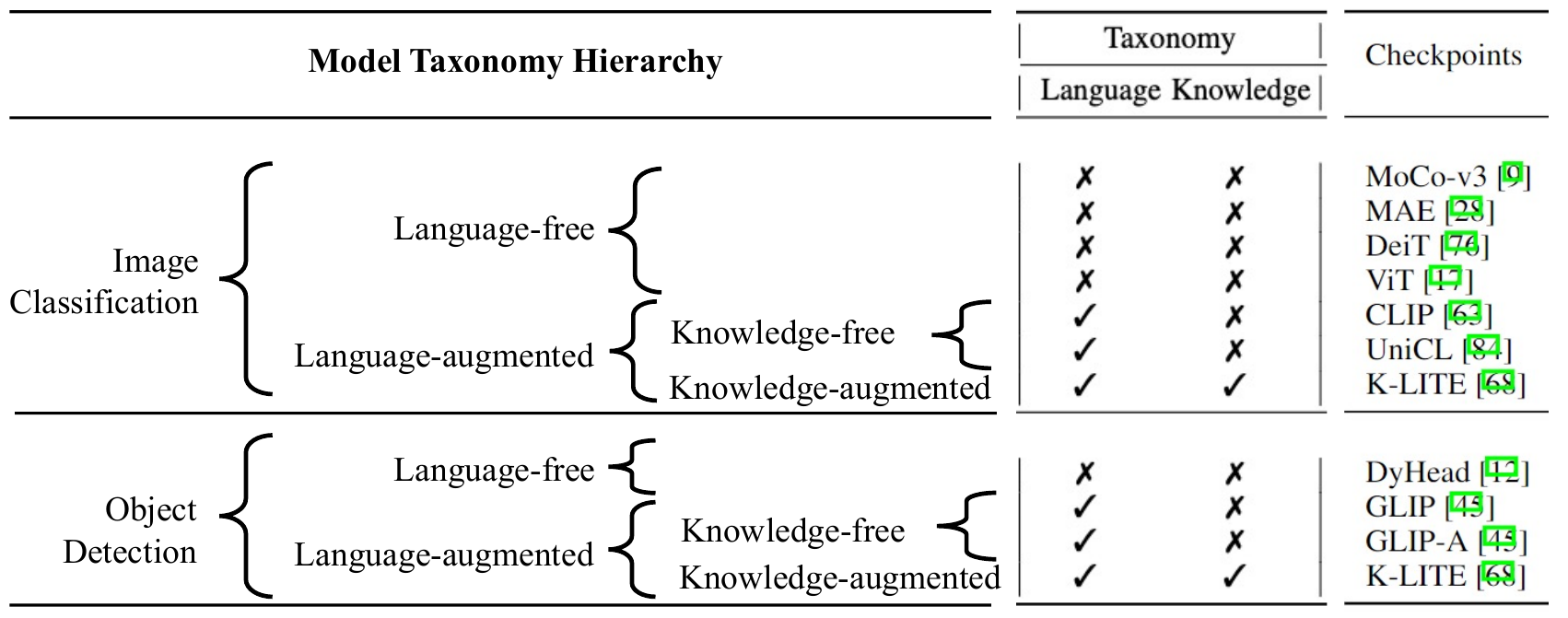}
  \end{center}
    \vspace{-0mm}
    \caption{A Taxonomy of Vision Pre-trained Models}
    \vspace{-0mm}
    \label{tab:model_taxonomy}
\end{table}

\begin{table*}[t!]
    \centering
    \footnotesize
    \scalebox{0.92}{
\begin{tabular}{@{}p{1.7cm}@{} c c| cccc}
    \toprule
       \multicolumn{3}{c|}{ \multirow{1}{*}{Pre-training Settings} }  &  \multicolumn{4}{c}{ 20 Image Classification Datasets }  \\
       \cmidrule{4-7} 
       Checkpoint  & Method & Dataset &   5-shot & 20-shot & 50-shot & Full-shot  \\
        \midrule
        \multicolumn{7}{c}{ \bf{Linear Probing} } \\
        CLIP$^\ddagger$ & Image-Text Contrast  & WebImageText (400M)   & 68.27 {\tiny $\pm$ 0.97} & 74.76 {\tiny $\pm$ 1.11} & 77.75 {\tiny $\pm$ 0.81} & 81.17 \\
        ViT$^\dagger$ & Supervised & ImageNet-22K (14M)  & 57.61 {\tiny $\pm$ 3.62} & 69.93 {\tiny $\pm$ 0.71} & 73.74 {\tiny $\pm$ 0.79} & 77.60 \\ 
        DeiT  & Supervised & ImageNet-1K (1.2M)   & 54.06 {\tiny $\pm$ 3.02} & 68.57 {\tiny $\pm$ 3.43} & 75.53 {\tiny $\pm$ 0.72} & 79.56 \\ 
        MAE & Self-Supervised & ImageNet-1K (1.2M)  & 33.37 {\tiny $\pm$ 1.98} & 48.03 {\tiny $\pm$ 2.70} & 58.26 {\tiny $\pm$ 0.84} & 68.70 \\  
        CAE  & Self-Supervised & ImageNet-1K (1.2M)  & 44.15 {\tiny $\pm$ 0.31} & 57.93 {\tiny $\pm$ 0.19} & 64.37 {\tiny $\pm$ 0.23} & 70.56 \\
        MoCo-v3 & Self-Supervised & ImageNet-1K (1.2M)  & 50.17 {\tiny $\pm$ 3.43} & 61.99 {\tiny $\pm$ 2.51} & 69.71 {\tiny $\pm$ 1.03} & 74.92 \\
        Random & - & -  & 
        19.64 {\tiny $\pm$ 1.68} & 23.89 {\tiny $\pm$ 1.47} & 26.86 {\tiny $\pm$ 0.69} & 31.64 \\        
        \midrule
         \multicolumn{7}{c}{ \bf{Fine-tuning } } \\
        CLIP$^\ddagger$  & Image-Text Contrast  & WebImageText (400M)   & 69.12 {\tiny $\pm$ 1.66} & 74.76 {\tiny $\pm$ 2.34} & 78.21 {\tiny $\pm$ 2.04} & 83.63 \\
        ViT$^\dagger$ & Supervised & ImageNet-22K (14M)   & 57.18 {\tiny $\pm$ 2.02} & 72.45 {\tiny $\pm$ 2.85} & 78.53 {\tiny $\pm$ 0.69} & 82.02 \\ 
        DeiT  & Supervised & ImageNet-1K (1.2M)   & 54.06 {\tiny $\pm$ 3.02} & 68.53 {\tiny $\pm$ 3.47} & 75.57 {\tiny $\pm$ 0.68} & 79.55 \\ 
        MAE & Self-Supervised & ImageNet-1K (1.2M)   & 36.10 {\tiny $\pm$ 3.25} & 54.13 {\tiny $\pm$ 3.86} & 65.86 {\tiny $\pm$ 2.42} & 74.43 \\ 
         CAE  & Self-Supervised & ImageNet-1K (1.2M)  & 37.87 {\tiny $\pm$ 1.03} & 58.04 {\tiny $\pm$ 2.07} & 71.39 {\tiny $\pm$ 0.79} & 77.79 \\
        MoCo-v3  & Self-Supervised & ImageNet-1K (1.2M)  & 39.30 {\tiny $\pm$ 3.84} & 58.75 {\tiny $\pm$ 5.55} & 70.33 {\tiny $\pm$ 1.64} & 77.71 \\
        Random & - & -  & 
        20.85 {\tiny $\pm$ 1.59} & 26.29 {\tiny $\pm$ 1.21} & 30.88 {\tiny $\pm$ 1.68} & 43.73 \\        
        \bottomrule
    \end{tabular}
    }
    \vspace{-0mm}
    \caption{Averaged scores on 20 IC datasets with the {\bf ViT-B16} network architecture. $^\ddagger$ CLIP is adapted using the proposed language-augmented initialization. $^\dagger$ ViT checkpoint is pre-trained on ImageNet-22K, then fine-tuned on ImageNet-1K. The zero-shot performance of CLIP is 59.96\%.
    }
    \label{tab:perf_ic_vitb16}
      \vspace{-2mm}
\end{table*}

\begin{table}[t!]
    \centering
    \footnotesize
    \setlength{\tabcolsep}{2.2pt}
    \scalebox{0.8}{
    \begin{tabular}{cc|c|cccccccccccccccccccc}
        \toprule
        Ckpt. & Shot & Score & \rotatebox[origin=l]{90}{Caltech101} & \rotatebox[origin=l]{90}{CIFAR10} & \rotatebox[origin=l]{90}{CIFAR100} & \rotatebox[origin=l]{90}{Country211} & \rotatebox[origin=l]{90}{DTD} & \rotatebox[origin=l]{90}{EuroSat} & \rotatebox[origin=l]{90}{FER2013} & \rotatebox[origin=l]{90}{FGVCAircraft} & \rotatebox[origin=l]{90}{Food101} & \rotatebox[origin=l]{90}{GTSRB} & \rotatebox[origin=l]{90}{HatefulMemes} & \rotatebox[origin=l]{90}{KittiDistance} & \rotatebox[origin=l]{90}{MNIST} & \rotatebox[origin=l]{90}{Flowers102} & \rotatebox[origin=l]{90}{OxfordPets} & \rotatebox[origin=l]{90}{PatchCamelyon} & \rotatebox[origin=l]{90}{SST2} & \rotatebox[origin=l]{90}{RESISC45} & \rotatebox[origin=l]{90}{StanfordCars} & \rotatebox[origin=l]{90}{VOC2007} \\
        \midrule
        \multicolumn{23}{c}{ \bf{Linear Probing} } \\
\multirow{4}{*}{CLIP} & 5 & 68.3 & 91.3 & 91.4 & 71.1 & 21.7 & 61.6 & 76.7 & 53.6 & 36.0 & 89.7 & 55.9 & 58.0 & 44.8 & 76.7 & 94.2 & 90.5 & 54.3 & 62.0 & 78.3 & 73.6 & 84.2 \\
 & 20 & 74.8 & 94.3 & 93.0 & 75.4 & 25.2 & 73.7 & 86.6 & 54.7 & 48.1 & 90.6 & 75.7 & 58.5 & 50.3 & 90.5 & 96.8 & 92.3 & 68.0 & 63.8 & 87.5 & 83.9 & 86.3 \\
 & 50 & 77.8 & 94.4 & 93.8 & 78.0 & 27.7 & 76.3 & 90.0 & 57.5 & 53.5 & 91.3 & 81.6 & 60.0 & 61.4 & 95.9 & 96.8 & 93.8 & 69.9 & 68.7 & 90.1 & 87.2 & 87.1 \\
 & full & 81.2 & 94.4 & 95.8 & 82.2 & 31.2 & 77.4 & 94.5 & 68.5 & 52.8 & 92.8 & 88.6 & 65.1 & 67.7 & 98.9 & 96.5 & 94.0 & 83.5 & 74.5 & 90.8 & 87.2 & 87.0 \\
 \cmidrule{4-23} 
\multirow{4}{*}{MAE} & 5 & 33.4 & 59.0 & 34.0 & 21.2 & 2.8 & 35.0 & 64.4 & 21.3 & 7.0 & 7.7 & 17.5 & 51.4 & 46.1 & 63.4 & 50.9 & 17.2 & 54.9 & 50.1 & 38.9 & 6.3 & 18.3 \\
 & 20 & 48.0 & 85.5 & 44.9 & 43.5 & 4.4 & 58.3 & 74.1 & 23.5 & 29.9 & 30.4 & 41.1 & 51.7 & 49.8 & 52.9 & 71.9 & 60.0 & 52.7 & 53.2 & 67.4 & 25.5 & 39.9 \\
 & 50 & 58.3 & 88.7 & 67.3 & 53.3 & 6.9 & 66.0 & 86.4 & 27.1 & 39.2 & 42.8 & 57.0 & 50.8 & 54.0 & 81.5 & 71.9 & 76.5 & 69.4 & 51.6 & 78.6 & 36.7 & 59.2 \\
 & full & 68.7 & 87.7 & 88.2 & 68.3 & 10.1 & 66.3 & 94.8 & 56.0 & 39.1 & 65.1 & 76.3 & 56.2 & 78.8 & 99.3 & 72.0 & 81.6 & 86.0 & 58.4 & 81.2 & 37.2 & 71.4 \\
\cmidrule{4-23} 
 \multirow{4}{*}{CAE} 
& 5 & 43.8 & 74.7 & 61.6 & 38.3 & 3.5 & 43.7 & 76.7 & 24.5 & 14.3 & 18.6 & 33.8 & 47.9 & 42.3 & 57.8 & 70.3 & 37.3 & 63.2 & 52.1 & 54.4 & 8.7 & 51.3 \\
    & 20 & 57.9 & 87.3 & 76.4 & 55.1 & 5.5 & 62.0 & 89.0  & 32.5 & 32.6 & 35.7 & 54.3 & 51.6 & 57.3 & 88.9 & 81.2 & 63.3 & 69.9 & 52.2 & 72.1 & 27.5  & 64.4 \\
    & 50 & 71.4 & 93.9 & 90.6 & 78.3 & 6.8 & 69.4 & 93.2 & 43.2 & 56.1 & 59.4 & 93.5 & 53.0 & 61.6 & 96.2 & 85.9 & 90.0 & 81.1 & 52.3 & 88.0 & 69.5 & 65.8 \\
 & full & 70.6 & 90.0 & 93.9 & 78.9 & 11.4 & 66.3 & 96.7 & 57.9 & 40.8 & 67.4 & 78.9 & 55.6 & 75.7 & 99.0 & 81.2 & 79.8 & 85.9 & 58.8 & 82.7 & 40.4 & 70.0 \\
  
  \cmidrule{4-23} 
\multirow{4}{*}{MoCo-v3} & 5 & 50.2 & 80.8 & 78.5 & 60.5 & 4.8 & 57.1 & 77.1 & 20.5 & 11.8 & 36.6 & 31.4 & 50.7 & 46.7 & 64.1 & 79.5 & 76.2 & 54.7 & 50.0 & 61.1 & 13.4 & 47.9 \\
 & 20 & 62.0 & 91.3 & 67.7 & 75.5 & 7.6 & 66.3 & 84.8 & 30.9 & 38.2 & 59.3 & 53.9 & 53.5 & 48.5 & 81.8 & 89.5 & 86.4 & 52.1 & 51.6 & 77.3 & 49.5 & 74.2 \\
 & 50 & 69.7 & 92.1 & 93.6 & 79.0 & 10.3 & 73.4 & 92.3 & 40.2 & 48.0 & 66.8 & 66.7 & 50.3 & 60.5 & 88.3 & 89.5 & 90.2 & 75.1 & 51.3 & 84.1 & 63.1 & 79.2 \\
 & full & 74.9 & 92.1 & 96.9 & 85.3 & 13.7 & 73.1 & 95.9 & 60.1 & 48.0 & 78.0 & 78.7 & 53.7 & 68.8 & 98.4 & 89.5 & 91.4 & 86.7 & 57.1 & 86.3 & 63.0 & 81.7 \\
  \cmidrule{4-23} 
\multirow{4}{*}{DeiT} & 5 & 54.1 & 86.2 & 70.1 & 61.5 & 4.4 & 52.9 & 62.5 & 14.5 & 24.1 & 41.9 & 46.7 & 51.1 & 47.6 & 83.8 & 82.7 & 87.8 & 51.5 & 50.1 & 63.4 & 27.6 & 70.9 \\
 & 20 & 68.6 & 93.9 & 91.2 & 73.7 & 6.2 & 68.7 & 90.7 & 35.2 & 34.1 & 61.5 & 86.7 & 50.8 & 52.4 & 90.7 & 92.7 & 91.9 & 66.7 & 51.7 & 82.7 & 68.8 & 81.1 \\
 & 50 & 75.5 & 94.7 & 94.2 & 82.0 & 8.8 & 73.9 & 94.4 & 40.8 & 60.6 & 73.2 & 96.5 & 53.4 & 69.7 & 98.1 & 92.7 & 93.4 & 77.4 & 52.2 & 89.4 & 82.9 & 82.3 \\
 & full & 79.6 & 94.9 & 98.2 & 89.6 & 14.1 & 72.8 & 98.2 & 69.3 & 59.3 & 84.5 & 98.8 & 44.3 & 82.0 & 99.6 & 92.4 & 93.9 & 89.9 & 52.6 & 90.8 & 83.0 & 83.1 \\
  \cmidrule{4-23} 
\multirow{4}{*}{ViT} & 5 & 57.6 & 93.2 & 88.2 & 75.4 & 6.8 & 63.9 & 70.0 & 25.2 & 22.7 & 59.0 & 29.9 & 48.5 & 46.5 & 68.3 & 99.2 & 89.6 & 61.3 & 49.9 & 57.9 & 27.6 & 69.2 \\
 & 20 & 69.9 & 95.6 & 94.8 & 84.0 & 11.5 & 75.7 & 86.5 & 45.4 & 40.5 & 81.7 & 51.1 & 53.5 & 57.1 & 87.7 & 99.2 & 92.6 & 72.0 & 52.4 & 79.7 & 53.9 & 83.7 \\
 & 50 & 73.7 & 96.0 & 96.4 & 86.8 & 15.2 & 78.8 & 91.5 & 50.0 & 48.5 & 85.1 & 62.1 & 51.0 & 60.1 & 91.7 & 99.2 & 93.9 & 77.7 & 51.5 & 85.4 & 67.3 & 86.6 \\
 & full & 77.6 & 95.9 & 98.2 & 89.8 & 16.6 & 78.9 & 96.0 & 64.5 & 47.8 & 89.6 & 76.5 & 55.1 & 69.3 & 98.2 & 99.2 & 94.8 & 85.5 & 54.6 & 86.6 & 67.5 & 87.3 \\
  \cmidrule{4-23} 
\multirow{4}{*}{Random} & 5 & 19.6 & 9.0 & 17.6 & 5.8 & 1.2 & 8.2 & 41.0 & 15.4 & 3.0 & 2.7 & 7.9 & 49.6 & 40.9 & 26.7 & 17.8 & 4.1 & 52.7 & 51.5 & 18.6 & 1.5 & 17.5 \\
 & 20 & 23.9 & 13.0 & 25.1 & 9.8 & 1.9 & 12.4 & 46.3 & 20.4 & 3.6 & 4.7 & 9.4 & 54.4 & 42.1 & 40.8 & 22.4 & 7.0 & 64.8 & 52.2 & 25.5 & 2.3 & 19.9 \\
 & 50 & 26.9 & 15.9 & 27.3 & 12.1 & 2.2 & 14.2 & 60.4 & 20.2 & 4.1 & 6.0 & 11.1 & 54.1 & 40.8 & 56.0 & 22.4 & 8.7 & 73.7 & 53.1 & 30.6 & 2.6 & 21.6 \\
 & full & 31.6 & 16.5 & 43.0 & 18.7 & 3.1 & 13.8 & 69.0 & 30.4 & 4.4 & 10.8 & 15.3 & 56.6 & 45.1 & 85.0 & 21.7 & 9.5 & 77.6 & 55.0 & 31.0 & 2.7 & 23.3 \\        \midrule
        \multicolumn{23}{c}{\bf{Fine-tuning}} \\
\multirow{4}{*}{CLIP} & 5 & 69.1 & 91.2 & 92.1 & 73.2 & 22.2 & 53.8 & 79.0 & 55.9 & 33.5 & 87.5 & 84.3 & 55.3 & 41.9 & 84.9 & 87.1 & 91.7 & 59.4 & 59.8 & 80.1 & 66.0 & 83.5 \\
 & 20 & 74.8 & 93.7 & 93.9 & 79.7 & 21.8 & 70.6 & 94.1 & 59.0 & 52.2 & 89.0 & 91.9 & 54.3 & 52.7 & 70.0 & 93.8 & 93.0 & 71.9 & 62.6 & 87.0 & 80.0 & 84.2 \\
 & 50 & 78.2 & 94.5 & 94.7 & 82.8 & 21.9 & 75.0 & 95.7 & 61.0 & 61.5 & 89.3 & 91.3 & 54.5 & 65.1 & 85.4 & 93.8 & 93.7 & 75.4 & 64.9 & 91.0 & 86.5 & 86.3 \\
 & full & 83.6 & 94.9 & 98.6 & 89.4 & 23.6 & 74.7 & 98.4 & 72.0 & 60.8 & 91.8 & 99.0 & 65.9 & 84.1 & 99.6 & 94.1 & 94.2 & 89.7 & 76.5 & 91.9 & 86.9 & 86.5 \\
  \cmidrule{4-23} 
\multirow{4}{*}{MAE} & 5 & 36.1 & 70.8 & 34.4 & 13.1 & 2.1 & 41.4 & 64.1 & 20.8 & 8.2 & 13.3 & 14.8 & 49.6 & 38.0 & 46.8 & 68.8 & 37.8 & 53.3 & 50.9 & 50.4 & 6.0 & 37.4 \\
 & 20 & 54.1 & 91.0 & 50.1 & 40.4 & 3.6 & 59.7 & 79.5 & 22.6 & 32.5 & 22.4 & 62.2 & 54.8 & 46.0 & 90.9 & 81.6 & 78.0 & 67.7 & 51.7 & 65.5 & 21.6 & 60.8 \\
 & 50 & 65.9 & 92.9 & 71.5 & 54.7 & 4.9 & 66.2 & 87.8 & 34.1 & 42.8 & 51.9 & 95.6 & 51.3 & 50.1 & 96.1 & 81.6 & 84.8 & 77.3 & 52.4 & 85.2 & 68.0 & 68.0 \\
 & full & 74.4 & 92.8 & 97.7 & 85.5 & 9.3 & 66.2 & 97.5 & 68.5 & 46.2 & 84.6 & 99.1 & 55.2 & 82.8 & 99.6 & 75.3 & 89.8 & 76.0 & 56.7 & 87.0 & 47.0 & 71.7 \\
  \cmidrule{4-23} 
    \multirow{4}{*}{CAE} & 5 & 39.2 & 74.4 & 54.2 & 30.3 & 1.8 & 47.5 & 68.2 & 18.9 & 5.8 & 18.1 & 10.9 & 48.4 & 35.3 & 22.6 & 73.3 & 51.9 & 50.0 & 52.7 & 58.2 & 3.7 & 57.7 \\
    & 20 & 58.0 & 89.3 & 30.6 & 62.1 & 4.9 & 62.3  & 73.6  & 24.9 & 36.3 & 30.9 & 84.9 & 49.7 & 56.0 & 66.1 & 86.0 & 84.0 & 75.8 & 52.4 & 75.3 & 50.4 & 65.2 \\
    & 50 & 71.4 & 93.9 & 90.6 & 78.3 & 6.8 & 69.4 & 93.2 & 43.2 & 56.1 & 59.4 & 93.5 & 53.0 & 61.6 & 96.2 & 85.9 & 90.0 & 81.1 & 52.3 & 88.0 & 69.5 & 65.8 \\
 & full & 77.8 & 93.2 & 98.6 & 89.1 & 12.8 & 68.0 & 98.1 & 68.8 & 44.3 & 87.3 & 99.2 & 57.1 & 84.3 & 99.8 & 87.6 & 92.1 & 91.8 & 56.7 & 89.9 & 61.6 & 75.4 \\
 
  \cmidrule{4-23} 
\multirow{4}{*}{MoCo-v3} & 5 & 39.3 & 73.7 & 70.3 & 17.4 & 2.3 & 45.6 & 60.0 & 13.5 & 7.2 & 27.6 & 16.5 & 50.8 & 43.5 & 18.1 & 65.7 & 77.1 & 50.9 & 50.7 & 58.2 & 11.2 & 25.7 \\
 & 20 & 58.8 & 91.9 & 58.4 & 59.2 & 5.0 & 63.4 & 69.7 & 19.8 & 47.4 & 55.5 & 86.7 & 53.5 & 48.5 & 53.4 & 85.8 & 87.4 & 51.5 & 51.4 & 78.5 & 49.2 & 59.2 \\
 & 50 & 70.3 & 92.8 & 89.1 & 77.5 & 6.9 & 71.3 & 92.6 & 31.0 & 53.4 & 63.2 & 96.5 & 50.9 & 57.3 & 94.3 & 85.8 & 90.2 & 74.2 & 50.4 & 87.3 & 66.2 & 75.7 \\
 & full & 77.7 & 93.3 & 98.1 & 88.7 & 11.7 & 71.3 & 97.3 & 68.3 & 51.9 & 84.1 & 98.8 & 54.5 & 80.5 & 99.6 & 87.1 & 90.9 & 91.4 & 52.5 & 88.6 & 67.9 & 77.6 \\
  \cmidrule{4-23} 
\multirow{4}{*}{DeiT} & 5 & 54.1 & 86.2 & 70.1 & 61.5 & 4.4 & 52.9 & 62.5 & 14.5 & 24.1 & 41.9 & 46.7 & 51.1 & 47.6 & 83.8 & 82.7 & 87.8 & 51.5 & 50.1 & 63.4 & 27.6 & 70.9 \\
 & 20 & 68.5 & 93.9 & 91.2 & 73.7 & 6.2 & 68.7 & 90.7 & 34.4 & 34.1 & 61.5 & 86.7 & 50.8 & 52.4 & 90.7 & 92.7 & 91.9 & 66.7 & 51.7 & 82.7 & 68.8 & 81.1 \\
 & 50 & 75.6 & 94.7 & 94.2 & 82.0 & 9.6 & 73.9 & 94.4 & 40.8 & 60.6 & 73.2 & 96.5 & 53.4 & 69.7 & 98.0 & 92.7 & 93.4 & 77.4 & 52.2 & 89.4 & 82.9 & 82.3 \\
 & full & 79.5 & 94.9 & 98.2 & 89.6 & 14.1 & 72.8 & 98.2 & 69.3 & 59.2 & 84.5 & 98.8 & 44.3 & 82.0 & 99.6 & 92.4 & 93.9 & 89.9 & 52.6 & 90.8 & 83.0 & 83.1 \\
  \cmidrule{4-23} 
\multirow{4}{*}{ViT} & 5 & 57.2 & 90.8 & 82.7 & 67.6 & 4.0 & 56.0 & 75.2 & 24.5 & 21.4 & 58.0 & 51.5 & 47.6 & 38.4 & 82.6 & 99.0 & 83.8 & 53.8 & 51.0 & 61.5 & 21.0 & 73.2 \\
 & 20 & 72.5 & 96.1 & 93.6 & 86.7 & 8.4 & 74.2 & 91.7 & 43.6 & 51.6 & 68.2 & 92.8 & 51.9 & 57.8 & 95.8 & 99.4 & 92.1 & 71.6 & 51.8 & 84.8 & 65.5 & 71.4 \\
 & 50 & 78.5 & 96.3 & 97.3 & 89.9 & 11.8 & 79.1 & 95.0 & 52.1 & 63.6 & 83.0 & 97.5 & 54.7 & 68.9 & 97.5 & 99.5 & 93.3 & 80.5 & 52.3 & 90.1 & 83.0 & 85.2 \\
 & full & 82.0 & 96.6 & 99.0 & 93.4 & 16.8 & 79.4 & 98.3 & 72.6 & 61.9 & 90.7 & 99.1 & 53.4 & 84.5 & 99.7 & 99.5 & 94.0 & 91.1 & 50.1 & 91.5 & 83.2 & 85.6 \\
  \cmidrule{4-23} 
\multirow{4}{*}{Random} & 5 & 20.9 & 12.4 & 16.2 & 6.6 & 1.3 & 9.4 & 38.3 & 19.9 & 3.2 & 3.2 & 8.6 & 52.4 & 41.7 & 18.6 & 25.4 & 4.7 & 62.3 & 51.1 & 21.7 & 1.8 & 18.3 \\
 & 20 & 26.3 & 24.5 & 25.3 & 13.1 & 2.2 & 16.4 & 55.3 & 20.2 & 5.1 & 5.9 & 16.6 & 52.8 & 35.2 & 38.0 & 38.3 & 8.2 & 65.7 & 50.6 & 29.1 & 3.4 & 20.1 \\
 & 50 & 30.9 & 27.5 & 31.1 & 19.6 & 3.0 & 19.9 & 67.3 & 21.9 & 6.5 & 8.5 & 28.5 & 56.0 & 42.2 & 42.7 & 38.3 & 11.6 & 73.5 & 52.3 & 41.2 & 3.7 & 22.2 \\
 & full & 43.7 & 28.3 & 65.9 & 41.4 & 4.0 & 21.8 & 85.3 & 41.0 & 7.2 & 34.0 & 83.0 & 55.7 & 50.5 & 95.6 & 35.1 & 15.3 & 81.8 & 55.6 & 44.5 & 2.6 & 26.1 \\
        \bottomrule
    \end{tabular}
    }
    \vspace{3mm}
    \caption{Breakdown results of \textbf{ViT-B16} checkpoints in Table~\ref{tab:perf_ic_vitb16}.
    }
    \label{table:experiment_breakdown_vitb16}
\end{table}

\subsection{Baseline with Vision Pre-trained Models}
\paragraph{Image Classification}
We consider seven checkpoints to produce baseline results for IC. In the main paper, we report the following four checkpoints.

\begin{itemize}[leftmargin=2.5mm]
\item \textit{Supervised ViT }~\cite{dosovitskiy2020image} represents a checkpoint for the traditional language-free visual models, where model training is performed on ImageNet-22K with cross-entropy loss.
\vspace{-1mm}
\item \textit{CLIP ViT}~\cite{radford2021learning} represents a checkpoint for the family of the language-augmented visual models, trained with 400M image-text pairs.
\vspace{-1mm}
\item \textit{UniCL Swin}~\cite{yang2022unicl} represents knowledge-free language-augmented visual models with Swin~\cite{liu2021Swin} as the visual backbone, trained in the academic setting with ImageNet-21K, which excludes ImageNet-1K categories from ImageNet-22K.
\vspace{-1mm}
\item \textit{KLITE, UniCL Swin}~\cite{shen2022klite} represents knowledge-enriched language-augmented visual models. Its pre-training setting is the same as UniCL Swin, but external knowledge such as Wiktionary is leveraged in model pre-training.
\end{itemize}

We also consider three popular language-free visual models in Appendix:

\begin{itemize}[leftmargin=2.5mm]
\item \textit{DeiT}~\cite{touvron2021training} represents a checkpoint for the supervised visual backbone, where model training is performed on ImageNet-1K with cross-entropy loss and advanced data augmentation and training schedule.
\vspace{-1mm}
\item \textit{MoCo}~\cite{chen2021empirical} represents a checkpoint for the family of augmented-view-based methods for image self-supervised learning, trained with images only in ImageNet-1K.
\vspace{-1mm}
\item \textit{MAE}~\cite{he2021masked} represents a checkpoint for the family of recent masked region (visual token) modeling based methods for image self-supervised learning, trained with images only in ImageNet-1K.
\vspace{-1mm}
\item \textit{CAE}~\cite{chen2022context} represents a checkpoint that benefits the separation of the representation learning (encoding) role and the pretext task completion role, trained with images only in ImageNet-1K.
\vspace{-1mm}

\end{itemize}

\paragraph{Object Detection}
We consider four checkpoints to produce baseline results for OD. They are for the academic track, as they are pre-trained on public datasets. All of them employ Swin-Tiny backbone~\cite{liu2021Swin}.
\begin{itemize}[leftmargin=2.5mm]
\item \textit{DyHead}~\cite{dai2021dynamic} represents a checkpoint for the traditional language-free object detector, where model  is pre-trained on Object365~\cite{shao2019objects365} without leveraging the category name information.
\vspace{-1mm}
\item \textit{GLIP}~\cite{li2021grounded} represents a checkpoint for the family of the language-augmented object detector, trained with Object365 and Flicker phrase grounding data~\cite{plummer2015flickr30k}.
\vspace{-1mm}
\item \textit{GLIP-A}~\cite{li2021grounded} represents knowledge-free language-augmented object detector, where model is trained on Object365 and the semantics of category names is leveraged.
\vspace{-1mm}
\item \textit{KLITE, GLIP-A}~\cite{shen2022klite} represents knowledge-enriched language-augmented object detector. Its training setting is the same as GLIP-A, except that Wiktionary knowledge is leveraged in model pre-training.
\end{itemize}

In summary, among four checkpoints for each problem, the first two are used to compare the state-of-the-art in language-free and language-augmented models, and latter two are used to compare the knowledge-free and knowledge-augmented models (both belongs to language-augmented models, as knowledge is presented as a structured form of language).

\subsection{Experimental Results of Different Model Checkpoints}
In Table~\ref{tab:perf_ic_vitb16}, we report IC performance with ViT-B16 pre-trained with representative methods, using different objectives and datasets.  We present its breakdown experimental results in Table~\ref{table:experiment_breakdown_vitb16}.  Note that all of the models are adapted to downstream datasets, using the same automatic hyper-parameter tuning process in our toolkit, and no model- / dataset-specific tuning is employed. This ensures fairness in model adaptation process, but may not represent the best transfer performance of each pre-trained model, if more careful tuning efforts are paid. Nevertheless, we believe the results represent the model transferability with affordable efforts, and use them as baseline results for \shortname{} benchmark.  

We found that the overall ranking of the models in the descending order: CLIP, ViT, DeiT, MoCo-v3, MAE.  Surprisingly, we found that MAE performs worse than MoCo, and both of them are worse than supervised method DeiT, though all three of them are pre-trained on the same ImageNet-1K dataset. We note that an similar observation is made in~\cite{jia2022visual}, when evaluated these checkpoints on a large range of downstream datasets. This is perhaps because the region-based pre-training tasks in MAE is can better capture region-level dependency (thus benefits dense prediction tasks such as object detection), while view-based pre-training tasks in MoCo can better capture image-level dependency (thus benefits image classification). ViT outperforms DeiT probably due to the larger pre-training dataset. CLIP performs the best.  To the best of our knowledge, language-augmented visual models such as CLIP enjoy the best scaling performance; In contrast, the scaling performance of language-free visual models are either less studied or less successful so far. 

In Table~\ref{table:additional_experiment_breakdown_lang_init}, we presented the comparisons of random and language-augmented initialization for language-image model
adaptation with more checkpoints under 5-shot settings. This includes ViT-Base and ViT-Large models of 
DeCLIP~\cite{li2021supervision},
OpenCLIP~\cite{ilharco_gabriel_2021_5143773} and CLIP~\cite{radford2021learning}.

In Table~\ref{table:experiment_breakdown_zeroshot_more_checkpoints}, we presented zero-shot results of more model checkpoints for both Industry and Academic Tracks. 
For Academic Tracks, we consider CLIP~\cite{radford2021learning}, DeCLIP~\cite{li2021supervision}, FILIP~\cite{yao2021filip}, SLIP~\cite{mu2021slip}, with network ViT-Base32 pre-trained on YFCC (15M).
For Industry Tracks,
we consdier DeCLIP,
OpenCLIP and CLIP, with models ranging from ViT-Base to ViT-Large, and training data ranging from 88M to 400M image-text pairs.

\subsection{Breakdown Experimental Results on CLIP}

We show the individual linear probing and finetuning scores for comparing the random and language-augmented initialization in Table~\ref{table:experiment_breakdown_lang_init}.  Language initialization consistently outperforms random initialization across different domains: sample efficiency, parameter efficiency, and different datasets. See Sec.~\ref{sec:toolkit} for more discussions on the design and the effectiveness of the language-augmented initializations.

\begin{table}[t!]
    \centering
    \footnotesize
    \setlength{\tabcolsep}{2.2pt}
    \scalebox{0.8}{
    \begin{tabular}{cc|c|cccccccccccccccccccc}
        \toprule
        Shot & {Lang-Init} & Score & \rotatebox[origin=l]{90}{Caltech101} & \rotatebox[origin=l]{90}{CIFAR10} & \rotatebox[origin=l]{90}{CIFAR100} & \rotatebox[origin=l]{90}{Country211} & \rotatebox[origin=l]{90}{DTD} & \rotatebox[origin=l]{90}{EuroSat} & \rotatebox[origin=l]{90}{FER2013} & \rotatebox[origin=l]{90}{FGVCAircraft} & \rotatebox[origin=l]{90}{Food101} & \rotatebox[origin=l]{90}{GTSRB} & \rotatebox[origin=l]{90}{HatefulMemes} & \rotatebox[origin=l]{90}{KittiDistance} & \rotatebox[origin=l]{90}{MNIST} & \rotatebox[origin=l]{90}{Flowers102} & \rotatebox[origin=l]{90}{OxfordPets} & \rotatebox[origin=l]{90}{PatchCamelyon} & \rotatebox[origin=l]{90}{SST2} & \rotatebox[origin=l]{90}{RESISC45} & \rotatebox[origin=l]{90}{StanfordCars} & \rotatebox[origin=l]{90}{VOC2007} \\
        \midrule
        \multicolumn{23}{c}{ \bf{Fine-tuning} } 
 \\
\multirow{2}{*}{5} & \xmark & 29.8 & 40.8 & 19.6 & 15.5 & 0.9 & 25.2 & 55.8 & 21.1 & 13.4 & 14.7 & 30.6 & 46.1 & 41.5 & 52.2 & 31.8 & 44.5 & 52.5 & 51.2 & 16.5 & 3.7 & 17.5 \\
 & \cmark & 63.3 & 88.8 & 91.3 & 73.0 & 16.6 & 51.8 & 79.3 & 52.2 & 23.1 & 84.0 & 60.4 & 55.8 & 44.3 & 60.5 & 67.3 & 86.9 & 61.8 & 59.2 & 70.8 & 56.3 & 82.4 \\
\multirow{2}{*}{20} & \xmark & 46.8 & 82.6 & 63.2 & 26.5 & 1.9 & 57.9 & 81.6 & 27.4 & 33.2 & 36.6 & 60.9 & 53.1 & 41.7 & 35.6 & 34.6 & 54.2 & 74.9 & 51.7 & 43.8 & 32.9 & 41.0 \\
 & \cmark & 72.2 & 93.3 & 91.9 & 76.0 & 17.2 & 60.0 & 90.4 & 57.9 & 42.7 & 84.2 & 92.0 & 53.9 & 46.4 & 93.1 & 86.4 & 90.8 & 72.4 & 59.4 & 82.9 & 69.9 & 82.9 \\
\multirow{2}{*}{50} & \xmark & 61.7 & 91.4 & 88.7 & 42.7 & 2.7 & 68.2 & 85.8 & 42.7 & 50.3 & 72.7 & 77.3 & 52.6 & 52.0 & 71.9 & 34.6 & 84.3 & 78.0 & 52.7 & 88.0 & 51.4 & 46.0 \\
 & \cmark & 75.7 & 94.0 & 93.3 & 79.1 & 17.5 & 71.7 & 94.9 & 58.7 & 51.6 & 85.1 & 95.2 & 55.0 & 59.1 & 89.7 & 86.4 & 91.1 & 78.6 & 62.0 & 88.4 & 76.8 & 85.7 \\
\multirow{2}{*}{full} & \xmark & 77.7 & 88.9 & 97.4 & 85.8 & 14.6 & 70.8 & 97.7 & 69.8 & 46.3 & 85.4 & 97.9 & 60.5 & 78.9 & 98.9 & 81.8 & 89.5 & 88.7 & 55.3 & 89.4 & 76.1 & 81.1 \\
 & \cmark & 80.3 & 94.0 & 97.8 & 87.0 & 19.1 & 70.0 & 98.1 & 68.8 & 50.7 & 87.7 & 98.5 & 61.9 & 81.0 & 99.5 & 88.5 & 91.6 & 91.0 & 70.6 & 89.4 & 75.8 & 85.7 \\
\midrule
        \multicolumn{23}{c}{ \bf{Linear Probing} } 
\\
\multirow{2}{*}{5} & \xmark & 58.1 & 88.1 & 87.0 & 56.1 & 10.1 & 58.1 & 73.8 & 33.9 & 28.2 & 70.0 & 52.8 & 51.0 & 40.9 & 77.5 & 89.5 & 66.5 & 57.0 & 49.4 & 75.3 & 53.1 & 43.3 \\
 & \cmark & 65.3 & 89.8 & 90.0 & 67.4 & 17.5 & 59.6 & 73.2 & 47.4 & 28.4 & 84.2 & 52.5 & 56.0 & 44.9 & 71.1 & 90.5 & 88.0 & 63.2 & 57.5 & 76.6 & 65.0 & 84.0 \\
\multirow{2}{*}{20} & \xmark & 70.0 & 92.2 & 91.0 & 69.2 & 16.6 & 71.0 & 81.2 & 48.6 & 39.8 & 81.3 & 73.1 & 51.3 & 51.3 & 92.4 & 93.8 & 83.7 & 65.4 & 58.0 & 84.4 & 73.0 & 82.1 \\
 & \cmark & 71.7 & 92.9 & 90.8 & 71.5 & 19.6 & 71.3 & 83.0 & 52.2 & 40.2 & 85.3 & 74.1 & 57.1 & 50.8 & 92.5 & 94.2 & 88.5 & 63.2 & 58.9 & 84.4 & 77.9 & 85.5 \\
\multirow{2}{*}{50} & \xmark & 74.1 & 92.8 & 92.1 & 73.7 & 21.1 & 74.5 & 88.1 & 53.6 & 44.3 & 84.0 & 80.5 & 51.3 & 58.7 & 95.1 & 93.8 & 88.2 & 75.2 & 62.3 & 87.0 & 81.0 & 84.2 \\
 & \cmark & 74.9 & 93.1 & 91.6 & 74.9 & 22.9 & 74.8 & 88.2 & 53.6 & 44.6 & 86.1 & 80.7 & 57.7 & 60.9 & 95.1 & 94.2 & 89.7 & 72.3 & 62.1 & 87.3 & 82.0 & 86.0 \\
\multirow{2}{*}{full} & \xmark & 78.4 & 92.7 & 94.5 & 79.6 & 25.2 & 74.0 & 93.4 & 67.8 & 44.3 & 88.1 & 86.9 & 64.0 & 65.8 & 98.8 & 93.9 & 89.9 & 83.2 & 71.4 & 88.1 & 80.8 & 85.0 \\
 & \cmark & 78.4 & 86.0 & 95.1 & 79.8 & 25.9 & 75.3 & 93.8 & 67.8 & 44.7 & 88.6 & 86.9 & 63.1 & 65.8 & 98.8 & 94.5 & 91.0 & 83.2 & 71.6 & 88.1 & 82.1 & 86.0 \\
        \bottomrule
    \end{tabular}
    }
    \vspace{3mm}
    \caption{Comparison of random and language-augmented initialization on CLIP (ViT-B32).
    }
    \label{table:experiment_breakdown_lang_init}
\end{table}

\begin{table}[t!]
    \centering
    \footnotesize
    \setlength{\tabcolsep}{2.2pt}
    \scalebox{0.78}{
    \begin{tabular}{ccc|c|cccccccccccccccccccc}
        \toprule
        \rotatebox[origin=l]{90}{Backbone} & \rotatebox[origin=l]{90}{Pretrain} & \rotatebox[origin=l]{90}{Language-Init} & \rotatebox[origin=l]{90}{Average Score} & \rotatebox[origin=l]{90}{Caltech101} & \rotatebox[origin=l]{90}{CIFAR10} & \rotatebox[origin=l]{90}{CIFAR100} & \rotatebox[origin=l]{90}{Country211} & \rotatebox[origin=l]{90}{DTD} & \rotatebox[origin=l]{90}{EuroSat} & \rotatebox[origin=l]{90}{FER2013} & \rotatebox[origin=l]{90}{FGVCAircraft} & \rotatebox[origin=l]{90}{Food101} & \rotatebox[origin=l]{90}{GTSRB} & \rotatebox[origin=l]{90}{HatefulMemes} & \rotatebox[origin=l]{90}{KittiDistance} & \rotatebox[origin=l]{90}{MNIST} & \rotatebox[origin=l]{90}{Flowers102} & \rotatebox[origin=l]{90}{OxfordPets} & \rotatebox[origin=l]{90}{PatchCamelyon} & \rotatebox[origin=l]{90}{SST2} & \rotatebox[origin=l]{90}{RESISC45} & \rotatebox[origin=l]{90}{StanfordCars} & \rotatebox[origin=l]{90}{VOC2007} \\
        \midrule
        \multicolumn{24}{c}{ \bf{Fine-tuning} } 
 \\
B32 & DeCLIP & \xmark & 58.8 & 88.2 & 78.0 & 59.4 & 6.0 & 58.3 & 73.8 & 20.1 & 26.4 & 61.4 & 66.0 & 51.5 & 30.2 & 77.0 & 98.1 & 74.9 & 53.1 & 52.7 & 68.2 & 57.6 & 75.5 \\
B32 & DeCLIP & \cmark & 64.5 & 92.9 & 91.6 & 77.5 & 11.7 & 55.2 & 77.1 & 38.4 & 20.0 & 76.2 & 69.7 & 54.5 & 43.6 & 73.1 & 95.5 & 85.6 & 61.4 & 52.0 & 71.7 & 60.2 & 82.3 \\
B32 & OpenCLIP & \xmark & 34.6 & 28.7 & 73.2 & 13.6 & 1.1 & 36.2 & 76.3 & 29.1 & 9.1 & 9.2 & 7.8 & 50.4 & 31.2 & 66.2 & 40.5 & 32.1 & 63.8 & 51.5 & 39.1 & 1.2 & 31.7 \\
B32 & OpenCLIP & \cmark & 64.8 & 91.6 & 91.6 & 74.1 & 10.0 & 54.3 & 72.2 & 46.6 & 23.0 & 79.0 & 82.3 & 54.4 & 33.8 & 85.9 & 83.8 & 86.1 & 62.8 & 53.1 & 76.0 & 53.7 & 82.2 \\
B16 & OpenCLIP & \xmark & 27.6 & 19.6 & 32.8 & 6.4 & 1.2 & 28.7 & 76.1 & 15.6 & 3.5 & 6.0 & 7.1 & 48.2 & 46.1 & 60.8 & 33.2 & 6.0 & 57.1 & 51.6 & 30.9 & 1.7 & 20.1 \\
B16 & OpenCLIP & \cmark & 66.2 & 86.8 & 91.5 & 74.6 & 17.0 & 60.6 & 79.8 & 45.2 & 15.4 & 84.2 & 60.6 & 54.1 & 34.1 & 85.8 & 86.0 & 88.2 & 67.4 & 55.0 & 72.5 & 82.9 & 83.1 \\
\midrule
        \multicolumn{24}{c}{ \bf{Linear Probing} } 
\\
B32 & DeCLIP & \xmark & 57.2 & 88.4 & 86.8 & 60.6 & 7.9 & 58.6 & 70.4 & 29.8 & 23.9 & 63.9 & 29.2 & 50.5 & 31.5 & 68.3 & 98.3 & 74.8 & 60.8 & 49.5 & 67.3 & 59.6 & 64.9 \\
B32 & DeCLIP & \cmark & 62.5 & 93.0 & 92.0 & 73.3 & 12.8 & 62.1 & 72.2 & 36.3 & 23.9 & 76.2 & 29.2 & 54.7 & 45.7 & 68.3 & 98.6 & 84.9 & 61.0 & 52.6 & 66.4 & 65.2 & 80.2 \\
B32 & OpenCLIP & \xmark & 61.9 & 89.7 & 88.1 & 64.2 & 8.1 & 53.3 & 78.8 & 33.2 & 28.8 & 68.6 & 64.4 & 50.6 & 36.6 & 80.7 & 92.2 & 72.7 & 61.1 & 52.5 & 73.3 & 74.3 & 67.2 \\
B32 & OpenCLIP & \cmark & 68.6 & 91.2 & 91.2 & 72.0 & 14.4 & 68.9 & 76.9 & 45.9 & 31.2 & 81.4 & 65.1 & 52.9 & 46.7 & 86.0 & 94.0 & 87.9 & 65.9 & 54.7 & 79.1 & 83.0 & 84.5 \\
B16 & OpenCLIP & \xmark & 62.9 & 90.2 & 85.5 & 63.9 & 9.2 & 65.6 & 78.3 & 24.3 & 33.0 & 74.8 & 62.3 & 51.9 & 30.8 & 88.1 & 94.0 & 74.5 & 52.0 & 50.2 & 78.7 & 77.2 & 73.1 \\
B16 & OpenCLIP & \cmark & 69.7 & 93.2 & 91.6 & 72.7 & 18.1 & 69.4 & 79.7 & 46.3 & 34.3 & 84.0 & 64.1 & 53.4 & 38.7 & 92.8 & 95.0 & 88.2 & 66.3 & 57.4 & 78.5 & 86.0 & 85.0 \\
L14 & OpenCLIP & \xmark & 66.5 & 91.7 & 92.1 & 70.0 & 12.1 & 66.3 & 80.1 & 36.4 & 37.2 & 81.0 & 72.0 & 51.9 & 27.1 & 87.6 & 96.0 & 81.0 & 53.2 & 52.2 & 81.4 & 84.1 & 76.6 \\
L14 & OpenCLIP & \cmark & 72.5 & 92.9 & 94.1 & 78.8 & 22.6 & 72.0 & 86.0 & 52.9 & 40.1 & 89.2 & 74.2 & 54.7 & 41.1 & 86.4 & 97.2 & 91.5 & 60.2 & 58.8 & 83.3 & 89.3 & 84.9 \\
L14 $^\dagger$ & CLIP & \xmark & 68.3 & 93.3 & 92.1 & 70.2 & 19.6 & 65.0 & 85.1 & 42.5 & 46.0 & 88.0 & 72.7 & 51.3 & 45.0 & 80.9 & 96.6 & 83.8 & 60.3 & 56.5 & 80.9 & 79.5 & 57.8 \\
L14 $^\dagger$ & CLIP & \cmark & 75.2 & 94.5 & 95.3 & 79.3 & 34.2 & 70.0 & 87.0 & 58.4 & 50.1 & 93.8 & 74.2 & 59.8 & 35.0 & 83.0 & 98.0 & 94.2 & 65.8 & 71.3 & 87.8 & 85.7 & 86.5 \\
        \bottomrule
    \end{tabular}
    }
    \vspace{3mm}
    \caption{Comparisons of random and language-augmented initialization for language-image model adaptation with more checkpoints under 5-shot settings. $^\dagger$ Input image size 336$\times$336 .
    }
    \label{table:additional_experiment_breakdown_lang_init}
\end{table}

\begin{table}[t!]
    \centering
    \footnotesize
    \setlength{\tabcolsep}{2.2pt}
    \scalebox{0.68}{
    \begin{tabular}{ccc|c|cccccccccccccccccccc}
        \toprule
        \rotatebox[origin=l]{90}{Backbone} & \rotatebox[origin=l]{90}{Pretrain Method} & \rotatebox[origin=l]{90}{Pretrain Dataset} & \rotatebox[origin=l]{90}{Average Score} & \rotatebox[origin=l]{90}{Caltech101} & \rotatebox[origin=l]{90}{CIFAR10} & \rotatebox[origin=l]{90}{CIFAR100} & \rotatebox[origin=l]{90}{Country211} & \rotatebox[origin=l]{90}{DTD} & \rotatebox[origin=l]{90}{EuroSat} & \rotatebox[origin=l]{90}{FER2013} & \rotatebox[origin=l]{90}{FGVCAircraft} & \rotatebox[origin=l]{90}{Food101} & \rotatebox[origin=l]{90}{GTSRB} & \rotatebox[origin=l]{90}{HatefulMemes} & \rotatebox[origin=l]{90}{KittiDistance} & \rotatebox[origin=l]{90}{MNIST} & \rotatebox[origin=l]{90}{Flowers102} & \rotatebox[origin=l]{90}{OxfordPets} & \rotatebox[origin=l]{90}{PatchCamelyon} & \rotatebox[origin=l]{90}{SST2} & \rotatebox[origin=l]{90}{RESISC45} & \rotatebox[origin=l]{90}{StanfordCars} & \rotatebox[origin=l]{90}{VOC2007} \\
        \midrule
        \multicolumn{24}{c}{ \bf{Academic Track} } \\
B32 & CLIP & YFCC (15M) & 32.0 & 55.9 & 70.2 & 33.7 & 5.1 & 15.6 & 29.9 & 23.3 & 2.5 & 32.1 & 5.6 & 53.5 & 39.9 & 14.3 & 48.7 & 19.1 & 50.0 & 49.0 & 17.3 & 2.3 & 71.6 \\
B32 & DeCLIP & YFCC (15M) & 37.9 & 69.1 & 85.3 & 55.5 & 8.8 & 26.3 & 27.5 & 29.8 & 2.9 & 48.6 & 10.4 & 51.7 & 28.4 & 11.1 & 59.8 & 34.9 & 50.6 & 49.9 & 25.0 & 4.0 & 77.5 \\
B32 & FILIP & YFCC (15M) & 34.5 & 65.1 & 83.6 & 50.8 & 7.5 & 23.2 & 23.4 & 23.3 & 3.0 & 40.8 & 7.4 & 50.8 & 24.2 & 7.9 & 49.5 & 22.5 & 51.8 & 49.9 & 25.9 & 3.1 & 77.1 \\
B32 & SLIP & YFCC (15M) & 31.2 & 58.8 & 69.5 & 39.0 & 5.1 & 14.0 & 19.5 & 22.8 & 1.3 & 32.8 & 6.7 & 52.9 & 29.0 & 10.3 & 45.9 & 24.4 & 50.0 & 49.9 & 17.5 & 2.2 & 71.6 \\
 \midrule
        \multicolumn{24}{c}{\bf{Industry Track}} \\
B32 & CLIP & WebImageText (400M) & 56.8 & 87.4 & 89.8 & 65.2 & 17.2 & 44.1 & 46.0 & 42.0 & 19.5 & 84.0 & 32.7 & 56.0 & 29.0 & 48.4 & 66.5 & 87.2 & 60.7 & 58.8 & 60.0 & 59.6 & 82.6 \\
B32 & DeCLIP & DeCLIP (88M) & 51.0 & 89.2 & 90.9 & 66.8 & 12.0 & 44.9 & 39.9 & 23.3 & 9.0 & 75.0 & 11.4 & 53.9 & 39.7 & 13.6 & 83.0 & 83.7 & 55.3 & 50.1 & 47.6 & 49.7 & 80.6 \\
B32 & OpenCLIP & LAION (400M) & 57.5 & 90.1 & 90.8 & 70.6 & 14.8 & 54.5 & 51.7 & 42.4 & 16.6 & 80.8 & 42.0 & 52.8 & 31.6 & 37.6 & 65.9 & 86.5 & 50.1 & 52.3 & 57.5 & 79.3 & 82.1 \\
B16 & CLIP & WebImageText (400M) & 60.0 & 88.9 & 90.8 & 68.2 & 22.8 & 44.8 & 54.7 & 48.5 & 24.3 & 88.7 & 43.5 & 58.1 & 27.0 & 52.0 & 69.4 & 89.0 & 54.0 & 60.9 & 65.6 & 64.8 & 83.7 \\
B16 & OpenCLIP & LAION (400M) & 59.1 & 90.3 & 90.2 & 70.0 & 17.4 & 48.7 & 48.6 & 44.9 & 15.3 & 83.2 & 38.6 & 53.4 & 23.9 & 71.1 & 63.7 & 87.6 & 51.0 & 57.2 & 63.6 & 81.6 & 82.2 \\
L14 & CLIP & WebImageText (400M) & 65.9 & 92.6 & 95.6 & 78.2 & 31.8 & 55.4 & 64.1 & 50.0 & 31.9 & 93.1 & 50.5 & 59.3 & 13.5 & 76.2 & 79.1 & 93.5 & 51.2 & 68.9 & 71.0 & 77.9 & 83.9 \\
L14 & OpenCLIP & LAION (400M) & 62.5 & 92.9 & 93.5 & 76.2 & 21.2 & 56.4 & 53.7 & 50.3 & 20.8 & 89.1 & 45.6 & 55.3 & 28.8 & 63.9 & 70.9 & 89.7 & 50.5 & 57.0 & 64.1 & 87.4 & 82.3 \\
L14 $\dagger$ & CLIP & WebImageText (400M) & 66.8 & 92.4 & 94.9 & 77.0 & 34.5 & 56.0 & 63.0 & 48.3 & 33.3 & 93.9 & 52.3 & 60.0 & 11.5 & 79.0 & 78.5 & 93.8 & 62.3 & 70.6 & 71.3 & 79.3 & 84.0 \\
        \bottomrule
    \end{tabular}
    }
    \vspace{3mm}
    \caption{Zero-shot results of more checkpoints in Academic and Industry Tracks. $^\dagger$ Input image size 336$\times$336.
    }
    \label{table:experiment_breakdown_zeroshot_more_checkpoints}
\end{table}

\section{Benefits of External Knowledge in Model Adaptation}
\label{sec:benefit_knowledge}

We also explore the benefits of the external knowledge to models that are  pre-trained without the external knowledge (\eg CLIP). On CLIP, we compare the effect of adding different combinations of external knowledge (Wiktionary, the nubmer of GPT3 knowledge items). The results are summarized in \ref{table:clip_knowledge_gain}, and detailed in Table~\ref{tab:benefit_zeroshot_clip_breakdown}.

In zero-shot settings, we find that when the external knowledge is available,  CLIP demonstrates consistent improvement on four datasets and considerable gains on the other three datasets.
This suggests that the knowledge can benefit language-image models (though varying between datasets) as a new language prompting technique for some datasets, even if the pre-trained model is trained without the external knowledge.

In few- / full-shot settings, we argue that the pre-trained model can \emph{selectively} incorporate different knowledge sources to achieve the best adaptation performance.  One simple strategy is to train the model with different knowledge sources, compare the split \emph{validation} accuracy of checkpoints with different knowledge sources, and use the best one for testing. We called it as {\bf knowledge-augmented adaptation}, in contrast to the baseline method {\bf knowledge-free adaptation}, where no collected external knowledge is employed at all.
We find such simple strategy is already effective for linear probing and fine-tuning CLIP.  As shown in Table.~\ref{table:clip_knowledge_gain}, knowledge-based adaptation of CLIP consistently improves over knowledge-free adaptation both in terms of accuracy and the number of wins. Notably, by selectively incorporating the external knowledge, it shows a significant 1.8 improvement for 5-shot CLIP fine-tuning.  Note that such gain comes for \emph{free}, even when the base CLIP model is \emph{not} pre-trained with the external knowledge.  We believe more sophisticated knowledge adaptation strategy can yield even better performance and we leave that to future work. 

These experiments show that the collected external knowledge on \shortname{} is a useful resource for improving the adaptation of language-augmented visual models.

\begin{table}[t!]
	\centering
    \scalebox{0.98}{
    \begin{tabular}{l | cc | cc}
    \toprule
         \multirow{2}{*}{Adaptation Methods} &
         \multicolumn{2}{c|}{5-shot} & \multicolumn{2}{c}{Full-shot} \\
           & LP & FT & LP & FT
          \\          
        \midrule
        Knowledge-free adaptation & 65.35 {\tiny $\pm$ 1.24} & 63.29 {\tiny $\pm$ 3.18} & 78.40 & 79.97 \\
        \rowcolor{gray!30} 
        Knowledge-augmented adaptation & {\bf 65.83} {\tiny $\pm$ 1.50} & {\bf 65.10} {\tiny $\pm$ 2.08} & {\bf 78.75} & {\bf 80.32} \\
        \midrule
        Gain & +0.48 & +1.81 & +0.35 & +0.35 \\
        \# {\bf \color{emerald} win} / {\bf \color{black} tie} / {\bf \color{orange} lose} & {\bf \color{emerald} 7} / {\bf \color{black} 8} / {\bf \color{orange} 5} & {\bf \color{emerald} 8} / {\bf \color{black} 8} / {\bf \color{orange} 4} & {\bf \color{emerald} 12} / {\bf \color{black} 4} / {\bf \color{orange} 4} & {\bf \color{emerald} 10} / {\bf \color{black} 5} / {\bf \color{orange} 5} \\
        \bottomrule
    \end{tabular}
    }
    \vspace{1mm}
    \caption{Benefits of adapting CLIP with external knowledge.}
    \label{table:clip_knowledge_gain}
\end{table}

\begin{table*}[h!]
    \centering
    \footnotesize
    \setlength{\tabcolsep}{2.2pt}
    \scalebox{0.82}{
    \begin{tabular}{cc|c|cccccccccccccccccccc}
        \toprule
        Wiki & \#GPT3 & mAcc & \rotatebox[origin=l]{90}{Caltech101} & \rotatebox[origin=l]{90}{CIFAR10} & \rotatebox[origin=l]{90}{CIFAR100} & \rotatebox[origin=l]{90}{Country211} & \rotatebox[origin=l]{90}{DTD} & \rotatebox[origin=l]{90}{EuroSat} & \rotatebox[origin=l]{90}{FER2013} & \rotatebox[origin=l]{90}{FGVCAircraft} & \rotatebox[origin=l]{90}{Food101} & \rotatebox[origin=l]{90}{GTSRB} & \rotatebox[origin=l]{90}{HatefulMemes} & \rotatebox[origin=l]{90}{KittiDistance} & \rotatebox[origin=l]{90}{MNIST} & \rotatebox[origin=l]{90}{Flowers102} & \rotatebox[origin=l]{90}{OxfordPets} & \rotatebox[origin=l]{90}{PatchCamelyon} & \rotatebox[origin=l]{90}{SST2} & \rotatebox[origin=l]{90}{RESISC45} & \rotatebox[origin=l]{90}{StanfordCars} & \rotatebox[origin=l]{90}{VOC2007} \\
        \midrule
        \multicolumn{23}{c}{ \bf{Zero-Shot} } \\
         & -- & 56.8 & 87.4 & 89.8 & 65.1 & 17.2 & 44.4 & 45.5 & 42.3 & 19.6 & 84.0 & 32.5 & 56.0 & 29.0 & 48.2 & 66.5 & 87.2 & 60.6 & 58.6 & 60.0 & 59.7 & 82.6 \\
        $\checkmark$ & -- & 52.1 & 83.6 & 85.4 & 56.1 & 13.2 & 44.4 & 40.3 & 39.6 & 18.4 & 79.8 & 28.9 & 55.5 & 27.3 & 10.6 & 66.2 & 81.0 & 52.4 & {\bf\cellcolor{emerald!30}62.2} & 57.8 & 59.7 & 80.1 \\
        $\checkmark$ & 1 & 53.3 & 86.8 & 88.4 & 57.6 & 14.9 & {\bf\cellcolor{emerald!30}47.0} & 36.6 & 42.0 & 18.4 & 81.8 & {\bf\cellcolor{emerald!30}34.0} & 55.5 & 28.3 & 19.1 & {\bf\cellcolor{emerald!30}67.6} & 85.3 & 56.6 & {\bf\cellcolor{emerald!30}61.9} & 58.1 & 45.4 & 81.3 \\
        $\checkmark$ & 5 & 54.2 & 87.3 & 88.8 & 63.9 & 16.0 & {\bf\cellcolor{emerald!30}50.1} & 41.1 & {\bf\cellcolor{emerald!30}43.4} & 18.5 & 82.3 & {\bf\cellcolor{emerald!30}36.4} & 55.5 & {\bf\cellcolor{emerald!30}32.2} & 11.9 & {\bf\cellcolor{emerald!30}69.5} & 87.0 & 52.9 & {\bf\cellcolor{emerald!30}62.0} & 58.6 & 45.1 & 82.0 \\
         & 1 & 53.2 & 86.1 & 87.6 & 61.5 & 14.7 & 43.6 & {\bf\cellcolor{emerald!30}51.2} & 33.3 & 18.5 & 80.0 & 31.4 & 55.5 & {\bf\cellcolor{emerald!30}33.2} & 23.1 & {\bf\cellcolor{emerald!30}66.6} & 84.5 & 51.5 & {\bf\cellcolor{emerald!30}61.2} & 53.7 & 45.4 & 81.1 \\
         & 5 & 54.5 & 87.0 & 88.7 & 63.9 & 16.1 & {\bf\cellcolor{emerald!30}49.5} & {\bf\cellcolor{emerald!30}50.9} & {\bf\cellcolor{emerald!30}44.0} & 18.6 & 81.9 & {\bf\cellcolor{emerald!30}35.5} & 55.5 & {\bf\cellcolor{emerald!30}31.2} & 14.7 & {\bf\cellcolor{emerald!30}69.4} & {\bf\cellcolor{emerald!30}87.3} & 49.9 & {\bf\cellcolor{emerald!30}62.2} & 57.5 & 45.1 & 82.0 \\
        \midrule
        \multicolumn{23}{c}{ \bf{5-Shot Linear Probing} } \\
 & -- & 65.3 & 89.8 & 90.0 & 67.4 & 17.5 & 59.6 & 73.2 & 47.4 & 28.4 & 84.2 & 52.5 & 56.0 & 44.9 & 71.1 & 90.5 & 88.0 & 63.2 & 57.5 & 76.6 & 65.0 & 84.0 \\
$\checkmark$ & 5 & 65.2 & 89.5 & 89.3 & 67.4 & 17.5 & {\bf\cellcolor{emerald!30}61.9} & 72.0 & {\bf\cellcolor{emerald!30}48.4} & {\bf\cellcolor{emerald!30}28.5} & 84.2 & 52.2 & 55.5 & 39.3 & {\bf\cellcolor{emerald!30}76.2} & {\bf\cellcolor{emerald!30}91.1} & {\bf\cellcolor{emerald!30}88.1} & {\bf\cellcolor{emerald!30}63.5} & {\bf\cellcolor{emerald!30}58.4} & 72.7 & 65.0 & 83.5 \\
$\checkmark$ & -- & {\bf\cellcolor{emerald!30}65.6} & 89.2 & {\bf\cellcolor{emerald!30}90.7} & 67.4 & 17.5 & {\bf\cellcolor{emerald!30}61.0} & {\bf\cellcolor{emerald!30}74.1} & 45.8 & 28.4 & 84.2 & 52.5 & 55.5 & 37.5 & {\bf\cellcolor{emerald!30}76.4} & {\bf\cellcolor{emerald!30}91.0} & 88.0 & {\bf\cellcolor{emerald!30}67.8} & {\bf\cellcolor{emerald!30}59.7} & 76.6 & 65.0 & 83.6 \\
 & 5 & {\bf\cellcolor{emerald!30}65.8} & 89.3 & 89.2 & 66.5 & 17.5 & {\bf\cellcolor{emerald!30}61.4} & {\bf\cellcolor{emerald!30}73.5} & {\bf\cellcolor{emerald!30}48.4} & {\bf\cellcolor{emerald!30}28.5} & 84.2 & {\bf\cellcolor{emerald!30}53.2} & 55.5 & {\bf\cellcolor{emerald!30}45.5} & {\bf\cellcolor{emerald!30}76.2} & {\bf\cellcolor{emerald!30}91.1} & {\bf\cellcolor{emerald!30}88.6} & {\bf\cellcolor{emerald!30}63.3} & {\bf\cellcolor{emerald!30}59.9} & 76.6 & 65.0 & 83.2 \\
        \midrule
        \multicolumn{23}{c}{ \bf{5-Shot Fine-tuning} } \\
 & -- & 63.3 & 88.8 & 91.3 & 73.0 & 16.6 & 51.8 & 79.3 & 52.2 & 23.1 & 84.0 & 60.4 & 55.8 & 44.3 & 60.5 & 67.3 & 86.9 & 61.8 & 59.2 & 70.8 & 56.3 & 82.4 \\
$\checkmark$ & 5 & 62.0 & 88.0 & 90.3 & 73.0 & 16.6 & {\bf\cellcolor{emerald!30}56.4} & {\bf\cellcolor{emerald!30}81.3} & 52.2 & 22.3 & 83.9 & 60.4 & 55.4 & {\bf\cellcolor{emerald!30}47.0} & 60.5 & {\bf\cellcolor{emerald!30}79.9} & {\bf\cellcolor{emerald!30}87.5} & {\bf\cellcolor{emerald!30}65.0} & {\bf\cellcolor{emerald!30}62.7} & 19.8 & 56.3 & 81.8 \\
$\checkmark$ & -- & {\bf\cellcolor{emerald!30}65.1} & 88.8 & 88.9 & 73.0 & 16.6 & 44.9 & 79.3 & 52.2 & {\bf\cellcolor{emerald!30}24.4} & {\bf\cellcolor{emerald!30}84.1} & {\bf\cellcolor{emerald!30}66.7} & 55.4 & {\bf\cellcolor{emerald!30}44.5} & {\bf\cellcolor{emerald!30}82.4} & {\bf\cellcolor{emerald!30}79.4} & 86.9 & 60.6 & {\bf\cellcolor{emerald!30}63.3} & {\bf\cellcolor{emerald!30}71.8} & 56.3 & 82.4 \\
 & 5 & {\bf\cellcolor{emerald!30}64.2} & 88.5 & 89.8 & 73.0 & 16.6 & {\bf\cellcolor{emerald!30}53.1} & {\bf\cellcolor{emerald!30}82.0} & 52.2 & 21.6 & 83.9 & 60.4 & 55.5 & 34.9 & {\bf\cellcolor{emerald!30}81.4} & {\bf\cellcolor{emerald!30}77.7} & {\bf\cellcolor{emerald!30}87.9} & 54.2 & {\bf\cellcolor{emerald!30}62.6} & 70.8 & 56.3 & 81.8 \\
        \midrule
        \multicolumn{23}{c}{ \bf{Full-Shot Linear Probing} } \\
 & -- & 78.4 & 86.0 & 95.1 & 79.8 & 25.9 & 75.3 & 93.8 & 67.8 & 44.7 & 88.6 & 86.9 & 63.1 & 65.8 & 98.8 & 94.5 & 91.0 & 83.2 & 71.6 & 88.1 & 82.1 & 86.0 \\
$\checkmark$ & 5 & {\bf\cellcolor{emerald!30}78.7} & {\bf\cellcolor{emerald!30}92.8} & 94.9 & 79.8 & 25.7 & 75.1 & 93.3 & 67.7 & {\bf\cellcolor{emerald!30}44.9} & 88.6 & {\bf\cellcolor{emerald!30}87.0} & {\bf\cellcolor{emerald!30}64.1} & {\bf\cellcolor{emerald!30}66.8} & {\bf\cellcolor{emerald!30}98.9} & {\bf\cellcolor{emerald!30}94.9} & 90.8 & {\bf\cellcolor{emerald!30}83.7} & 70.3 & 87.0 & 81.9 & 85.8 \\
$\checkmark$ & -- & {\bf\cellcolor{emerald!30}78.8} & {\bf\cellcolor{emerald!30}93.2} & {\bf\cellcolor{emerald!30}95.2} & {\bf\cellcolor{emerald!30}79.9} & 25.7 & 73.5 & 93.3 & {\bf\cellcolor{emerald!30}67.8} & {\bf\cellcolor{emerald!30}44.8} & 88.2 & {\bf\cellcolor{emerald!30}86.9} & {\bf\cellcolor{emerald!30}64.1} & 65.8 & 98.8 & {\bf\cellcolor{emerald!30}94.9} & {\bf\cellcolor{emerald!30}91.1} & {\bf\cellcolor{emerald!30}83.7} & 71.6 & {\bf\cellcolor{emerald!30}88.3} & {\bf\cellcolor{emerald!30}82.2} & 86.0 \\
 & 5 & {\bf\cellcolor{emerald!30}78.6} & {\bf\cellcolor{emerald!30}93.1} & 94.9 & 79.8 & 25.7 & 74.7 & 93.4 & 65.4 & 44.6 & 88.6 & {\bf\cellcolor{emerald!30}87.0} & {\bf\cellcolor{emerald!30}64.1} & {\bf\cellcolor{emerald!30}66.7} & {\bf\cellcolor{emerald!30}98.9} & {\bf\cellcolor{emerald!30}94.9} & 90.8 & {\bf\cellcolor{emerald!30}83.7} & 70.3 & {\bf\cellcolor{emerald!30}88.3} & 81.9 & 85.7 \\        \midrule
        \multicolumn{23}{c}{ \bf{Full-Shot Fine-tuning} } \\
 & -- & 80.0 & 93.1 & 97.5 & 87.3 & 19.2 & 70.9 & 98.0 & 70.2 & 47.7 & 88.0 & 98.5 & 61.1 & 81.9 & 99.5 & 87.3 & 90.7 & 90.6 & 66.7 & 89.4 & 76.1 & 85.5 \\
$\checkmark$ & 5 & {\bf\cellcolor{emerald!30}80.0} & {\bf\cellcolor{emerald!30}93.5} & {\bf\cellcolor{emerald!30}97.5} & 84.4 & {\bf\cellcolor{emerald!30}19.4} & {\bf\cellcolor{emerald!30}72.5} & 97.9 & 69.5 & 47.7 & 87.8 & {\bf\cellcolor{emerald!30}98.5} & 61.1 & 81.3 & {\bf\cellcolor{emerald!30}99.5} & {\bf\cellcolor{emerald!30}89.5} & {\bf\cellcolor{emerald!30}92.5} & 90.1 & 66.7 & {\bf\cellcolor{emerald!30}90.0} & 76.1 & 85.4 \\
$\checkmark$ & -- & {\bf\cellcolor{emerald!30}80.1} & 93.0 & 97.4 & 87.3 & 19.2 & {\bf\cellcolor{emerald!30}73.0} & 98.0 & 70.0 & 47.7 & 87.6 & {\bf\cellcolor{emerald!30}98.5} & 61.1 & 80.9 & 99.5 & {\bf\cellcolor{emerald!30}89.6} & {\bf\cellcolor{emerald!30}92.2} & 89.2 & 66.7 & {\bf\cellcolor{emerald!30}89.5} & {\bf\cellcolor{emerald!30}76.2} & 85.5 \\
 & 5 & {\bf\cellcolor{emerald!30}80.3} & {\bf\cellcolor{emerald!30}93.6} & {\bf\cellcolor{emerald!30}97.6} & {\bf\cellcolor{emerald!30}87.4} & 19.2 & 70.9 & {\bf\cellcolor{emerald!30}98.1} & {\bf\cellcolor{emerald!30}71.7} & 47.7 & 87.8 & 98.4 & 61.1 & 81.6 & 99.3 & {\bf\cellcolor{emerald!30}89.3} & {\bf\cellcolor{emerald!30}92.5} & 87.3 & {\bf\cellcolor{emerald!30}70.9} & {\bf\cellcolor{emerald!30}90.0} & 76.1 & {\bf\cellcolor{emerald!30}85.9} \\        \bottomrule
    \end{tabular}
    }
    \vspace{-0mm}
    \caption{Benefit of external knowledge for CLIP. For adaptation with linear probing and fine-tuning, we make use of the external knowledge when it has a higher validation accuracy.}
    \label{tab:benefit_zeroshot_clip_breakdown}
      \vspace{-3mm}
\end{table*}

\end{document}